\documentstyle[11pt,rotate,lingmacros]{report}  

\input{psfig}

\makeatletter
\renewcommand{\baselinestretch}{1.5}

\def\@xfloat#1[#2]{\ifhmode \@bsphack\@floatpenalty -\@Mii\else
   \@floatpenalty-\@Miii\fi\def\@captype{#1}\ifinner
      \@parmoderr\@floatpenalty\z@
      \else\@next\@currbox\@freelist{\@tempcnta\csname ftype@#1\endcsname
      \multiply\@tempcnta\@xxxii\advance\@tempcnta\sixt@@n \@tfor \@tempa
      :=#2\do {\if\@tempa h\advance\@tempcnta \@ne\fi \if\@tempa
      t\advance\@tempcnta \tw@\fi \if\@tempa b\advance\@tempcnta 4\relax\fi
      \if\@tempa p\advance\@tempcnta 8\relax\fi
      }\global\count\@currbox\@tempcnta}\@fltovf\fi
      \global\setbox\@currbox\vbox\bgroup
      \def\baselinestretch{1}\@normalsize \boxmaxdepth\z@ \hsize\columnwidth
      \@parboxrestore}
\long\def\@footnotetext#1{\insert\footins{\def\baselinestretch{1}\footnotesize
    \interlinepenalty\interfootnotelinepenalty 
    \splittopskip\footnotesep
    \splitmaxdepth \dp\strutbox \floatingpenalty \@MM
    \hsize\columnwidth \@parboxrestore
   \edef\@currentlabel{\csname p@footnote\endcsname\@thefnmark}\@makefntext
    {\rule{\z@}{\footnotesep}\ignorespaces
      #1\strut}}}

\def\dept#1{\gdef\@dept{#1}}

\def\groupchair#1{\gdef\@groupchair{#1}}
\def\submitdate#1{\gdef\@submitdate{#1}}
\def\copyrightyear#1{\gdef\@copyrightyear{#1}} 
\def\@title{}\def\@author{}\def\@dept{Computer and Information Science}
\def\@groupchair{}%
\def\@submitdate{\ifcase\the\month\or
  January\or February\or March\or April\or May\or June\or
  July\or August\or September\or October\or November\or December\fi
  \space \number\the\year}
\ifnum\month=12
    \@tempcnta=\year \advance\@tempcnta by 1
    \edef\@copyrightyear{\number\the\@tempcnta}
\else
    \def\@copyrightyear{\number\the\year}
\fi
\newif\ifcopyright \newif\iffigurespage \newif\iftablespage
\copyrighttrue \figurespagetrue \tablespagetrue

\def\titlep{%
	\thispagestyle{empty}%
	\null\vskip.1in%
	\begin{center}
		{\Large\sc\expandafter{\@title}} 
		\vskip.3in%
		{\large \expandafter{\@author}} 
		\vskip.3in%
		{\large {\sc A Dissertation} \\
		in \\
		{\expandafter{\@dept}}}
		\vfill
		Presented to the Faculties of the University of Pennsylvania
		in Partial Fulfillment of the Requirements
		for the Degree of Doctor of Philosophy \\ 
		\vskip.3in%
		1993
	\end{center}
	\vskip1in%
	\hrule width7cm 
	\vskip.1in%
Mark J. Steedman\\
	Supervisor of Dissertation\\
	\vskip.5in%
	\hrule width7cm 
	\vskip.1in%
	\noindent
	\@groupchair \\ 
	Graduate Group Chairperson \\
	\vfill\newpage}

\def\copyrightpage{%
	\thispagestyle{empty}%
	\null\vfill
	\begin{center}
		\Large\copyright\ Copyright\\[3mm]
		\@author\\[3mm]
	       1993 
	\end{center}
	\vfill\newpage}

\def\beforepreface{%
	\pagenumbering{roman}%
	\pagestyle{plain}%
	\titlep%
	\ifcopyright\copyrightpage\fi}

\def\prefacesection#1{%
	\pagestyle{plain}%
	\chapter*{#1}
	\addcontentsline{toc}{chapter}{#1}}

\def\afterpreface{\newpage
	\tableofcontents
	\newpage
	\iftablespage
		{\addvspace{10pt}
		\let\saveaddvspace=\addvspace
		\def\addvspace##1{}
		\listoftables
		\let\addvspace=\saveaddvspace}
		\newpage
	\fi
	\iffigurespage
		{\addvspace{10pt}
		\let\saveaddvspace=\addvspace
		\def\addvspace##1{}
		\listoffigures
		\let\addvspace=\saveaddvspace}
		\newpage
	\fi
	\pagenumbering{arabic}
	\pagestyle{plain}}

\makeatother

\renewcommand{\baselinestretch}{1.0}

\newcommand{\bkslf}{$ \backslash $}		

\newcommand{\RULE}[2]		                
{\begin{tabular}{c}
 \( {#1} \) \\ \hline \( {#2} \)
\end{tabular}}

\newcommand{\reffig}[1]{figure~\ref{#1}}	
\newcommand{\refch}[1]{chapter~\ref{#1}}

\newcommand{\refsec}[1]{section~\ref{#1}}

\newcommand{\refpage}[1]{page~\pageref{#1}}
\newcommand{\refapp}[1]{Appendix~\ref{#1}}

\newcommand{\SetMyTabs}{\verb+  + \= \verb+  + \= \verb+  +   
         \= \verb+  + \= \verb+  + \= \verb+  + \= \verb+  + \=  
         \verb+  + \= \verb+  + \= \verb+  + \= \verb+  + \= \kill}      
\newcommand{\comment}[1]{}
\newcommand{\GoesTo}{\(\longrightarrow\)}
\newcommand{\forw}{\verb+>+}
\newcommand{\back}{\verb+<+}
\newcommand{\up}{\(\mid\)}
\newcommand{\cB}{\mbox{\bf B}} 
\newcommand{\cT}{\mbox{\bf T}} 
\newcommand{\cS}{\mbox{\bf S}}  
\newcommand{\SUB}[1]{ \( \! \!_{ {#1}}\)}
\newcommand{\SSUB}[1]{ \( \! \!_{_{ {#1}}}\)}
\newcommand{\SUP}[1]{ \( \! \!^{ {#1}}\)}

\newcommand{\Center}[1]{\begin{center} #1 \end{center}}

\newcommand{\komment}[1]{{\em \ \ \% #1}} 

\newcommand{\pf}[1]{\noindent{\bf proof }  {#1} \hfill~$\Box$}
\newcommand{\implies}{\: \supset \:}
\newcommand{\And}{\: \wedge \:}
\newcommand{\arr}{-\!\!\!\!\rightarrow}   			
\newcommand{\larr}{\leftarrow\!\!\!\!-}				
\newcommand{\darr}{-\!\!\!\!\rightarrow\!\!\!\!\!\rightarrow}   
\newcommand{\invdarr}{\leftarrow\!\!\!\!\!\leftarrow\!\!\!\! -} 
\newcommand{\drpl}{\invdarr\!\!\!\!\!\!\darr}   		
\newcommand{\CTRarr}{\mbox{$\arr_{\mbox{ctr}}$}}
\newtheorem{defi}{Definition}
\newtheorem{LEMMA}{Lemma}
\newtheorem{COROLLARY}{Corollary}
\newtheorem{thm}{Theorem}

\newcommand{\lemma}[1]{\begin{LEMMA} {\rm {#1}} \end{LEMMA}}
\newcommand{\corollary}[1]{\begin{COROLLARY} {\rm {#1}} \end{COROLLARY}}
\newcommand{\theorem}[1]{\begin{thm} {\rm {#1}} \end{thm}}
\newcommand{\kases}{    
   \vspace{-3.2mm}
   \begin{list}
 {???}
 {\setlength{\leftmargin}{4.1mm}
  \setlength{\labelwidth}{2mm}}
  \setlength{\parsep}{1mm}
  \setlength{\itemsep}{0.01mm}
  \setlength{\topsep}{0.01mm}
  \setlength{\parskip}{0mm}
  \setlength{\parindent}{0mm}}
\newcommand{\Endkases}{\end{list}\vspace{-3.2mm}}
\newcommand{\kase}[1]{\item[case {#1}:\ ]}
\newcommand{\etal}{{\em et al.}}
\newcommand{\UL}{\_\hspace{-0.6mm}\_}  
\newcommand{\Kons}{\verb+|+}
\newcommand{\VEEPEE}{{\sl VP}}
\newcommand{\tuple}[1]{\(\langle\)#1\(\rangle\)}

\newcommand{\zp}{\rule{3mm}{0ex}}
\newcommand{\State}[5]
{\mbox{\vbox{state #1:\\
\parbox[t]{2em}{\verb+ + B:} \parbox[t]{\ExampleWidth}{#2}\\
\parbox[t]{2em}{\verb+ + S:} \parbox[t]{\ExampleWidth}{#3}\\
\parbox[t]{2em}{\verb+ + I:} \parbox[t]{\ExampleWidth}{#4}\\
\parbox[t]{2em}{\verb+ + P:} \parbox[t]{\ExampleWidth}{#5}
}}}
\newcommand{\mytable}[1]{

\vspace{1mm}
\centerline{ #1 }

}

\newcommand{\IndentABit}[1]{\rule{5em}{0ex}{#1}}
\newcommand{\WhatsUp}{{\mbox{{\tt\^{ }}}}}
\newcommand{\DD}{$\diamond$\ }

\newcommand{\?}{$ \backslash $}		  

\newlength{\DerivSep}                
\newlength{\DefaultLineThickness}    
\newcommand{\NLN}[1]{\LN[0pt]{\rule{0pt}{1ex}}{#1}}

\makeatletter

\newsavebox{\LNbox}
\newlength{\LNlength}

\def\LN{\@ifnextchar[{\LNz}{\LNz[\DefaultLineThickness]}}

\def\LNz[#1]#2#3{{%
\savebox{\LNbox}{%
\setlength{\arrayrulewidth}{#1}%
\begin{tabular}{@{\hspace{.1em}}c@{\hspace{.1em}}}{#2}\\[.2ex]%
\hline\raisebox{-.2ex}{#3}\end{tabular}}%
\settowidth{\LNlength}{\usebox{\LNbox}}%
\makebox[\LNlength]%
{\begin{minipage}[t]{\LNlength}\usebox{\LNbox}\end{minipage}}}}

\newsavebox{\UNb}
\newlength{\UNl}

\def\UN{\@ifnextchar[{\UNz}{\UNz[\DefaultLineThickness]}}

\def\UNz[#1]#2#3#4{{
\savebox{\UNb}{#3}\settowidth{\UNl}{\usebox{\UNb}}%
\begin{minipage}[t]{\UNl}{\begin{center}\makebox[\UNl]{\usebox{\UNb}}\\
\makebox[\UNl][l]{%
\raisebox{-1.0ex}[.1ex][-.2ex]{\rule[0.65ex]{\UNl}{#1}\hspace{.1em}{#4}}}\\
{#2}\end{center}}\end{minipage}}}

\newsavebox{\BNba}
\newsavebox{\BNbb}
\newlength{\BNla}
\newlength{\BNlb}
\newlength{\BNl}

\def\BN{\@ifnextchar[{\BNz}{\BNz[\DefaultLineThickness]}}

\def\BNz[#1]#2#3#4#5{{
\savebox{\BNba}{#3}\settowidth{\BNla}{\usebox{\BNba}}%
\savebox{\BNbb}{#4}\settowidth{\BNlb}{\usebox{\BNbb}}\setlength{\BNl}{\BNlb}%
\addtolength{\BNl}{\BNla}%
\addtolength{\BNl}{\DerivSep}%
\begin{minipage}[t]{\BNl}{\begin{center}\makebox[\BNl]{%
\usebox{\BNba}%
\hspace*{\DerivSep}%
\usebox{\BNbb}}\\
\makebox[\BNl][l]{%
\raisebox{-1.0ex}[.1ex][-.2ex]{\rule[0.65ex]{\BNl}{#1}\hspace{.1em}{#5}}}\\
{#2}\end{center}}\end{minipage}}}

\newsavebox{\TNba}
\newsavebox{\TNbb}
\newsavebox{\TNbc}
\newlength{\TNla}
\newlength{\TNlb}
\newlength{\TNlc}
\newlength{\TNl}

\def\TN{\@ifnextchar[{\TNz}{\TNz[\DefaultLineThickness]}}

\def\TNz[#1]#2#3#4#5#6{
\savebox{\TNba}{#3}\settowidth{\TNla}{\usebox{\TNba}}%
\savebox{\TNbb}{#4}\settowidth{\TNlb}{\usebox{\TNbb}}%
\savebox{\TNbc}{#5}\settowidth{\TNlc}{\usebox{\TNbc}}\setlength{\TNl}{\TNlc}%
\addtolength{\TNl}{\TNla}%
\addtolength{\TNl}{\TNlb}%
\addtolength{\TNl}{2\DerivSep}%
\begin{minipage}[t]{\TNl}{\begin{center}\makebox[\TNl]{%
\makebox[\TNla]{\usebox{\TNba}}%
\hspace*{\DerivSep}\makebox[\TNlb]{\usebox{\TNbb}}%
\hspace*{\DerivSep}\makebox[\TNlc]{\usebox{\TNbc}}}\\
\makebox[\TNl][l]{%
\raisebox{-1.0ex}[.1ex][-.2ex]{\rule[0.65ex]{\TNl}{#1}\hspace{.1em}{#6}}}\\
{#2}\end{center}}\end{minipage}}

\newsavebox{\PivotBox}
\newlength{\PivotLength}
\newcommand{\SetPivot}[1]{%
\savebox{\PivotBox}{#1}%
\settowidth{\PivotLength}{\usebox{\PivotBox}}}

\newcommand{\UsePivot}{\usebox{\PivotBox}}

\def\UNp{\@ifnextchar[{\UNpz}{\UNpz[\DefaultLineThickness]}}

\def\UNpz[#1]#2#3#4{{
\savebox{\UNb}{#3}\settowidth{\UNl}{\usebox{\UNb}}%
\begin{minipage}[t]{\UNl}{%
\begin{center}\makebox[\UNl]{\usebox{\UNb}}\\
\makebox[\UNl]{%
\rule{\PivotLength}{0pt}%
\rule{\DerivSep}{0pt}%
\addtolength{\UNl}{-\PivotLength}%
\addtolength{\UNl}{-\DerivSep}%
\makebox[\UNl]{%
\begin{minipage}{\UNl}%
\begin{center}%
\makebox[\UNl][l]{%
\raisebox{-1.3ex}[.1ex][-.2ex]{\rule[0.55ex]{\UNl}{#1}\hspace{.1em}{#4}}}\\
\addtolength{\UNl}{\PivotLength}%
\addtolength{\UNl}{\DerivSep}%
{#2}%
\end{center}%
\end{minipage}}}%
\end{center}}%
\end{minipage}}}

\def\BNp{\@ifnextchar[{\BNpz}{\BNpz[\DefaultLineThickness]}}

\def\BNpz[#1]#2#3#4#5{{
\savebox{\BNba}{#3}\settowidth{\BNla}{\usebox{\BNba}}%
\savebox{\BNbb}{#4}\settowidth{\BNlb}{\usebox{\BNbb}}\setlength{\BNl}{\BNlb}%
\addtolength{\BNl}{\BNla}%
\addtolength{\BNl}{\DerivSep}%
\begin{minipage}[t]{\BNl}{%
\begin{center}\makebox[\BNl]{%
\makebox[\BNla]{\usebox{\BNba}}\hspace*{\DerivSep}\makebox[\BNlb]{%
\usebox{\BNbb}}}\\
\makebox[\BNl]{%
\rule{\PivotLength}{0pt}%
\rule{\DerivSep}{0pt}%
\addtolength{\BNl}{-\PivotLength}%
\addtolength{\BNl}{-\DerivSep}%
\makebox[\BNl]{%
\begin{minipage}{\BNl}%
\begin{center}%
\makebox[\BNl][l]{%
\raisebox{-1.3ex}[.1ex][-.2ex]{\rule[0.55ex]{\BNl}{#1}\hspace{.1em}{#5}}}\\
\addtolength{\BNl}{\PivotLength}%
\addtolength{\BNl}{\DerivSep}%
{#2}%
\end{center}%
\end{minipage}}}%
\end{center}}%
\end{minipage}}}

\def\TNp{\@ifnextchar[{\TNpz}{\TNpz[\DefaultLineThickness]}}

\def\TNpz[#1]#2#3#4#5#6{{
\savebox{\TNba}{#3}\settowidth{\TNla}{\usebox{\TNba}}%
\savebox{\TNbb}{#4}\settowidth{\TNlb}{\usebox{\TNbb}}%
\savebox{\TNbc}{#5}\settowidth{\TNlc}{\usebox{\TNbc}}\setlength{\TNl}{\TNlc}%
\addtolength{\TNl}{\TNla}%
\addtolength{\TNl}{\TNlb}%
\addtolength{\TNl}{\DerivSep}%
\addtolength{\TNl}{\DerivSep}%
\begin{minipage}[t]{\TNl}{%
\begin{center}\makebox[\TNl]{%
\makebox[\TNla]{\usebox{\TNba}}\hspace*{\DerivSep}%
\makebox[\TNlb]{\usebox{\TNbb}}\hspace{\DerivSep}%
\makebox[\TNlc]{\usebox{\TNbc}}}\\
\makebox[\TNl]{%
\rule{\PivotLength}{0pt}%
\rule{\DerivSep}{0pt}%
\addtolength{\TNl}{-\PivotLength}%
\addtolength{\TNl}{-\DerivSep}%
\makebox[\TNl]{%
\begin{minipage}{\TNl}%
\begin{center}%
\makebox[\TNl][l]{\raisebox{-1.3ex}[.1ex][-.2ex]{\rule[0.55ex]{\TNl}{#1}%
\hspace{.1em}{#6}}}\\
\addtolength{\TNl}{\PivotLength}%
\addtolength{\TNl}{\DerivSep}%
{#2}%
\end{center}%
\end{minipage}}}%
\end{center}}%
\end{minipage}}}

\newcommand{\maxdim}[2]{\ifdim#1>#2 #1 \else #2 \fi} 

\makeatother

\newtheorem{example}{}    

\newlength{\ExampleWidth}
\newcommand{\startx} 
 {\begin{example}
  \rm
  \ \ \ 
  \begin{minipage}[t]{\ExampleWidth}}  

\newcommand{\startxl}[1] 
 {\begin{example}
  \label{#1}
  \ \ \ 
  \rm
  \begin{minipage}[t]{\ExampleWidth}}  

\newcommand{\stopx} 
 {\end{minipage}
  \end{example}}

\newcommand{\startxr}[1] 
 {(#1)
  \ \ \ 
  \begin{minipage}[t]{\ExampleWidth}}  

\newcommand{\stopxr} 
 {\par\end{minipage}}

\makeatletter
\def\@citex[#1]#2{\if@filesw\immediate\write\@auxout{\string\citation{#2}}\fi
  \def\@citea{}\@cite{\@for\@citeb:=#2\do
    {\@citea\def\@citea{;\penalty\@m\ }\@ifundefined
       {b@\@citeb}{{\bf ?}\@warning
       {Citation `\@citeb' on page \thepage \space undefined}}%
{\csname b@\@citeb\endcsname}}}{#1}}
\let\@internalcite\cite
\def\cite{\def\citename##1{##1, }\@internalcite}
\def\shortcite{\def\citename##1{}\@internalcite}
\def\newcite{\leavevmode\def\citename##1{{##1} (}\@internalciteb}

\def\@citexb[#1]#2{\if@filesw\immediate\write\@auxout{\string\citation{#2}}\fi
  \def\@citea{}\@newcite{\@for\@citeb:=#2\do
    {\@citea\def\@citea{;\penalty\@m\ }\@ifundefined
       {b@\@citeb}{{\bf ?}\@warning
       {Citation `\@citeb' on page \thepage \space undefined}}%
\hbox{\csname b@\@citeb\endcsname}}}{#1}}
\def\@internalciteb{\@ifnextchar [{\@tempswatrue\@citexb}%
{\@tempswafalse\@citexb[]}}

\def\@newcite#1#2{{#1\if@tempswa, #2\fi)}}

\def\@biblabel#1{\def\citename##1{##1}[#1]\hfill}

\def\@cite#1#2{({#1\if@tempswa , #2\fi})}

\def\thebibliography#1{\section*{Bibliography\@mkboth
 {Bibliography}{Bibliography}}\list
 {}{\setlength{\labelwidth}{0pt}\setlength{\leftmargin}{20pt}
 \setlength{\itemindent}{-20pt}}
 \def\newblock{\hskip .11em plus .33em minus -.07em}
 \sloppy\clubpenalty4000\widowpenalty4000
 \sfcode`\.=1000\relax}

\def\@lbibitem[#1]#2{\item[]\if@filesw 
      { \def\protect##1{\string ##1\space}\immediate
        \write\@auxout{\string\bibcite{#2}{#1}}\fi\ignorespaces}}

\def\@bibitem#1{\item\if@filesw \immediate\write\@auxout
       {\string\bibcite{#1}{\the\c@enumi}}\fi\ignorespaces}

\makeatother

\newlength{\RevealingThickness}
\newif\ifprediction                     
\predictiontrue

\begin{document}
\setlength{\parskip}{\medskipamount}
\setlength{\parindent}{0 in}

\title{A Computational Model of Syntactic Processing: \\
Ambiguity Resolution from Interpretation}
\author{Michael Niv}
\groupchair{Mark J. Steedman}
\tablespagefalse
\copyrighttrue
\beforepreface
\prefacesection{Acknowledgements}
My deepest feelings of gratitude and indebtedness are to my advisor and mentor
Mark Steedman.  Mark has taught me not just linguistics and cognition, but
also about thinking and behaving like a scientist.  He tirelessly and
carefully read through each of the many, many drafts of papers I have given
him, including this thesis, providing meticulous and insightful comments; and
spent countless hours explaining and debating these comments with me.  I would
guess that every single point I discuss in this thesis reflects his
contributions.

This thesis, like me, is a product of a community.  The Cognitive Science
community at Penn brings forth a beautifully dissonant buzz of intellectual
activity.  The diversity of outlooks, paradigms, methods, results and opinions
about the study of the mind provides an ideal environment in which to go
shopping for one's own direction.  It is impossible to identify the individual
members and visitors to IRCS (Institute for Research in Cognitive Science)
that have shaped my perspective, so I'll just thank the two people who are
responsible for enabling the tremendous growth of IRCS during my stay at Penn:
Aravind Joshi and Lila Gleitman.

I am very grateful to my doctoral committee: Janet Fodor, Aravind Joshi, Mitch
Marcus and Ellen Prince for their helpful comments on this document, in
particular to Ellen and Janet for long and fruitful discussions about the
first parts of the thesis, and to Mitch for making the Penn Treebank available
to me.

My colleagues Breck Baldwin, Barbara Di Eugenio, Bob Frank, Dan Hardt, Jamie
Henderson, Beth Ann Hockey, Young-Suk Lee, Owen Rambow, Phil Resnik, Robert
Rubinoff, Lyn Walker, and many other members of the Computational Linguistics
group at Penn provided me with much help, support and friendship throughout my
graduate studies.  IRCS postdocs have been particularly helpful: I have
learned a great deal from Sandeep Prasada, Jeff Siskind, and Mark Hepple.

This work has benefitted greatly from suggestions and advice by Ellen Bard,
Julie Boland, Lyn Frazier, Susan Garnsey, Mark Johnson, Robert Ladd, Don
Mitchell, Dick Oehrle, Stu Shieber, Val Tannen, Henry Thompson, Amy Weinberg,
Steve Whittaker, and Bill Woods.  Rich Pito was very helpful in extending his
treebank searching program {\tt tgrep} to accommodate my needs.

I am grateful to Bonnie Webber for offerring me my first opportunity to do
research --- the TraumAID project, and for her financial support during my
first few years at Penn.  The research reported in this document was supported
by the following grants: DARPA N00014-90-J-1863, ARO DAAL03-89-C-0031, NSF IRI
90-16592, Ben Franklin 91S.3078C-1.  I thank the Computational Linguistics
faculty members for their sustained financial support.  My thanks also go to
Jow and Oyi, Ali Baba, Yue Kee, and the Kims for sustenance.

I am very grateful to Barbara, Ramesh, Bob, Sandeep, and to the official
members of the late night Wawa crew: Patrick, Anuj, Tilman, especially to
Young-Suk, for the beginnings of life-long friendships.

I've had a few really wonderful teachers: Ruta in kindergarten, Mme. O'Connor
for French in high school, Joe O'Rourke in college, and Val Tannen in graduate
school.  They each possess the rare and precious talent of sparking curiosity
and intellectual excitement in their students.

My most important teachers have been my parents, Yaffa and Avigdor.  I thank
them and my two sisters Adi and Tamar for encouraging my curiosity (and coming
to terms with just how curious I have become).

\newpage
\addcontentsline{toc}{chapter}{Abstract}

\begin{center}
{\Large 
{\bf Abstract\\[1mm]}
A Computational Model of Syntactic Processing:\\[-.8mm]
Ambiguity Resolution from Interpretation\\[2mm]
\large
Michael Niv\\[1mm]
Mark J. Steedman (Supervisor)
}
\end{center}

Syntactic ambiguity abounds in natural language, yet humans have no difficulty
coping with it. In fact, the process of ambiguity resolution is almost always
unconscious.  But it is not infallible, however, as example 1 demonstrates.

1. The horse raced past the barn fell.
   
This sentence is perfectly grammatical, as is evident when it appears in the
following context:

\begin{tabbing}
2. \= Two horses were being shown off to a prospective buyer.  
One was raced past a meadow \\
   \> and the other was raced past a barn.
\end{tabbing}

Grammatical yet unprocessable sentences such as 1 are called `garden-path
sentences.' Their existence provides an opportunity to investigate the human
sentence processing mechanism by studying how and when it fails.  The aim of
this thesis is to construct a computational model of language understanding
which can predict processing difficulty.  The data to be modeled are known
examples of garden path and non-garden path sentences, and other results from
psycholinguistics.

It is widely believed that there are two distinct loci of computation in
sentence processing: syntactic parsing and semantic interpretation.  One
longstanding controversy is which of these two modules bears responsibility for
the immediate resolution of ambiguity.  My claim is that it is the latter, and
that the syntactic processing module is a very simple device which blindly and
faithfully constructs all possible analyses for the sentence up to the current
point of processing.  The interpretive module serves as a filter, occasionally
discarding certain of these analyses which it deems less appropriate for the
ongoing discourse than their competitors.

This document is divided into three parts.  The first is introductory, and
reviews a selection of proposals from the sentence processing literature.  The
second part explores a body of data which has been adduced in support of a
theory of structural preferences --- one that is inconsistent with the present
claim.  I show how the current proposal can be specified to account for the
available data, and moreover to predict where structural preference theories
will go wrong.  The third part is a theoretical investigation of how well the
proposed architecture can be realized using current conceptions of linguistic
competence.  In it, I present a parsing algorithm and a meaning-based ambiguity
resolution method.

\afterpreface
\setcounter{page}{1}

\chapter{Introduction}
\label{ch:intro}

The question I address here is how people deal with the linguistic ambiguity
which pervades natural language discourse.  I focus on syntactic ambiguity.
The task is to construct a detailed theory of the sentence processing
mechanism, its components, and the nature and dynamics of their interaction.

The data to be accounted for are measurements of processing difficulty (or lack
thereof) in various sentence types, both in and out of context.  Current
methods of measuring processing difficulty are often crude (e.g.\ naive
understandability judgements for a list of sentences) or, at best, indirect
(e.g.\ spatially diffuse EEG response patterns, and chronometric measurement of
cross-modal lexical priming, self-paced word-by-word reading, eye movement).
Nevertheless observations of processing difficulty very often show remarkable
and unmistakable regularity.  This regularity is the data to be
explained.\footnote{ In this work I do not address the strength of processing
difficulty effects, nor the issue of how humans cope with processing difficulty
(e.g.\ by rereading the offending passage). The aim is solely to account for
those linguistic environments which give rise to processing difficulty.}

Many models of human sentence processing have been put forth.  Most try to
account for processing difficulty by positing some properties of the {\em
parsing\/} component of the linguistic cognitive apparatus.

Frazier and Fodor (1978)\nocite{sausage} and Marcus (1980)\nocite{Marcus80} are
well known examples which attempt to derive a wide variety of phenomena from
memory limitations in the processor.

Theories have also been proposed in which the parser embodies a preference for
certain analyses over certain others.  Frazier and her colleagues have
advocated preferences for certain structural configurations.  Pritchett has
argued for preference for maximizing the degree to which the principles of
grammar are satisfied at every step of the parsing process.

Distinct from these parser-based theories of processing difficulty, is a theory
advocated by Crain, Steedman and Altmann, (CSA, hereinafter) which ascribes the
locus of ambiguity resolution preferences to higher-level interpretive
components, as opposed to the lower-level syntactic parsing component.  CSA
describe this architecture in broad terms, and apply it in detail to a fairly
narrow class of phenomena, essentially modifier attachment ambiguity.  In this
dissertation I argue for a conception of the syntactic processor which is a
generalization of CSA's proposal.  My claim is that the syntactic processor is
the simplest imaginable: all it represents is syntactic analyses of the input.
It is not responsible for resolution of ambiguity --- that task is performed by
the interpreter.

This document is divided into three parts.  The first is introductory, and
reviews a selection of proposals from the sentence processing literature, much
of which implicitly assume a specialized syntactic processor.  It concludes
with a detailed statement of the central claim of the dissertation.  The second
part, chapters 3 and 4, explores a body of data which has been adduced in
support of a theory of structural preferences --- one that is inconsistent with
the present claim.  In these chapters I show how the current proposal can be
specified to account for the available data, and moreover to predict where
structural preference theories will go wrong.  The third part, chapters 5 and
6, is a theoretical investigation of how well the proposed architecture can be
realized using current conceptions of linguistic competence.  Chapter~5
addresses issues of parsing --- it is an attempt to carry out Steedman's
(1994)\nocite{GandP} program of simplifying the theory of the parser by
adopting a competence grammar which defines more `incremental' analyses than
other grammars.  Chapter~6 is a synthesis of the parser developed in chapter~5
and the competence-base ambiguity resolution criteria developed in previous
chapters.  It describes an implemented computer model intended to demonstrate
the viability of the central claim.  Chapter~7 provides a conclusion and
suggests areas of further research.

\chapter{Previous Work}
\label{ch:survey}

In this chapter, I review a selected sample of the sentence processing
literature.  Of the many issues which any proposed model of human sentence
processing must address, I focus on two --- the role of memory limitations, and
the extent of `deliberation' which precedes ambiguity resolution.  The reader
is referred to Gibson (1991)\nocite{Gibson91} for a general review (and cogent
critique) of the literature.

\section{Memory Limitation}

Considering that the process of sentence understanding is successfully
implemented by the computational mechanism of the human brain, one may ask
about the nature of the architectural features of this computational device:
what is the relation among the various subcomponents --- lexical, syntactic,
and interpretive processes; and what sorts of limitations are imposed on
computational and memory resources by the finite `hardware' dedicated to the
task?  I begin with the latter question and focus on memory limitations.

\subsection{Representing an Analysis}

The most familiar demonstration that the processing system does not find all
grammatically possible analyses of a string with equal ease is the classic
example from Chomsky and Miller (1963)\nocite{ChomskyMiller63}:

\startxl{RatCat}
The rat that the cat that the dog bit chased died.
\stopx

Miller and Chomsky accounted for this in automaton theoretic terms --- the
processor cannot be interrupted while processing a constituent of type X to
process another constituent of type X.  More recent work, (Gibson 1991; Joshi
1990;\nocite{Joshi90} Rambow and Joshi 1993\nocite{RambowJoshi93}) consider a
variety of constructions in English and German which give rise to
center-embedding-like effects, and come to similar (though not identical)
conclusions: as it proceeds incrementally through the input string, the
underlying automaton is incapable of maintaining a large number of separate
pieces of the input which are not integrated together.  I return to this issue
in sections \ref{sec:DisConsequences} and
\ref{sec:UnambiguousStructures}.  Difficulty with sentences such as 
\ref{RatCat} arise
independently of syntactic ambiguity --- they indicate an inherent limitation
in the processor in representing the linguistic structure which they require.
The conclusion that this difficulty results from memory constraints in the
processor is unchallenged in the recent literature, as far as I know.

\subsection{Representing Competing Analyses}

The question of whether memory limitations are responsible for another form of
processing difficulty, namely so-called {\em garden path sentences,} as in
\ref{HorseRaced}, is much more controversial.

\startxl{HorseRaced}
The horse raced past the barn fell.
\stopx

With this sentence, there is no question that the processor is capable of
representing the necessary linguistic structure --- the grammatically identical
sentence in \ref{HorseRidden} causes no processing difficulty.

\startxl{HorseRidden}
The horse ridden past the barn collapsed.
\stopx

Authors such as Frazier and Fodor (1978)\nocite{sausage} and Marcus
(1980)\nocite{Marcus80} (see Mitchell, Corley and Garnham 1992\nocite{MCG};
Weinberg 1993\nocite{Weinberg93} for more recent incarnations of the two works,
respectively) have argued that when the processor encounters the (local,
temporary) ambiguity in the word `raced' in \ref{HorseRaced}, it is incapable
of keeping track of both available analyses of the input until the arrival of
the disambiguating information. That is, memory limitations {\em force\/} a
commitment.  Other authors (Crain and Steedman 1985\nocite{CrainSteedman85};
Altmann and Steedman 1988\nocite{AltmannSteedman88}; McClelland, St.\ John and
Taraban 1989\nocite{McClellandStJohnTaraban89}; Gibson 1991\nocite{Gibson91};
Pritchett 1992; Spivey-Knowlton, Trueswell and Tanenhaus
1993\nocite{SpiveyTrueswellTanenhaus93}, {\em inter alia}) have argued that the
processor considers all grammatically available analyses and picks among the
alternatives according to certain preferences (these authors differ widely
about what the preferences are).  I now consider a few of these papers in more
detail.

\subsubsection{The Sausage Machine}
\label{sec:SausageMachine}

Frazier and Fodor (1978)\nocite{sausage} proposed an architecture for the
syntactic processor whose central characteristic is a stage of processing whose
working memory is limited.  Their proposal is that the sentence processing
mechanism is comprised of modules:

\begin{quote}
The Preliminary Phrase Packager (PPP) is a `shortsighted' device, which peers
at the incoming sentence through a narrow window which subtends only a few
words at a time.  It is also insensitive in some respects to the
well-formedness rules of the language.  The Sentence Structure Supervisor (SSS)
can survey the whole phrase marker for the sentence as it is computed, and it
can keep track of dependencies between items that are widely separated in the
sentence and of long-term structural commitments which are acquired as the
analysis proceeds. (p. 292)
\end{quote}

Interesting predictions of processing difficulty arise for situations where the
PPP imposes the incorrect bracketing (or chunking) on a substring of the input.
Frazier and Fodor characterize the PPP has having a memory size of roughly six
words, and attempting, at any point to ``...group as may items as it can into a
single phrasal package.'' (p. 306) Aside from predicting difficulties with
center-embedded sentences (e.g.\ in \ref{RatCat} the PPP might try to chunk
``the rat that the cat'' into one package.) their account makes interesting
predictions with respect to modifier attachment.  Consider

\startxl{Lake}
We went to the lake to swim quickly.
\stopx

Their account predicts that the PPP will attempt to structure `quickly' with
the material immediately to its left, namely `to swim' rather than with `went'.
This prediction is not made when the adverbial consists of more words, e.g.\
\ref{TooCold}

\startxl{TooCold}
We went to the lake to swim but the weather was too cold.
\stopx

In \ref{TooCold}, the adverbial clause `but...' cannot fit into the PPP
together with `to swim' so the PPP puts the two constituents into separate
packages, and the SSS has the opportunity to decide how to attach the three
packages

\startxl{TooColdx}
[We went to the lake] [to swim] [but the weather was too cold.]
\stopx

The time-pressure under which the processor is operating --- faced with quickly
incoming words --- leads Frazier and Fodor to make another prediction about
attachment ambiguity resolution, namely, that syntactically `simplest' analyses
will be found first, thus preferred.  This was formalized by Frazier
(1978)\nocite{Frazier78}

\startxl{MAdef}
{\bf Minimal Attachment: } Attach incoming material into the phrase-marker
being constructed using the fewest nodes consistent with the well-formedness
rules of the language.
\stopx

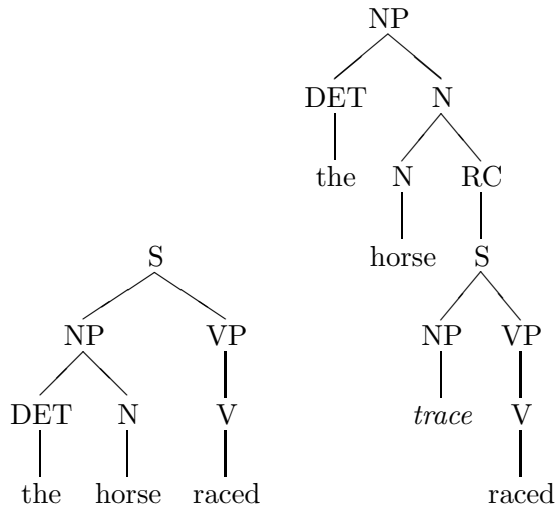
\begin{figure}
\centerline{\mbox{\mbox{
  \begin{picture}(93.22749999999999,114.0)(0,0)
   \put(54.94,101.0){\makebox(1,1){\rule[-1ex]{0ex}{3ex}{\normalsize S}}}
   \put(54.94,96.72){\line(-3,-2){26.913}}
   \put(54.94,96.72){\line(3,-2){26.912}}
   \put(28.027,71.0){\makebox(1,1){\rule[-1ex]{0ex}{3ex}{\normalsize NP}}}
   \put(28.027,66.72){\line(-1,-1){16.425}}
   \put(28.027,66.72){\line(1,-1){16.425}}
   \put(11.602,41.0){\makebox(1,1){\rule[-1ex]{0ex}{3ex}{\normalsize DET}}}
   \put(11.602,36.72){\line(0,-1){17.44}}
   \put(11.602,11.0){\makebox(1,1){\rule[-1ex]{0ex}{3ex}{\normalsize the}}}
   \put(44.452,41.0){\makebox(1,1){\rule[-1ex]{0ex}{3ex}{\normalsize N}}}
   \put(44.452,36.72){\line(0,-1){17.44}}
   \put(44.452,11.0){\makebox(1,1){\rule[-1ex]{0ex}{3ex}{\normalsize horse}}}
   \put(81.852,71.0){\makebox(1,1){\rule[-1ex]{0ex}{3ex}{\normalsize VP}}}
   \put(81.852,66.72){\line(0,-1){17.44}}
   \put(81.852,41.0){\makebox(1,1){\rule[-1ex]{0ex}{3ex}{\normalsize V}}}
   \put(81.852,36.72){\line(0,-1){17.44}}
   \put(81.852,11.0){\makebox(1,1){\rule[-1ex]{0ex}{3ex}{\normalsize raced}}}
   \put(0.0,0.0){\makebox(1,1)[l]{\rule[-1ex]{0ex}{3ex}{\normalsize}}}
  \end{picture}
}\ \ \ \mbox{
  \begin{picture}(93.27250000000001,204.0)(0,0)
   \put(31.747,191.0){\makebox(1,1){\rule[-1ex]{0ex}{3ex}{\normalsize NP}}}
   \put(31.747,186.72){\line(-6,-5){20.145}}
   \put(31.747,186.72){\line(6,-5){20.144}}
   \put(11.602,161.0){\makebox(1,1){\rule[-1ex]{0ex}{3ex}{\normalsize DET}}}
   \put(11.602,156.72){\line(0,-1){17.44}}
   \put(11.602,131.0){\makebox(1,1){\rule[-1ex]{0ex}{3ex}{\normalsize the}}}
   \put(51.891,161.0){\makebox(1,1){\rule[-1ex]{0ex}{3ex}{\normalsize N}}}
   \put(51.891,156.72){\line(-5,-6){14.946}}
   \put(51.891,156.72){\line(5,-6){14.947}}
   \put(36.945,131.0){\makebox(1,1){\rule[-1ex]{0ex}{3ex}{\normalsize N}}}
   \put(36.945,126.72){\line(0,-1){17.44}}
   \put(36.945,101.0){\makebox(1,1){\rule[-1ex]{0ex}{3ex}{\normalsize horse}}}
   \put(66.838,131.0){\makebox(1,1){\rule[-1ex]{0ex}{3ex}{\normalsize RC}}}
   \put(66.838,126.72){\line(0,-1){17.44}}
   \put(66.838,101.0){\makebox(1,1){\rule[-1ex]{0ex}{3ex}{\normalsize S}}}
   \put(66.838,96.72){\line(-5,-6){15.061}}
   \put(66.838,96.72){\line(5,-6){15.06}}
   \put(51.777,71.0){\makebox(1,1){\rule[-1ex]{0ex}{3ex}{\normalsize NP}}}
   \put(51.777,66.72){\line(0,-1){17.44}}
   \put(51.777,41.0){\makebox(1,1){\rule[-1ex]{0ex}{3ex}{\normalsize 
{\em{trace}}}}}
   \put(81.898,71.0){\makebox(1,1){\rule[-1ex]{0ex}{3ex}{\normalsize VP}}}
   \put(81.898,66.72){\line(0,-1){17.44}}
   \put(81.898,41.0){\makebox(1,1){\rule[-1ex]{0ex}{3ex}{\normalsize V}}}
   \put(81.898,36.72){\line(0,-1){17.44}}
   \put(81.898,11.0){\makebox(1,1){\rule[-1ex]{0ex}{3ex}{\normalsize raced}}}
   \put(0.0,0.0){\makebox(1,1)[l]{\rule[-1ex]{0ex}{3ex}{\normalsize}}}
  \end{picture}
}}}
\caption{Main-verb and reduced-relative-clause analyses of 
\protect{\ref{HorseRaced}}}
\label{TwoAnalyses}
\end{figure}

Minimal Attachment predicts that the main-verb analysis of `raced' in
\ref{HorseRaced} will be initially pursued, as can be seen by the relative
syntactic complexity of the main verb and reduced-relative analyses in
\reffig{TwoAnalyses}

Minimal Attachment similarly predicts that the sentences in \ref{MAegs} each
give rise to a garden path.

\startxl{MAegs}
a. The cop shot the spy with the binoculars.\\ b. The doctor told the patient
that he was having trouble with to leave.
\stopx

In more recent work, Frazier and her colleagues (Rayner, Carlson and Frazier
1983\nocite{RCF83}; Frazier 1990\nocite{Frazier90}) propose a different
modularization of the language processing faculty: the {\em syntactic
processor\/} constructs a single analysis of the incoming words according to
structurally defined criteria such as Minimal Attachment above.  The {\em
thematic processor\/} considers the phrasal constituents that the syntactic
processor has found and considers in parallel, all the possible thematic
combinations of these constituents.  When it finds a better thematic
combination than the one being constructed by the syntactic processor, it
interrupts the syntactic processor, telling it to reanalyze the sentence.

\subsubsection{PARSIFAL}
\label{sec:parsifal}

Marcus (1980)\nocite{Marcus80} seeks to reconcile the apparent speed and
efficiency of the human sentence processing mechanism with traditional parsing
techniques (for ambiguous grammars) which are significantly slower.  Standard
parsing algorithm require time which is either polynomial or exponential in the
length of the string, but humans do not require words to arrive more slowly as
the input string --- the sentence --- becomes longer.  Marcus concludes that
the processor must be able to make all parsing decisions in a bounded amount of
time (i.e.\ using a bounded number processing steps).  He proposes an automaton
model, which he calls {\em Parsifal.} This model is a production system which
has a data store and set of pattern-action rules.  To achieve a bound on the
amount of time required by the processor to make its move, Marcus bounds the
portion of the processor's memory which is `visible' to the rules.  The store
has two components: a parse stack, and a buffer of three cells, each capable of
storing one constituents.  The rules may only mention the syntactic category of
the content of each of the cells, and (roughly) the top of the parse stack.
The processor proceeds {\em deterministically\/} in the sense that any
structure it builds (by attaching constituents from the buffer into the stack)
may not be destroyed.  When the processor reaches an ambiguity, it may either
resolve it, or it may leave one or more constituents uncombined in the buffer,
provided there is room.  If there is no room in the buffer for new
constituents, the processor is {\em forced\/} to make a commitment, which may
result in a garden path.

An account of garden paths which is based strictly on the 3-cell memory limit
quickly runs into empirical difficulties.  Pritchett (1988)\nocite{Pritchett88}
provides the following examples which can be resolved within a 3-cell buffer,
but nevertheless appear to be garden paths.  (see Gibson (1991) for a detailed
critique of Marcus's parser)

\startxl{AgainstParsifal}
a. The boat floated quickly sank.\\
b. Without her money would be hard to find.\\
c. While Tegan sang a song played on the radio.
\stopx

\subsubsection{Minimal Commitment}

Marcus, Hindle and Fleck (1983)\nocite{MarcusHindleFleck83} propose an
architecture which maintains the 3-cell buffer of Marcus's earlier work, but
factors the procedural pattern-action rules into a more elegant collection of
structural description rules and an engine which applies them.  While the rules
of grammar are about {\em direct dominance\/} of nodes in the phrase marker,
the processor maintains partial specifications by means of {\em dominance\/}
statements, and other devices.  Preserving the determinism in Marcus's parser,
their processor may not retract any assertions about the phrase marker that it
is constructing.  Weinberg (1993)\nocite{Weinberg93} adopts Marcus \etal's
proposal of partial descriptions of phrase structure,\footnote{Weinberg's
(partial) structural description include statements of dominance, direct
dominance, linear precedence, and partial category specification using
features.} but jettisons altogether the idea of a bounded buffer.  Instead,
Weinberg adopts an arguably less stipulative account of garden path sentences:

\startxl{PQI}
{\bf Principle of Quick Interpretation:}
The parser attaches arguments using the smallest number of dominance
statements and features necessary to assign grammatically relevant
properties.
\stopx

This account predicts a garden path whenever the commitments necessitated by
the Principle of Quick Interpretation turns out to be inconsistent with
subsequent material.  No garden path is predicted in cases where the commitment
(i.e.\ partial description) constructed by the Principle of Quick
Interpretation is consistent with the rest of the string.

For an illustration of Weinberg's parser, consider

\startxl{Eric}
a. I knew Eric.\\
b. I knew Eric was a great guy.
\stopx

Weinberg's account entails that neither \ref{Eric}a nor b is a garden path.
This follows from the description that the processor builds after encountering
the prefix `I knew Eric':

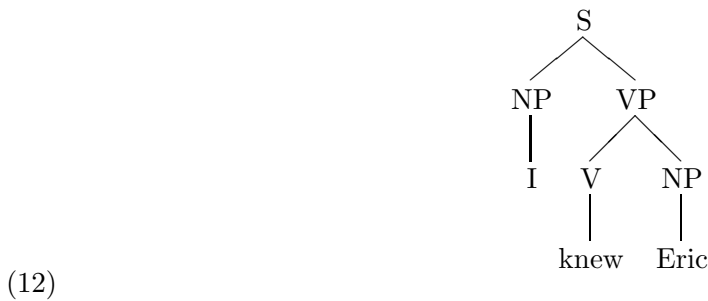
\startxl{EricTree}
\centerline{
  \begin{picture}(75.255,114.0)(0,0)
   \put(27.538,101.0){\makebox(1,1){\rule[-1ex]{0ex}{3ex}{\normalsize S}}}
   \put(27.538,96.72){\line(-6,-5){19.803}}
   \put(27.538,96.72){\line(6,-5){19.803}}
   \put(7.735,71.0){\makebox(1,1){\rule[-1ex]{0ex}{3ex}{\normalsize NP}}}
   \put(7.735,66.72){\line(0,-1){17.44}}
   \put(7.735,41.0){\makebox(1,1){\rule[-1ex]{0ex}{3ex}{\normalsize I}}}
   \put(47.341,71.0){\makebox(1,1){\rule[-1ex]{0ex}{3ex}{\normalsize VP}}}
   \put(47.341,66.72){\line(-1,-1){17.221}}
   \put(47.341,66.72){\line(1,-1){17.221}}
   \put(30.12,41.0){\makebox(1,1){\rule[-1ex]{0ex}{3ex}{\normalsize V}}}
   \put(30.12,36.72){\line(0,-1){17.44}}
   \put(30.12,11.0){\makebox(1,1){\rule[-1ex]{0ex}{3ex}{\normalsize knew}}}
   \put(64.562,41.0){\makebox(1,1){\rule[-1ex]{0ex}{3ex}{\normalsize NP}}}
   \put(64.562,36.72){\line(0,-1){17.44}}
   \put(64.562,11.0){\makebox(1,1){\rule[-1ex]{0ex}{3ex}{\normalsize Eric}}}
   \put(0.0,0.0){\makebox(1,1)[l]{\rule[-1ex]{0ex}{3ex}{\normalsize}}}
  \end{picture}
}
\stopx

(where the links are express dominance, not direct dominance). \ref{EricTree}
is compatible with either the direct dominance interpretation of \ref{EricTree}
or with the analysis necessary for \ref{Eric}b, where an S node intervenes
between the VP node and the [$_{np}$~Eric] node.  \ref{Eric} is in contrast
with

\startxl{MaryFood}
After Mary ate the food disappeared from the table.
\stopx

When the processor encounters `the food', the Principle Of Quick Interpretation
commits it to the fact that the VP headed by `ate' dominates the NP `the food'.
This commitment is inconsistent with the rest of the string, so a garden path
is correctly predicted.

Weinberg's proposal is that the sentence processor's working memory is limited
to hold exactly one structural representation.  Unlike Frazier and Fodor's and
Marcus's proposals, this limitation is confined to the representation of
ambiguity --- Weinberg's memory limitation makes no predictions about
difficulty with center embedding.

The three proposals above all share the fundamental property that the processor
pursues only one analysis at a time.  This has been called `serial' processing
as well as `determinism'.  Standing in contrast to serial processing are
proposals that the processor constructs representations for the various
ambiguous analyses available at any point.\footnote{One could argue that
Frazier \etal's model is a mix of serial (syntactic) and parallel (thematic)
processing, but what is relevant here is the question of whether the {\em
initial syntactic analysis\/} is carried out in serial or parallel.} Of the
many `parallel' proposals in the literature, I shall review only two: Gibson's
(1991)\nocite{Gibson91} proposal of processing load and breakdown; and the
parallel weak-interaction model of Crain, Steedman and Altmann (CSA) (Crain and
Steedman 1985\nocite{CrainSteedman85}; Altmann and Steedman
1988\nocite{AltmannSteedman88}).

\subsubsection{Gibson (1991)}

Gibson (1991)\nocite{Gibson91} proposes that the human sentence processing
mechanism pursues all grammatically available analyses in parallel as it
processes the string, discarding those analyses which are `too costly' --- that
is, when the cost of one analysis, A, exceeds that of another analysis, B, by
more than P Processing Load Units, A is discarded, necessitating conscious
effort to reconstruct should it be subsequently necessary.  The cost of an
analysis is the sum of Processing Loads which it incurs by virtue of having
certain memory-consuming properties.  Within Gibson's model, a theory of
sentence processing consists in a precisely defined collection of
memory-consuming properties and a numeric cost associated with each.
Considering a variety of data (mostly introspective judgements of processing
difficulty sentences) Gibson proposes a collection of four memory-consuming
properties: three have to do with failures to identify the relations among the
various constituents in the string (cf.\ Chomsky's (1986)\nocite{Chomsky86}
Principle of Full Interpretation); the fourth property associates a cost with
the need to access a constituent which is not the most recent.  Gibson
concentrates on syntactic properties, which he considers the most tractable to
investigate. He acknowledges that a complete theory of sentence processing
would likely require augmenting his set of properties with ``lexical, semantic,
pragmatic and discourse-level properties which are associated with significant
processing loads.''

\subsubsection{Crain, Steedman and Altmann}

Crain and Steedman (1985)\nocite{CrainSteedman85} and Altmann and Steedman
(1988)\nocite{AltmannSteedman88} report a collection of experiments which
militate against a model in which the syntactic processor operates in a serial
or deterministic fashion.  Consider the local ambiguity in \ref{DocPat}a,
illustrated in
\ref{DocPat}b and c.

\startxl{DocPat}
\begin{tabbing}
a. The psychologist told the wife that...\\
b. The psychologist told the wife that he was having trouble with her 
	husband.\\
c. \=The psychologist told the wife that he was having trouble with 
	to leave her\\
   \>husband.
\end{tabbing}
\stopx

A model where the syntactic processor operates serially would predict that the
ambiguity would be resolved on some structural grounds (e.g.\ Minimal
Attachment\footnote{CSA address their arguments specifically against Minimal
Attachment, but it applies to other any structural preference strategies such
as those in the proposals of Weinberg, above, Pritchett
(1992)\nocite{Pritchett92} and others.}) presumably toward the complementizer
analysis of `that', as in \ref{DocPat}b, not the relativizer analysis in
\ref{DocPat}c.   This resolution would occur independently of {\em meaning\/}
 of the constituents in question.  But Crain and Steedman found that depending
on compatibility with the {\em discourse context\/} the processor can be made
to select either analysis.  When there were two wives in the discourse context,
\ref{DocPat}b was a garden path --- reflecting a commitment toward a further
restrictor on the set of candidate referents.  When there was one wife in the
discourse, \ref{DocPat}c was a garden path.  This basic finding was replicated
using a different ambiguous structure and methodologies by Altmann and Steedman
(1988); Sedivy and Spivey-Knowlton 1993\nocite{SedivySpivey93}; and Altmann,
Garnham and Dennis (1992)\nocite{AltmannGarnhamDennis92}.  Given the
sensitivity to the meaning of the various alternatives, CSA argue that the
processor must be explicitly weighing the sensibleness of the alternatives.  It
follows that the interpreter receives representations, in parallel from the
syntactic processor of all available syntactic analyses.

Neither Gibson nor CSA discuss explicit bounds on the number of analyses that
are maintained by the processor at any time.  This just means that unlike
Marcus's proposal and the Sausage machine, it is only the preference criteria
themselves, not the memory bounds that bear the explanatory role for ambiguity
resolution behavior.  It must be emphasized that neither parallel model above
requires that the processor be able to represent the potentially exponentially
proliferating set of ambiguous analyses for a multiply ambiguous string ---
whenever the processor's preference reaches some threshold, it discards the
less-preferred analyses, thus keeping the size of analysis-set manageable.
Indeed, most ambiguities are resolved very quickly, making the processor appear
as if it operates serially. There is additional experimental evidence in
support of a parallel model of the sentence processor.

\subsubsection{Gorrell (1989)}

Gorrell (1989)\nocite{Gorrell89} used a lexical decision task\footnote{Where in
the middle of reading a sentence, the subject is presented with a word and has
to quickly respond with whether it is a word of the English language.  It has
been argued (Wright and Garrett 1984)\nocite{WrightGarrett84} that this task is
facilitated if the target word `fits in' at the point in the sentence that the
subject is processing.} to show that both analysis of a temporarily ambiguous
sentence are maintained --- that is, the ultimately dispreferred analysis
exerts an effect of lexical decision facilitation.  With sentences such as
\ref{GorellEgs}, Gorrell used target words (is, has, must) which are consistent
with the dispreferred (complex) analysis, and found facilitation in both the
Ambiguous and Complex conditions, but not for the unambiguous Simplex
condition. Presentation of the sentences were interrupted at the points marked
with \DD for presentation of the target word.

\startxl{GorellEgs}
\begin{tabbing}
NP/S Ambiguity\\
X \= Ambiguous: \= \kill
\> Simplex:   \> It's obvious that Holmes saved the son of the banker 
\DD right away\\
\> Ambiguous: \> It's obvious that Holmes suspected the son of the \\
\>            \> banker \DD (right away/was guilty)\\
\> Complex:   \> It's obvious that Holmes realized the son of the banker 
\DD was guilty\\[4mm]
Main Verb/Participle ambiguity\\
\> Simplex:   \> The company was loaned money at low rates 
\DD to ensure high volume\\
\> Ambiguous: \> The company loaned money at low rates 
\DD (to ensure high volume/\\
\>            \> decided to begin expanding)\\
\> Complex:   \> The company they loaned money at low rates 
\DD decided to begin\\
\>            \> expanding
\end{tabbing}
\stopx

\subsubsection{Hickok, Pickering and Nicol}

Additional experiments by Hickok (1993)\nocite{Hickok93} and Nicol and
Pickering (1993)\nocite{NicolPickering93} confirm Gorrell's findings.  Working
independently, these researchers considered the local ambiguity used by CSA in
\ref{DocPat}.  Using the method of antecedent reactivation\footnote{Where at
the position of a WH trace, the lexical decision times for words which are
semantically related to the antecedent of the trace are facilitated.  (Swinney
\etal\ 1988)\nocite{SwinneyEtal88} (See Fodor 1989\nocite{Fodor89} for a
review.)} they found that the relative clause analysis, which is strongly
dispreferred to the complement clause analysis, is still `active' and causes
reactivation of the WH trace at the position marked with \DD.

\startxl{HickokEg}
The girl swore to the dentist that a group of angry people called \DD the
office about the incident.
\stopx

Hickok used visual computer-paced presentation of the sentence, while Nicol and
Pickering used cross-modal priming --- the sentences were presented auditorily
and the target word was presented visually.  Results from the two experiments
consistently show reactivation of the WH-antecedent.  This result is quite
surprising given the remarkable extent to which subjects are garden pathed when
faced with a string such as \ref{HickokEg2}.  It suggests that dispreferred
analyses are not discarded outright --- they just fade away.

\startxl{HickokEg2}
The girl swore to the dentist that a group of angry people called that she was
going to quit.
\stopx

\subsubsection{MacDonald, Just and Carpenter (1992)}

MacDonald, Just and Carpenter (1992)\nocite{MacDonaldJustCarpenter92} argue
that how quickly dispreferred analyses fade away is subject to individual
variations in short term memory.  MacDonald \etal\ rated their subjects on
their performance on the Reading Span Task --- a task in which the subject
reads a list of unrelated sentences, keeping track of the the last word in each
sentence.  At the end of the list, the subject must recall the final words.
Subjects vary substantially on the length of the list for which they can
perform the task accurately. Score on this task is positively correlated with a
variety of language performance scores including SAT verbal score.  The theory
that MacDonald \etal\ propose is that high-span subjects maintain ambiguities
for longer periods of time.  This theory makes the interesting and
counter-intuitive prediction that for locally ambiguous sentences which are
disambiguated consistently with the preferred analysis, high-span readers would
have to work harder than low-span readers, since they would also be maintaining
the doomed non-preferred analysis.  This is indeed what they found. They
compared the locally ambiguous sentence in
\ref{Midnight}a to an unambiguous control, \ref{Midnight}b, and 
to the non-garden-pathing main-verb analysis in \ref{Midnight}c.

\startxl{Midnight}
\begin{tabbing}
a. The experienced soldiers warned about the dangers conducted the midnight 
raid.\\
b. \=The experienced soldiers who were told about the dangers conducted the\\
   \>midnight raid.\\
c. The experienced soldiers warned about the dangers before the midnight raid.
\end{tabbing}
\stopx

They found that high-span readers could cope better with the ambiguity in
\ref{Midnight}a:  On a reading comprehension task, high-span readers performed 
better than low-span readers (63--64\% correct versus 52--56\% correct ---
almost at chance --- on true/false questions) This confirms the relevance of
the reading span task to some aspects of reading ability.  More interestingly,
MacDonald \etal\ found that for the main-verb sentences, as in \ref{Midnight}c,
high span readers took significantly {\em more\/} time to read the last word of
the sentence.  For high span readers there was a very slight\footnote{This
effect reached statistical significance only when data from many experiments
(with slightly different conditions) were pooled together.} elevation in the
reading time of the ambiguous region `warned about the dangers' in the
ambiguous sentences
\ref{Midnight}a and c, as compared to the locally unambiguous \ref{Midnight}b.
This is clear evidence of the additional burden which maintaining the
possibility of a reduced-relative analysis imposes on high-span readers.
Slight though this effect is, it does constitute an online measure of the cost
of maintaining multiple analyses in parallel.

\subsubsection{Summary}

The existence of garden path sentences leads to the inescapable conclusion that
not all syntactic analyses are maintained indefinitely.  The stronger
conclusion, that multiple syntactic analyses are {\em never\/} retained from
word to word is inconsistent with three sorts of psycholinguistic evidence:
\begin{enumerate}
\item The meaning of the various competing analyses are compared, hence
computed, requiring the identification of syntactic relations.  (Note that
proposals such as the thematic processor of Frazier and her colleagues do not
specify how the interpretive module can identify which of the many possible
relations among constituents are potentially allowed by grammatical analyses of
the string which the processor has not chosen.  Different languages impose
different restrictions on which constituents may be combined, so syntactic
analysis must precede interpretation.)
\item The `discarded' reading still manifests certain signs of life on 
sufficiently sensitive
tests, such as the lexical decision task.
\item For readers who show signs of coping better with ambiguity, the 
dispreferred reading exacts a measurable processing cost.
\end{enumerate}

\section{Deliberation before Ambiguity Resolution}

Aside from memory limitations assumed by a model, another dimension along which
the various proposals vary is the nature and amount of computation that
precedes ambiguity resolution.  The two logically extreme positions have each
been advocated --- that any processing whatsoever, including arbitrarily
complex inference can precede ambiguity resolution, and that ambiguity is not
even identified by the processor online, let alone deliberated on.  Some papers
advocate intermediate positions.  In this section, I present a few papers
arranged in approximately increasing order of amount of pre-resolution
deliberation.

\subsection{Shieber and Pereira}
\label{sec:ShieberPereira}

Shieber (1983)\nocite{Shieber83} and Pereira
(1985)\nocite{Pereira85}\footnote{written at roughly the same time} propose a
technique for constructing a deterministic automaton given a (potentially
nondeterministic) grammar.  The automaton's memory consists of a stack of
symbols (grammatical categories) and a register which stores the name of one of
a bounded number of states which the automaton is in.  It is equipped with a
pre-compiled action table which completely determines what move it should take
next (add/remove items from the stack, change the state it is in) based on the
current state, the next word in the input string, and the top-of-stack symbol.
This action table is constructed from a grammar using a well-known grammar
compilation technique (LR parsing, Aho and Johnson
1974\nocite{AhoJohnson74}). If the grammar is locally ambiguous, the
compilation technique results in certain entries in the action table containing
sets of actions, each corresponding to a different analysis.  Shieber and
Pereira show how structural preference strategies such as Minimal Attachment
\ref{MAdef} and Lexical Preference (see below) can be used to resolve such
indeterminacies in the action table {\em at compile time.} The resulting
deterministic automaton will therefore follow the path of action consistent
with the minimally attached reading and not even detect the possibility of
another analysis.

\subsection{Syntactic `Optimality'}

A variety of proposals (Frazier and Fodor 1978; Rayner, Carlson, and Frazier
1983; Weinberg 1993; Pritchett 1992, {\em inter alia}) posit structural
preference criteria.  None of these proposals concretely specify the algorithm
by which the processor finds the preferred parse.  Presumably this involves
some sort of search over the space of analyses possible for the input so far.
For example, Frazier's Minimal Attachment principle could be made to fall out
of a processor which tries to integrate the next word into the current phrase
marker by trying all combinations in parallel and stopping as soon as it has
found the first grammatical solution.  In none of these proposals does any
non-syntactic information enter into the process of determining the first-pass
analysis.

\subsection{Lexical Association}

Ford, Bresnan and Kaplan (1982)\nocite{FBK82} argue that aside from purely
structural ambiguity resolution criteria, the processor is also sensitive to
the `strength' of association between certain words like verbs and the nouns
they take as arguments.  They conducted a questionnaire experiment in which
they presented participants with an ambiguous sentence such as
\ref{WantedDress}, and asked them to identify which reading they got first.

\startxl{WantedDress}
The woman wanted the dress on the rack.
\stopx

They found that by changing the main verb, they could significantly alter the
ambiguity resolution preferences observed.  For example, \ref{WantedDress} was
resolved 90\% of the time with the PP modifying `dress' and 10\% of the time
modifying `wanted'; however when `wanted' was replaced with `positioned', the
preferences reversed from 90~vs.~10 to 30~vs.~70.  Ford \etal\ incorporate such
preferences into a serial processing algorithm --- their processor considers
the set of possible rules at any point, applying both lexical preference and
general structurally-state rules to decide which rule to apply next.

\subsection{Explicit Consideration of Syntactic Choices}

The models of Marcus (1980) and Gibson (1991) explicitly reason about the
various syntactic alternatives available at any point.  Marcus's system
contains rules for {\em differential diagnosis\/} of local structural
ambiguity.  These rules consider the current collection of constituents and
decide how to combine them.  Gibson's system explicitly constructs all
grammatically available structures and applies preference metrics to adjudicate
among them.  While both systems adhere to solely syntactic criteria for
ambiguity resolution, their authors acknowledge the need for certain
meaning-based preferences in more complete/realistic versions of their work
(Gibson 1991 chapter~9, esp.\ p.~186; Marcus 1980 chapter~10).

\subsection{The Weakly Interactive Model}
\label{sec:WeaklyInteractive}

CSA argue that the syntactic processor constructs all grammatically available
analyses and the interpreter evaluates these analyses according to
meaning-based criteria.  While the criterion they propose,
\ref{PrincipleOfParsimony} requires potentially very elaborate inferences to
apply, their actual experiments rely on relatively easy to compute aspects of
meaning.

\startxl{PrincipleOfParsimony}
{\em Principle of Parsimony:} (Crain and Steedman 1985)\\ If there is a reading
that carries fewer unsatisfied but consistent presuppositions or entailments
than any other, then, other criteria of plausibility being equal, that reading
will be adopted as most plausible by the hearer, and the presuppositions in
question will be incorporated in his or her [mental] model [of the discourse].
\stopx

In their experiments, Crain and Steedman presented a locally ambiguous sentence
such as
\ref{CrainSteedmanBis} in two different contexts, as exemplified in 
\ref{CrainSteedmanContexts}.

\startxl{CrainSteedmanBis}
The psychologist told the wife that he was having trouble with to leave her
husband.
\stopx

\startxl{CrainSteedmanContexts}
a. One couple context:\\
A psychologist was counseling a married couple.  
One member of the pair was fighting with him
but the other was nice to him.\\[1ex]
b. Two  couple context:\\
A psychologist was counseling two married couples.  
One of the couple was fighting with him
but the other was nice to him.
\stopx

The inference which their subjects evidently were computing were first, going
{}from a married couple (or two) to a part of the couple, namely a wife;
second, determining whether the definite expression `the wife' referred
uniquely, presumably by determining whether the cardinality of the set of wives
was greater than one.  In another experiment, Crain and Steedman (1985) found
effects of plausibility in how often subjects garden pathed on examples such as

\startxl{CrainTwo}
a. The teachers taught by the Berlitz method passed the test.\\ 
b. The children taught by the Berlitz method passed the test.
\stopx

This is evidence that subjects use online the knowledge that teachers typically
{\em teach\/} and children typically {\em are taught.} Again, one may argue
that this sort of knowledge could conceivably be fairly directly represented
and is very quick to access (see Resnik 1993\nocite{Resnik93}).  Plausibility
effects on the reduced-relative/main-verb ambiguity in
\ref{CrainTwo} have since been found by many researchers (Pearlmutter and 
MacDonald 1992\nocite{PearlmutterMacDonald92}; Trueswell, Tanenhaus and Garnsey
1992\nocite{TTG92} {\em inter alia}).  Trueswell and Tanenhaus
(1991)\nocite{TrueswellTanenhaus91} have found that subjects are sensitive to
the temporal coherence of the discourse when parsing reduced relative clauses.
For example `The student caught cheating...' is more likely to be interpreted
as a reduced relative when the discourse is in the future tense than when it is
in the past tense.

Marslen-Wilson and Young (cited in Marslen-Wilson and Tyler,
1987\nocite{Marslen-WilsonTyler87}) conducted an experiment which shows
immediate effects of a rather complex inference process.  They placed ambiguous
phrases such as `flying planes' and `visiting relatives' in contexts which
inferentially favor one of their two readings.

\startxl{MarslenWilsonYoung}
a. If you want a cheap holiday, visiting relatives...\\
b. If you have a spare bedroom, visiting relatives...
\stopx

Subjects listened to an audio tape of these materials, and, at end of the
fragment, they were presented with a written word.  Their task was to read the
word outloud --- the so-called cross-modal naming task.  The words of interests
were `is' and `are', consistent with the
\ref{MarslenWilsonYoung}a and b meanings, respectively.  Marslen-Wilson and 
Young found significant effects of plausibility on subjects' reaction times,
indicating that the relatively complex inference required is brought to bear on
the {\em immediately following\/} word.  It is not clear just how much
inference is brought to bear on a word-by-word basis, this is due in part, no
doubt, to our current inability to objectively assess the complexity of
inference.

\subsection{Discussion}

There is a substantial and growing body of evidence in support of the claim
that the human sentence processing mechanism consults a variety of information
sources before it resolves ambiguities:

\begin{itemize}
\item properties of particular lexical items such as their preferred 
subcategorization frames (Ford, Bresnan Kaplan 1982; Garnsey, Lotocky and
McConkie 1992\nocite{GLM92}; Trueswell, Tanenhaus and Kello 1993\nocite{TTK93};
Juliano and Tanenhaus 1993\nocite{JulianoTanenhaus93})\footnote{For information
such as verb subcategorization frame preferences, it is very hard to tease
apart whether the information is associated with the lexical entry for the
verb, or with the `deeper' representation of the concept (e.g.\ of the verb)
and how it is associated to other concepts (e.g.\ its arguments) to which it is
being related by the sentence.  Current research on practical applications of
natural language technology, in trying to avoid the complexity of knowledge
representation, has been quite successful in assuming rich relation among
words.  Collecting lexical cooccurrence statistics from large text corpora,
researchers (e.g.\ Hindle and Rooth 1993\nocite{HindleRooth93}) are able to
construct ambiguity resolution algorithms which perform significantly better
than ones based hand-coded domain knowledge.  In fact, it is surprising to see
just how far cooccurrence-based statistical approaches to approximating natural
language can go --- Church (1988)\nocite{Church88} presents an algorithm for
determining the form-class of words in text.  This algorithm is trained on
hand-tagged text; it performs no syntactic analysis of its input, it only keeps
track of the form-class frequency for each word, and the frequency of
consecutive form-class tags in text.  Using this remarkably impoverished
approximation of the linguistic phenomena of English, Church's algorithm was
able to achieve form-class determination performance of better than 90\%.  The
success of these algorithms can serve as a demonstration of how easy it is to
`cheat' by attributing complex behavior using association-based strength of
representations of surface observable objects such as words.}
\item semantic properties associated more or less directly with the words in 
the sentence (e.g.\ married couple $\longrightarrow$ wife, teachers teach, from
Crain and Steedman 1985; cheap vacation $\longrightarrow$ visiting relatives,
{}from Marslen-Wilson and Young)
\item fit of the linguistic expression into the current discourse
(e.g.\ definite reference --- CSA; coherence of tense --- Trueswell and
Tanenhaus 1991\nocite{TrueswellTanenhaus91})
\end{itemize}

There is not, however, a consensus that the language processing architecture is
indeed parallel, and highly deliberative.  Mitchell, Corley and Garnham
(1992)\nocite{MCG} argue that there are separate syntactic and thematic
processors (see \refsec{sec:SausageMachine}); while the thematic processor does
consider the meanings of the various combinations of the words in the string so
far, the syntactic processor pursues only one analysis.  The thematic processor
can come to suspect that the syntactic processor may be pursuing the wrong
analysis and alert it very quickly to change course.  This quick alert
strategy, which Mitchell
\etal\ refer to as {\em stitch in time\/} can sometimes trick the processing 
system into a garden path.  The consequence of this argument is that if one
trains one's psycholinguistic measurement apparatus on the exact point in the
process, one could catch the syntactic processor constructing the minimally
attached analysis only to have this analysis abandoned in favor of the
contextually appropriate analysis a few hundred milliseconds later.  This issue
is currently being debated, with researchers on both sides refining their
experimental techniques.  (see Altmann, Garnham and Dennis
1992\nocite{AltmannGarnhamDennis92}).

\section{The Central Claim}

The claim that I argue in this dissertation is that the parsing mechanism a
straightforward device which blindly and faithfully applies the knowledge of
language (syntactic competence) directly to its input, allowing the
interpretation module to impose its preferences in case of ambiguity.  At each
point in processing, the parser constructs all available syntactic analyses for
the input thus far.  The interpreter considers the set of available analyses
and what each would mean, and selects a subset to discard.  The parser deletes
these analyses and extends the remaining ones with the next incoming word,
repeating the process until it is either exhausted the input string, or it is
stuck --- none of the non-discarded analyses has a grammatical continuation in
the next input word.

The following are immediate consequences of this claim:
\begin{enumerate}
\item There are no structural preferences (e.g.\ Minimal Attachment) 
encoded or implemented by 
the parser.
\item All ambiguity resolution decisions among grammatically licensed analyses 
stem directly from the linguistic competence: (in the broadest sense of
 the term)
\begin{itemize}
\item plausibility of the message carried by the analysis
\item quality of fit of this message into the current discourse
\item felicity of the constructions used in the utterance to express the
 message
\item the relative frequency of use of a certain construction or 
lexical item\footnote{On the
assumption that the knowledge of language specifies quantitative 
`frequency' information,
e.g.\ subcategorization frame preference.}
\end{itemize}  
That is, when resolving ambiguity, the hearer answers the question 
``which of these
grammatically possible analyses is the one that the speaker is most
 likely trying to
communicate to me?''
\item Each of the four criteria in 2 above can be investigated
independently of syntactic 
ambiguity.
\item The parser uses a direct representation of the competence grammar,
as opposed to some specially processed encoding intended solely for the task 
of parsing.
\item Certain parsing effects which have been heretofore explained by memory 
bounds in
the parser have explanations elsewhere:
\begin{itemize}
\item Parsing does not always proceed serially or deterministically.
\item Buffer-limitation-based predictions of how long ambiguity can be
maintained and when it must be resolved (e.g.\ Marcus's 3-cell buffer, 
the Sausage Machine's
6 word window) will predict either too long an ambiguity-maintenance 
period (in case
disambiguating information is available early) or too short a period, 
(in case disambiguating
information is not available.)
\item True memory-load effects, which would arise in artificial situations 
where many locally ambiguous readings are available for the input string but no
disambiguating information is applicable, will result from a diffuse shortage
in attentional resources needed to keep track of the many analyses in parallel,
in analogy with an overloaded multi-user computer which exhibits gradual
performance degradation.
\end{itemize}
\end{enumerate}

\chapter{Accounting for Recency Phenomena}
\label{ch:informational}

\newcommand{\sbar}{$\overline{\mbox{S}}$}

In the previous chapter I reviewed evidence that the ambiguity resolution
process is sensitive to a variety of aspects of `sensibleness' of the competing
analyses: real-world plausibility, felicity of definite reference, and temporal
coherence.  In this chapter and the next, I consider a collection of
ambiguities which seem, at first glance, to be resolved by criteria other than
sensibleness.  I will argue that when the notion of sensibleness is broadened
to encompass the degree of fit to current discourse situation, these
ambiguities receive a straight-forward sensibleness-based account.

\section{Right Association}

Kimball (1973)\nocite{Kimball73} proposes the parsing strategy of Right
Association (RA).  RA resolves modifiers attachment ambiguities by attaching at
the lowest syntactically permissible position along the right frontier of the
phrase marker.  Many authors (among them
\nocite{Wilks85} \nocite{Schubert86}\nocite{Whittemore90}\nocite{Weischedel91} 
Wilks 1985, Schubert 1986, Whittemore \etal\ 1990, and Weischedel \etal\ 1991)
incorporate RA into their parsing systems, yet none rely on it solely,
integrating it instead with ambiguity resolution preferences derived from
word/constituent/concept co-occurrence based criteria.  On its own, RA performs
rather well, given its simplicity, but it is far from adequate: Whittemore
\etal\ evaluate RA's performance on PP attachment using a corpus derived from
computer-mediated dialog.  They find that RA makes correct predictions 55\% of
the time.  Weischedel \etal, using a corpus of news stories, report a 75\%
success rate on the general case of attachment using a strategy Closest
Attachment which is essentially RA.  In the works just cited, RA plays a
relatively minor role, as compared with co-occurrence based preferences.

The status of RA is very puzzling.  Consider:

\startxl{jsttb}
a. John said that Bill left yesterday.\\
b. John said that Bill will leave yesterday.
\stopx

\startxl{China}
In China, however, there isn't likely to be any silver lining {\em because the
economy remains guided primarily by the state.} \\(from the Penn Treebank
corpus of Wall Street Journal articles)
\stopx

On the one hand, many naive informants do not see the ambiguity of \ref{jsttb}a
and are often confused by the semantically unambiguous \ref{jsttb}b --- a
strong RA effect.  On the other hand \ref{China} violates RA with impunity.
What is it that makes RA operate so strongly in
\ref{jsttb} but disappear in \ref{China}?  In the rest of this chapter, I 
argue that the high attachment of the adverbial encodes a commitment about the
information structure of the sentence which is infelicitous with the
information carried in
\ref{jsttb} but not with that in \ref{China}.  This commitment is about the
{\em volume of information\/} encoded in various constituents in the sentence,
and the feature which encodes this commitment is word (constituent) order.

\section{Information Volume}

Quirk \etal\ (1985)\nocite{Quirk85} define {\em end weight\/} as the tendency
to place material with more information content after material with less
information content.  This notion is closely related with {\em end focus\/}
which is stated in terms of importance of the contribution of the constituent,
(not merely the quantity of lexical material.)  These two principles operate in
an additive fashion.  Quirk \etal\ use them to account for a variety of
phenomena, among them:

\begin{tabbing}
genitive NPs:  \\
XXX \= * \= \kill
    \>   \> the shock of his resignation\\
    \> * \> his resignation's shock\\
it-extraposition: \\
    \>   \> It bothered me that she left quickly.\\
    \> ? \> That she left quickly bothered me.
\end{tabbing}

\begin{table*}
\begin{center}
\begin{tabular}{l@{\hspace{1mm}}l@{\hspace{1mm}}l@{\hspace{1mm}}l}
 & John sold it today.                            
& *& John sold today it.\\
 & John sold the newspapers today.                
& ?& John sold today the newspapers.\\
 & John sold his rusty socket-wrench set today.   
&  & John sold today his rusty socket-wrench set. \\
?& John sold his collection of 45RPM Elvis        
&  & John sold today his collection of 45RPM \\
 & \ \ records today.                             
&  &  \ \ Elvis records.\\
?& John sold his collection of old newspapers\hspace{-3mm}&  
						 
& John sold today his collection of old news- \\
 & \ \ from before the Civil War today.           
&  & \ \ papers from before the Civil War.
\end{tabular}
\end{center}
\caption{Illustration of heaviness and word order}
\label{elvis}
\end{table*}

Information volume clearly plays a role in modifier attachment, as shown in
table~\ref{elvis}.  My claim is that what is wrong with sentences such as
\ref{jsttb} is the violation, in the high attachment, of the principle of end
weight.  While violations of the principle of end weight in unambiguous
sentences (e.g.\ those in table~\ref{elvis}) cause little grief, as they are
easily accommodated by the hearer/reader, the online decision process of
ambiguity resolution could well be much more sensitive to small differences in
the degree of violation.  In particular, it would seem that in \ref{jsttb}b,
the weight-based preference for low attachment has a chance to influence the
parser before the temporal inference based preference for high attachment.

I am aware of no work which attempts to systematically tease apart the notion
of amount of linguistic material (measured in words or morphemes) from the
notion of amount of information communicated (in the pragmatic sense).  In this
document I use the term {\em information volume\/} to refer to a vague
combination of these two notions, on the assumption that they are highly
correlated in actual speech and text. To further simplify and operationalize
the definition of information volume, I classify single word constituents and
simple NPs as low information volume and constituents which include a clause as
high information volume.  In
\refsec{VolumeAndSensibleness} I argue that a very significant determinant of 
information volume is the pragmatic information carried by the constituent, not
by length of its surface realization.

\section{A study}

The consequence of my claim is that low information volume adverbials cannot be
placed after high volume arguments, while high volume adverbials are not
subject to such a constraint.  When the speaker wishes to convey the
information in \ref{jsttb}a (high attachment), there are other word-orders
available, namely,

\startxl{permu}
a. Yesterday John said that Bill left.\\ 
b. John said yesterday that Bill left.
\stopx

If the claim is correct then when a single word adverbial modifies a VP which
contains a high information volume argument, the adverbial will tend to appear
either before the VP or between the verb and the argument.  High volume
adverbials should be immune from this pressure.

To verify this prediction, I conducted an investigation of the Penn Treebank
corpus of about 1~million words of syntactically annotated text from the Wall
Street Journal.  Unfortunately, the corpus does not currently distinguish
between arguments and adjuncts --- they are both annotated as daughters of VP.
Since at this time, I do not have a dictionary-based method for distinguishing
\mbox{(VP asked (\sbar\ when...))} from \mbox{(VP left (\sbar\ when...))}, my
search cannot include all adverbials, only those which could never (or rarely)
serve as arguments.  I therefore restricted my search to subgroups of the
adverbials.

\begin{enumerate}
\item \sbar s whose complementizers participate overwhelmingly in adjuncts:
{\em after although as because before besides but by despite even lest
meanwhile once provided should since so though unless until upon whereas
while.}
\item single word adverbials:  {\em now however then already here too
  recently instead often later once yet previously especially
  again earlier soon ever first indeed sharply largely usually
  together quickly closely directly alone sometimes yesterday}
\end{enumerate}

The particular words were chosen solely on the basis of frequency in the
corpus, without `peeking' at their word-order behavior\footnote{Each adverbial
can appear in at least one position before the argument to the verb (sentence
initial, preverb, between verb and argument) and at least one
post-verbal-argument position (end of VP, end of S).}.

For arguments, I only considered NPs and \sbar s with complementizer {\em
that,} and the zero complementizer.

The results of this investigation appear the following table:

\newcommand{\zuzksata}{\hspace{1mm}}
\newcommand{\zuzktsatb}{\hspace{2mm}}
\newcommand{\zuzksatc}{\hspace{1mm}}
\newcommand{\zuzktsatd}{\hspace{2mm}}

\vspace{1ex}

\centerline{
\begin{tabular}{|l||r|r||r|r|}
\hline 
adverbial:  & \multicolumn{2}{c||}{single word} & \multicolumn{2}{|c|}{clausal}
\\ \hline
arg type    & pre-arg           & post-arg         & pre-arg            
 & post-arg \\ \hline
low volume  & { 760 \zuzksata}  & {399 \zuzktsatb} & {  13  \zuzksatc}    
&  {  597 \zuzktsatd}\\
high volume & { 267 \zuzksata}  & {  5 \zuzktsatb} & {   7  \zuzksatc}   
&  {   45
\zuzktsatd}\\
\hline 
total      &  {1027 \zuzksata}  &   {404 \zuzktsatb} & {  20  \zuzksatc}  
 &  {  642 \zuzktsatd}\\
\hline
\end{tabular}}

\vspace{1ex}

Of 1431 occurrences of single word adverbials, 404 (28.2\%) appear after the
argument.  If we consider only cases where the verb takes a high volume
argument (defined as one which contains an S), of the 273 occurrences, only 5
(1.8\%) appear after the argument.  This interaction with the information
volume of the argument is statistically significant ($\chi^2 = 115.5, p<.001$).

Clausal adverbials tend to be placed after the verbal argument: only 20 out of
the 662 occurrences of clausal adverbials appear at a position before the
argument of the verb.  Even when the argument is high in information volume,
clausal adverbials appear on the right: 45 out of a total of 52 clausal
adverbials (86.5\%).

\ref{China} and \ref{Kravis} are two examples of RA-violating sentences
which I have found.
\startxl{Kravis}
According to department policy, prosecutors must make a strong showing
that lawyers' fees came from assets tainted by illegal profits {\em before
any attempts at seizure are made.}

\stopx

To summarize: low information volume adverbials tend to appear before a high
volume argument and high information volume adverbials tend to appear after it.
The prediction is thus confirmed.

RA is at a loss to explain this sensitivity to information volume.  Even a
revision of RA, such as the one proposed by Schubert (1986) which is sensitive
to the size of the modifier and of the modified constituent, would still
require additional stipulation to explain the apparent conspiracy between a
{\em parsing\/} strategy and tendencies in {\em generator\/} to produce
sentences with the word-order properties observed above.  This also applies to
Frazier and Fodor's (1978)\nocite{sausage} Sausage Machine model which accounts
for RA effects using a narrow window in the parser (see
\refsec{sec:SausageMachine}).

\section{A Potential Practical Application}
\newcommand{\subvp}{$_{\mbox{vp}}$\ \ }
\newcommand{\subs}{$_{\mbox{s}}$\ \ }
\newcommand{\subx}{$_{\mbox{x}}$\ \ }
\newcommand{\suby}{$_{\mbox{y}}$\ \ }

How can we exploit the findings above in our design of practical parsers?
Clearly RA seems to work extremely well for single word adverbials, but how
about clausal adverbials?  To investigate this, I conducted another search of
the corpus, this time considering only ambiguous attachment sites.  I found all
structures matching the following two low-attached schemata\footnote{By * I
mean match 0 or more daughters.  By [\subx ... [\suby ]] I mean constituent x
contains constituent y as a rightmost descendant.  By [\subx ... [\suby ] ... ]
I mean constituent x contains constituent y as a descendant. }

\begin{tabbing}
\ \ low VP attached: \= [\subvp ... [\subs * [\subvp * adv *] * ] ...] \\
\ \ low S attached:  \> [\subvp ... [\subs * adv *] ...] \\
\end{tabbing}
\vspace{-4mm}
\noindent and the following two high-attached schemata
\vspace{-2mm}
\begin{tabbing}
\ \ high VP attached:  \=  [\subvp v * [... [\subs ]] adv *] \\
\ \ high S attached:   \>  [\subs * [... [\subvp ... [\subs ]]] adv * ]
\end{tabbing}

\noindent The results are summarized in the following table:

\newcommand{\zuz}{\hspace{5mm}}
\newcommand{\zuzi}{\hspace{3mm}}

\centerline{
\begin{tabular}{|l|r|r|r|}
\hline
adverb-type  & low-attached  &  high-att. &  \% high. \\ \hline
single word  &   {1116 \zuz} &   { 10 \zuzi}   &  { 0.8\%  \zuz} \\
clausal      &   { 817 \zuz} &   {194 \zuzi}   &  {19.2\%  \zuz} \\
\hline
\end{tabular}}

As expected, with single-word adverbials, RA is almost always right, failing
only 0.8\% of the time.\footnote{There is an interesting putative
counterexample to the generalization that only low information volume
adverbials give rise to recency effects, shown in (i). (I am grateful to Bill
Woods for bringing this example to my attention)

\noindent(i)  The Smiths saw the Grand Canyon flying to California.

\noindent Here there is a remarkably strong tendency to take the participial 
phrase `flying to California' as (belonging to) an argument to `saw'.  The more
plausible reading treats the participle as modifying the matrix subject, or the
matrix predication. I claim that this effect is not residual RA, but rather, it
stems from a subtle pragmatic infelicity in the `plausible' construal of the
participle.  My intuition is that when a participial adverbial is felicitously
used, the relation between the adverbial and the matrix predication is not
merely cotemporaneity but rather, the adverbial must be a {\em relevant\/} to
matrix predication.  An informal survey of post-head participles that do not
appear in construction with their heads (e.g.\ `spent the weekend writing a
paper') reveals that they most often appear delimited by a comma, and are
`relevantly' related to their heads, serving such rhetorical purposes as
evidence, consequence, elaboration, and exception.  I did not find examples of
mere cotemporaneity or scene-setting.  In fact, for scene-setting functions,
one tends to add the word `while' to the participle.  So the subtle infelicity
in (ii) can remedied as in (iii) or in (iv).

\noindent(ii) John collapsed flying to California.

\noindent(iii) John collapsed while flying to California.

\noindent(iv) John collapsed trying to run his third marathon in as many days.

\noindent In (i), the matrix attachment of the participial makes only the 
infelicitous scene-setting/cotemporaneity relation available, so the system is
forced to the ECM analysis which for all it can determine online, could have a
felicitous ending, or a slow to compute metaphorical interpretation.} However,
with clausal adverbials, RA is incorrect almost one out of five times.

\section{Information Volume and Sensibleness}
\label{VolumeAndSensibleness}
Let us return to the question of whether the attachment preferences discussed
above are indeed consistent with the thesis of sensibleness-based ambiguity
resolution.  If it turns out that information volume is simply a measure of
surface complexity (words, morphemes, phrase marker tree depth, etc.) then
there is no role for interpretation and sensibleness to play --- it follows
that the competence grammar marks information volume as a feature on certain
nodes and assigns a graded penalty of some kind to certain sequences of
volume-markings.  While the idea of graded penalty against certain structural
configurations is not new (cf.\ subjacency) the requirement for a
\mbox{$\pm$High-Volume} feature is rather odd.  Still, there is nothing in this
view which is inconsistent with the main thesis.

The other possible domain over which to define information volume is pragmatics
--- whatever Grice's (1975)\nocite{Grice75} maxims of quantity and manner are
about, that is, informativeness of the contribution and brevity/prolixity,
respectively.  If this is the case, then constituents are not marked by the
syntactic processor with their information volume. All that the syntax
determines is constituent order.  This constituent order can encode the
commitment that constituent X must carry less information than constituent Y.
The actual information volume is determined by the interpreter.  Such
determinations may be inconsistent with the order-based commitments, in which
case the analysis is deemed less sensible.

I would like to suggest that the pragmatic sense is a strong, if not an
exclusive\footnote{Ford, Bresnan and Kaplan (1982)\nocite{FBK82} point out that
RA effects are sensitive to the syntactic category of the more recent
attachment site.  They contrast (i) with (ii).

\noindent(i) Martha notified us that Joe died by express mail.

\noindent(ii) Martha notified us of Joe's death by express mail.

It is quite clear that the absurd RA reading is more prominent in (i) than in
(ii).  This is rather surprising because on informational terms, I can see no
notion of information by which `that Joe died' bears any more information than
`of Joe's death'.} determinant of information volume.  Here is one example:

The acceptability of verb-particle constructions clearly has to do with
information volume:

\startxl{Smashing}
\begin{tabbing}
a. * \=Joe called the friend who had crashed into his new car up.\\
b.   \>Joe called up the friend who had crashed into his new car.\\
c.   \>Joe called his friend up.\\
d.   \>Joe called up his friend.
\end{tabbing}
\stopx

It has been widely noted that pronouns are very awkward in post-verb-particle
positions:

\startxl{PissOff}
\begin{tabbing}
a. * \=This pissed off him.\\
b.   \>This pissed off Bob.
\end{tabbing}
\stopx

The reason, I claim, for the relative acceptability of \ref{PissOff}b is the
accommodation, by the hearer of (the possibility of) a context which places new
information in the NP Bob, e.g.

\startxl{PissOffDammit}
Mary passed John and Bob in the corridor without even saying hello.\\ 
Surprisingly, this only pissed off Bob.  John didn't seem to mind.
\stopx

\ref{PissOff}a can be made acceptable if the pronoun `him' is replaced by a
deictic accompanies by physical pointing --- that is, increasing the amount of
information associated with the word.

Returning to the central example of this chapter, let us consider the dialog in
\ref{Sabbatical}. When appropriately intoned, this dialog shows that a 
constituent like `that Bill will leave', which is construed as bearing high
information volume when it appears out of context in \ref{jsttb}b can indeed
bear low information volume when it expresses a proposition or concept which is
already {\em given in the discourse}.\footnote{I am grateful to Ellen Prince
for this example.}

\startxl{Sabbatical}
\begin{tabbing}
A: \= John said that Bill will leave next week, and that\\ 
   \> Mary will go on sabbatical
in September.\\
B: \> Oh really?  When did he announce all this? \\
A: \> He said that Bill will leave yesterday, and he told\\
   \> us about Mary's sabbatical this morning.
\end{tabbing}
\stopx

\section{Conclusion}

I have argued that the apparent variability in the applicability of Right
Association can be explained if we consider the information volume of the
constituents involved.  I have demonstrated that in at least one written genre,
low information volume adverbials are rarely produced after high volume
arguments --- precisely the configuration which causes the strongest RA-type
effects.  Considering the significant influence of pragmatic content on the
degree of information volume, the interaction between information volume and
constituent order provides a sensibleness-based account for the resolution of a
class of modifier attachment ambiguities.

\chapter{Other Constructions}
\label{ch:leftovers} 

Two often-discussed structural ambiguities have not been mentioned so far:
\startxl{HeardJoke}
John has heard the joke is offensive.
\stopx
\startxl{cannibals}
When the cannibals ate the missionaries drank.
\stopx

I will refer to the ambiguity in \ref{HeardJoke} as {\em NP vs.\ S complement,}
and to the garden-path effect in \ref{cannibals} as the {\em Late Closure
Effect\/} --- the term which Frazier and Fodor (1978)\nocite{sausage}
introduced.\footnote{\ref{HeardJoke} and \ref{cannibals} are intuitively garden
paths. One might argue that given the strong bias for jokes being heard and
cannibals eating missionaries, the structures in \ref{HeardJoke} and
\ref{cannibals} is irrelevant. But it is equally plausible that someone hears
some fact, and that cannibals engage in an (intransitive) eating activity, so
the question remains of why these strings are resolved as they are.}

In this chapter I consider the psycholinguistic evidence available about these
ambiguities, and consider two different ways of accounting for these and other
data.  The first proposal, which I call {\em disconnectedness theory\/} is a
formalization of many accounts of processing difficulty that appear in the
literature.  The second, which I call {\em Avoid New Subjects\/} has not been
proposed before in relation to ambiguity resolution.  I then consider the
evidence available for distinguishing these two accounts, ultimately trying to
show that disconnectedness theory makes some incorrect predictions.

\section{Available Evidence}
\label{sec:AvailableEvidence}

\subsection{NP vs.\ S complement}
\label{sec:NPvsS}

Advocates of structural ambiguity resolution strategies have argued that the
ambiguity in \ref{FR82m} is initially resolved by Minimal Attachment.

\startxl{FR82m}
Tom heard the latest gossip about the new neighbors was false.
\stopx

Frazier and Rayner (1982)\nocite{FrazierRayner82} used eye tracking to find
that for sentences such as \ref{FR82m} people slowed down when reading the
disambiguation region {\em was false.} Holmes, Kennedy and Murray
(1987)\nocite{HKM87} used a subject-paced word-by-word cumulative display
experiment to show that the slowdown which Frazier and Rayner observed persists
even when the ambiguity is removed, by the introduction of an overt
complementizer.  With experimental materials such as \ref{HKM87m}

\startxl{HKM87m} 
(TR) The maid disclosed the safe's location within the house to the officer.\\
(TC) The maid disclosed that the safe's location within the house had been
changed.\\ (RC) The maid disclosed the safe's location within the house had
been changed.
\stopx 

they found that in the disambiguation region (either {\em to the officer,} or
{\em had been changed\/}) the transitive verb sentence (TR) was read
substantially faster than the other two sentences.  The that-complement (TC)
sentence was read slightly faster than the reduced complement (RC) sentence.

In response, Rayner and Frazier (1987)\nocite{RaynerFrazier87} ran an an
eye-movement experiment which contradicted the conclusions of Holmes \etal\
(1987).  Using materials which were similar (but not identical) to those of
Holmes \etal\ (1987), they found that at the disambiguation region, TC was read
the fastest, followed by TR, followed by the ambiguous RC, consistent with the
theory of Minimal Attachment.

In turn, Kennedy, Murray, Jennings, and Reid (1989)\nocite{KMJR89} argued that
Rayner and Frazier (1987)\nocite{RaynerFrazier87} introduced serious artifacts
into their eye-tracking data by presenting their material on multiple lines and
not controlling for the resulting right-to-left eye-movement.  Kennedy
\etal\ also criticized other technical aspects of Rayner and Frazier's
experiment.  Kennedy \etal\ ran an eye-tracking study using the materials from
Holmes \etal\ (1987).  They found that TC and RC sentences were read
significantly slower in the disambiguation region than TR sentences.  They
found no reliable difference between TC and RC.  In a further experiment to
test the effect of line-breaks, they found statistically significant effects
whose nature was rather difficult to interpret.  They took this as evidence
that line-breaks do indeed introduce artifacts.

In summary, there is evidence that S-complement sentences --- TC and RC above
--- take longer to comprehend than comparable NP-complement sentences.

Another debate is whether RC sentences take longer to read than TC sentences,
and under what conditions.  Quite a few researchers have investigated the
question of whether RC sentences cause a garden-path effect when the matrix
verb `prefers' an NP rather than a clausal complement.

Kennedy \etal\ (1989) partitioned the materials for their first experiment
according to the bias of the matrix verb --- NP versus clausal.  They found no
effects of verb-bias on first-pass reading time, but found a statistically
non-significant effect of verb-bias on eye regressions initiated from the
disambiguating zone (i.e.\ indications of backtracking/confusion).  For both
kinds of verbs there were more regressions in the RC condition than the TC
condition, but the difference was greater for NP-bias verbs.  However Kennedy
\etal\ did not demonstrate that they accurately identified verb biases.

Many groups of researchers report experiments specifically designed to
investigate verb-bias effects on the extent of garden-path in RC sentences.  I
report the work of four: Holmes, Stowe and Cupples (1989)\nocite{HSC89};
Ferreira and Henderson (1990)\nocite{FH90}; Garnsey, Lotocky and McConkie
(1992); and Trueswell, Tanenhaus and Kello (1993)\nocite{TTK93}.  I refer to
them as HSC, FH, GLM, and TTK, respectively.  HSC, GLM, and TTK ran two-phase
experiments.  In the first phase subjects were asked to complete sentences such
as `He suspected...'.  Statistics from these data were used to assess verbs'
biases.  Two groups of verbs were selected: NP-preference and S-preference.
HSC's criterion for counting a verb as having a bias was a 15\% or greater
imbalance in subjects' responses.  When assessing a verb's argument structure,
they lumped TC and RC responses into one category.  GLM and TTK kept separate
tallies of these two kinds of complements.  This difference is significant
because the ambiguity in question is really between a TR and an RC analysis.
FH did not use a questionnaire, instead verbs were selected ``either on the
basis of normative data collected by Connine, Ferreira, Jones, Clifton and
Frazier (1984)\nocite{CFJCF84} or according to the intuitions of the
experimenters.''  The study of Connine \etal\ asked subjects to write a
sentence for each of a group of verbs.  They did not specify that the verb must
immediately follow an agent-subject, and this might have certain effects on the
way their data can be interpreted for the purpose at hand.

HSC considered the effects of two factors upon the degree of the garden
path-effect: verb-bias and plausibility of the post-verb NP as a direct object.
Their materials were of the form

\startxl{HSCm}
\begin{tabbing}
NP-bias verb\\
\ \= TC \ \ \= Implausible: \= \kill
\> TC \> Plausible:   \> The reporter saw that her friend was not succeeding.\\
\>    \> Implausible: \> The reporter saw that her method was not succeeding.\\
\> RC \> Plausible:   \> The reporter saw her friend was not succeeding.\\
\>    \> Implausible: \> The reporter saw her method was not succeeding.\\[3mm]
Clausal-bias verb\\
\> TC \> Implaus.: \= \kill
\> TC \> Plaus.:   \> The candidate doubted that his sincerity would be 
appreciated.\\
\>    \> Implaus.: \> The candidate doubted that his champagne would be 
appreciated.\\
\> RC \> Plaus.:   \> The candidate doubted his sincerity would be 
appreciated.\\
\>    \> Implaus.: \> The candidate doubted his champagne would be 
appreciated.
\end{tabbing}
\stopx

They tested the efficacy of the plausibility manipulation by asking subjects to
rate sentences such as `the reporter saw her method.'  This is an inadequate
test of the online plausibility of the NP analysis: Just because a subject
rejects the string as a sentence, it does not mean that the subject would,
online, reject the NP analysis for `her method' --- doing so would commit the
subject (depending on one's theory of grammar) to rejecting strings such as
`The reporter saw her method fail miserably when interviewing athletes.'  or
`The doctor found the fever discouraging.'  This criticism applies to the
majority of their `implausible' experimental materials, though unevenly for the
NP and S-bias verbs.  I therefore omit their findings with respect to this
factor.

They conducted three self-paced experiments varying the method of presentation
of the materials.  Their first experiment used a self-paced word-by word
cumulative display.  After each word the subject had to decide whether the
string is grammatical so far.  This resulted in remarkably slow reading times
--- three times slower than in eye-movement experiments.  The RC condition was
slower at the disambiguation region than the TC condition, this difference was
enhanced in the NP-bias condition, consistent with their theory that lexical
information is incorporated into the parsing process.  Advocates of Minimal
Attachment often argue that slow presentation methods may not be sensitive to
first-pass processing, and that it is not at all surprising that lexical
information is incorporated at the later stage tapped by this sort of
experiment.  Addressing this criticism, HSC ran another word-by-word self-paced
experiment, but this time subjects were required to repeat the sentence when it
was finished.  This resulted in somewhat faster reading times, still, roughly
twice as slow as in eye movement.  The findings in the second experiment were
comparable to those of the first, although the garden-path was detected roughly
one word later, and the differences in reading time were not quite as large.

One problem with cumulative displays is that subjects may be employ a strategy
of pressing the self-paced button faster than they are actually reading the
words.  In a third experiment, HSC used a non-cumulative display, where the
letters of each word in the sentence except the one being read were replaced
with underscores.  Instead of manipulating the plausibility of the ambiguous
NP, they manipulated its length.  Two examples are:

\startxl{HSC3m}
The lawyer heard (that) the story (about the accident) was not really true.\\
The reporter saw (that) the woman (who had arrived) was not very calm.
\stopx

For NP-bias verbs, at the first word of disambiguating region (`was' in
\ref{HSCm} and \ref{HSC3m}) the RC condition took 60 ms longer than the 
TC (530 ms versus 470 ms per word).  The difference between RC and TC was a
slightly larger for short NPs but there was no statistically significant
interaction between NP length and overtness of complementizer.

For clausal-bias verbs, RC sentences with long ambiguous NPs had a reading time
of roughly 520 ms for the disambiguating word, whereas the three other
conditions (i.e.\ the two TC conditions and the short NP RC condition) required
roughly 470ms.  For short NPs, the difference between TC and RC was not
significant, whereas for long NPs it was (the magnitudes were approximately 5
ms and 47 ms, respectively.)

In summary, the third experiment confirmed the first two by showing more
processing difficulty for the NP-bias verbs than the complement-bias verbs.  In
addition, it showed that when the ambiguous NP is long, readers tend to
interpret it as an argument to the matrix verb even for complement-bias verbs.

Ferreira and Henderson (1990)\nocite{FH90} attempted to dispute the claim that
lexical bias is incorporated into the processor's first pass
ambiguity-resolution decisions.  They conducted three experiments using three
different experimental procedures: eye-movement, non-cumulative self-paced
reading, and cumulative self-paced reading.  They used the same materials for
all three experiments.  One example is:

\startxl{FHm}
\begin{tabbing}
\ \ \= TC \ \ \= \kill
NP-bias\\
 \> TC \> Bill wrote that Jill arrived safely today.\\
 \> RC \> Bill wrote Jill arrived safely today.\\
Clausal-bias\\
 \> TC \> Bill hoped that Jill arrived safely today.\\
 \> RC \> Bill hoped Jill arrived safely today.
\end{tabbing}
\stopx

In their first two experiments, FH found no influence of verb bias on the
garden-path effect between the RC and TC conditions.  They did find a weak
influence in the third experiment.  These results support their claim that
Minimal Attachment is relevant for first-pass processing, and lexical
properties such as argument-preferences are considered only in subsequent
processing.  But there are some serious flaws with their experiment.  First,
they did not demonstrate that their (at least partially) intuitively-arrived-at
verb categories do indeed correspond to frequency of use or any other measure
of argument-structure bias.  Second, many of their examples give rise to
semantic implausibilities in the NP reading, e.g.\

\startxl{FHp}
Jan warned the fire...\\
Ed asserted eggs...\\
Ed disputed eggs...
\stopx

These implausibility problems affect the the NP-bias materials and S-bias
materials in different frequencies, thus introducing a serious potential source
of difficulty.

Garnsey, Lotocky and McConkie (1992)\nocite{GLM92} conducted one experiment
where they tested whether lexical-bias information is quickly incorporated by
the processor.  They used an eye-movement experiment with materials such as

\startxl{GLMm}
\begin{tabbing}
\ \ \= TC \ \ \= \kill
NP-bias\\
 \> TC \> The manager overheard that the employees were planning a surprise 
party. \\
 \> RC \> The manager overheard the employees were planning a surprise party.\\
Clausal-bias\\
 \> TC \> The manager suspected that the employees were planning a surprise 
party. \\
 \> RC \> The manager suspected the employees were planning a surprise party. 
\end{tabbing}
\stopx

They found a statistically significant interaction between complementizer
presence and verb-bias.  This effect appears at the first fixation on the first
disambiguating word (`were').

Trueswell, Tanenhaus and Kello (1993)\nocite{TTK93} argued that
subcategorization information is incorporated into the analysis at the earliest
point possible, and furthermore that the effect of verb subcategorization is
not categorical, but graded, reflecting `preferences' which show up in both
production and parsing tasks.  They used a cross-modal naming task with
materials such as:

\startxl{TTK1}
The old man insisted/accepted (that) $\ldots$
\stopx

The visually presented targets were `he' and `him'.  TTK found that for TR bias
verbs, absence of a complementizer commits the reader to an NP-complement
analysis, as can be seen by facilitation of `him' relative to `he'.
S-complement bias verbs ranged in the effect of the complementizer: TC-bias
verbs tended to require the complementizer in order to activate the
S-complement analysis, whereas RC-bias verbs showed activation of the
S-complement analysis even in the absence of the complementizer.  TTK found
converging evidence from two other experiments which used non-cumulative
word-by-word self-paced reading and eye tracking, respectively with materials
such as:

\startxl{TTK2}
The student forgot/hoped (that) the solution was in the back of the book.
\stopx

They found garden path effects, as can measured by the slowdown in the
disambiguating region, in RC sentences with TR-bias verbs.  For S-complement
bias verbs, the extent of garden path depended on how frequently the verb
appears with a that-complement versus zero-complement.  TTK used two forms of
statistical analysis: They used ANOVA to argue for an effect of TR-bias versus
S-bias verbs, and a regression statistic to argue for a correlation between the
strength of complementizer preference and the degree of garden path for the
S-complement bias verbs.

Ignoring the older (potentially problematic) self-paced studies, where FH's
results conflict with GLM's and TTK's, the latter studies are more believable:
FH's failure to find an effect (of verb bias) could be due to a variety of
factors. (as discussed at length in Trueswell
\etal\ 1993).

{}From the experiments listed above, I conclude the following:
\begin{enumerate}
\item Absence of a complementizer in a RC string can lead to processing 
difficulty.
\item The magnitude of this difficulty is often less than that of standard 
garden-path sentences.
\item Some of this difficulty might well persist when the complementizer 
is present.
\item There is some evidence suggesting that the magnitude of the effect 
becomes is higher
when the ambiguous NP is long.
\item The magnitude of the difficulty is sensitive to the subcategorization
possibility/preference of the matrix verb.
\end{enumerate}

In short, the evidence for the strong influence of lexical factors is clear.
But there is some evidence that some processing difficulty is residually
associated with sentential complements, independent of ambiguity and lexical
factors.

\subsection{Late Closure Ambiguity}
\label{sec:LCA}

Frazier and Rayner (1982)\nocite{FrazierRayner82} argue that Frazier's (1978)
structural preference principle of Late Closure \ref{LCdef} is what is
responsible for the garden path in \ref{LCeg}a.

\startxl{LCdef}
{\bf Late Closure}\\ When possible, attach incoming lexical items into the
clause or phrase currently being processed (i.e.\ the lowest possible
nonterminal node dominating the last item analyzed).
\stopx

\startxl{LCeg}
a. Since Jay always jogs a mile seems like a short distance to him.\\
b. Since Jay always jogs a mile this seems like a short distance to him.
\stopx

Late closure can similarly account for the processing difficulty in \ref{her}
and \ref{stalin}.

\startxl{her}
Without her contributions failed to come in.  (From Pritchett
1987\nocite{Pritchett87})
\stopx

\startxl{stalin}
When they were on the verge of winning the war against Hitler, Stalin,
Churchill and Roosevelt met in Yalta to divide up postwar Europe.  (From Ladd
1992\nocite{Ladd29})

\stopx

Frazier and Rayner's garden-path theory distinguishes two components of the
human language processor, which I will follow Mitchell
(1987)\nocite{Mitchell87} in calling the Assembler and the Monitor.  The
Assembler very quickly hypothesizes a syntactic structure for the words
encountered thus far.  The Monitor evaluates this hypothesis and sometimes
initiates backtracking when it detects a semantic problem. The Assembler uses
quickly-computable strategies like Minimal Attachment and Late Closure.
Mitchell (1987) investigated a prediction of the garden-path theory --- that
the Assembler only pays attention to major grammatical categories of the
incoming lexical items.  Finer distinctions, such as verb subcategorization
frames, are only considered in the Monitor.  It follows that Late Closure
effects, as in \ref{LCeg} should persist even when the first verb is purely
intransitive, as in \ref{sneeze}

\startxl{sneeze}
After the child had sneezed the doctor prescribed a course of injections.
\stopx

Mitchell's used subject-paced reading-time measurement. Instead of a
word-by-word procedure, each keypress would present the next segment of the
sentence.  Segments were fairly large --- each test sentence was divided into
only two segments.

As he predicted, Mitchell found garden path effects when the segment boundary
was after the ambiguous NP `the doctor'.  But this effect could arise as an
artifact of the way he segmented his materials, leading the reader to construe
each segment as clause.\footnote{Just because readers try to make sense of a
constituent such as `after the child had sneezed the doctor', it does not mean
that they are ignoring subcategorization information.  It is not clear that
putatively intransitive verbs such as `sneeze', `burp', `sleep' are indeed
ungrammatical when used transitively, or merely lead to implausibilities.  It
could be that in a sentence like
\ref{sneeze}, whatever it is that is responsible for the late closure effect 
is driving the interpreter to come up with a transitive verb interpretation.
Transitive uses of many putatively intransitive verbs are not impossible ---
consider `slept his fare share', `burped {\em Yankee Doodle}', `sneezed her way
out of the office', `sneezed his brains out'.} To address this criticism,
Mitchell and his colleagues (Adams, Clifton and Mitchell
1991\nocite{AdamsEtal91}) conducted an eye tracking study.  Using materials as
in \ref{Adams1}, they manipulated the availability of a transitive reading for
the verb, and whether or not there was a disambiguating comma after the
preposed clause.

\startxl{Adams1}
After the dog struggled/scratched (,) the vet took off the muzzle.
\stopx

Their results suggest that when the comma was omitted, subjects attempted to
construe the ambiguous NP `the vet' as the object of the preceding verb, even
when it was purely intransitive.  In other words, the lexical property of
intransitivity was not as effective as the comma in avoiding a transitive
analysis.

Stowe (1989, experiment 1)\nocite{Stowe89} provides evidence that directly
contradicts Mitchell's claim that verb subcategorization information is ignored
by the first phase of sentence processing.  Stowe exploited the phenomenon of
causative/ergative alternation, exemplified in \ref{CausativeErgative} to show
that readers are immediately sensitive to the subcategorization frames
available for the verb.

\startxl{CausativeErgative}
\begin{tabbing}
Causative:  \= John moved the pencil.\\
Ergative:   \> The pencil moved.
\end{tabbing}
\stopx

Using materials such as those in \ref{Stowem}, Stowe manipulated the
plausibility of the subject as causal agent, effectively changing
subcategorization frame of the verb.

\startxl{Stowem}
\begin{tabbing}
\ \ \ Inanimate: \= \kill
Ambiguous: \\
\ \ \ Animate: \ \> Before the police stopped the driver was already 
getting nervous.\\
\ \ \ Inanimate: \> Before the truck stopped the driver was already 
getting nervous.\\
Unambiguous:\\
\ \ \ Animate:   \> Before the police stopped at the restaurant the 
driver was already...\\
\ \ \ Inanimate: \> Before the truck stopped at the restaurant the 
driver was already...
\end{tabbing}
\stopx

Late Closure predicts that in the ambiguous conditions, the disambiguating word
(`was' in \ref{Stowem}) should not vary when the animacy of the subject is
manipulated.  Stowe's account of early use of lexically-specified information
predicts a garden path effect at `was' only in the ambiguous-animate condition.
And this is exactly what she found.  Using a subject-paced word-by-word
cumulative display task where subjects were required to monitor the
grammaticality of the string, she found significantly elevated reading times
and `ungrammatical' responses at the first disambiguating word in the
ambiguous-animate condition, and nowhere else.  The experimental technique used
by Stowe has often been criticized as too slow for detecting first-pass
processes, But I am aware of no experiments which challenge Stowe's result.

In a followup experiment, Stowe investigated the interaction between lexical
preferences and plausibility.  She used materials such as those in
\ref{Stowem2}.

\startxl{Stowem2}
\begin{tabbing}
\ \ \ Implausible: \= \kill
Animate: \\
\ \ \ Plausible: \ \> When the police stopped the driver became very 
frightened.\\
\ \ \ Implausible: \> When the police stopped the silence became very 
frightening.\\
Inanimate:\\
\ \ \ Plausible:   \> When the truck stopped the driver became very 
frightened.\\
\ \ \ Implausible: \> When the truck stopped the silence became very 
frightening.
\end{tabbing}
\stopx

She used the same procedure as in her first experiment --- a subject-paced
word-by-word cumulative display task where subjects were required to monitor
the grammaticality of the string at each word.  Aside from the animacy effect
observed in her first experiment, Stowe found that ``$\ldots$ the
implausibility of the subject NP [`silence' in \ref{Stowem2}] to serve as an
object of the preceding verb is noted as soon as the word itself appears.''
(p. 339) Stowe also observed ``The most perplexing point about the results of
Experiment 2 is that people apparently become aware of the unsuitability of the
NP to be an object of the preceding verb even when there is evidence that they
expect an intransitive verb structure.  [i.e.\ in the Inanimate conditions]''
(p. 341)

In summary, While the issue of whether verb-subcategorization information comes
to bear immediately on resolving the late-closure ambiguity is not definitively
settled, the available evidence suggests that it does.  Nevertheless, there is
still evidence for some residual effects (Late Closure, and preference for NP
complements over S complements) that lexical properties alone cannot account
for.  Below is additional evidence for this claim:

Consider

\startxl{realize}
John finally realized just how wrong he had been remained to be seen.
\stopx

The main verb, `realize', is biased toward a sentential
complement,\footnote{Verb data from five sources confirm this:\\

\mytable{
\begin{tabular}{|lrrrrl|}\hline &  NP     &   RC   &   TC    & RC+TC 
& units \\ \hline
Trueswell \etal (1993)          &   7     &   35   &   58    &  93   
& \% in completion task\\
Garnsey \etal\ (1992)           &  13     &   31   &   46    &  77   
& \% in completion task 
					(Garnsey p.c.\ 1992)\\
Connine \etal\ (1984)	        &  11     &    ?   &    ?    &  26   
& frequency in questionnaire\\
Brown corpus                    &  37     &   64   &   78    &  142  
& raw frequency\\
Wall Street Journal corpus      &  18     &   16   &   15    &  31   
& raw frequency \\ \hline
\end{tabular}}
 }
yet there is still a perceptible garden path.

Lexical bias alone also fails to account for all of the late closure effect.
In

\startxl{returned}
When Mary returned her new car was missing. 
\stopx

the verb `return' occurs more frequently without an object than with
one.\footnote{ The verb `return' occurs in the Brown and Wall Street corpora
as follows:

\mytable{\begin{tabular}{|lcc|} \hline
corpus      		& transitive  & intransitive \\ \hline
Wall Street Journal 	&	36	  & \hspace{.5em}75 \\
Brown       		& 	18        &    128 \\ \hline
\end{tabular}}
}
\footnote{ Note that while the verb `return' has both an intransitive and a 
transitive subcategorization frame, it is different from a verb like `eat'
which is transitive, but may drop its object.  It may be the case that
object-drop uses require a process of accommodating an implicit object.  While
difficulties with this process could potentially account for the garden path in
\ref{cannibals} they cannot account for a garden path in
\ref{returned}.}
Nor can lexical bias account for the garden-path sentences \ref{her} and
\ref{stalin}, repeated here as \ref{herBis} and \ref{stalinBis}. 

\startxl{herBis}
Without her contributions failed to come in.
\stopx

\startxl{stalinBis}
When they were on the verge of winning the war against Hitler, 
Stalin, Churchill and Roosevelt met in Yalta to divide up postwar Europe.
\stopx

In the rest of this chapter I investigate two theories to account for these
preferences.

\section{Degree of Disconnectedness}
\label{sec:Disconnectedness}

One idea that has been recently put to use by Pritchett
(1988)\nocite{Pritchett88} and Gibson (1991)\nocite{Gibson91}, but goes back at
least as far as Eady and Fodor (1981)\nocite{EadyFodor81} and Marslen-Wilson
and Tyler (1980)\nocite{Marslen-WilsonTyler80} is that the processor has
difficulty keeping around many fragments for which it does not yet have
semantic connections, and thus prefers better-integrated analyses.  In this
section I give one formalization of this idea and show how it can account for
many different pieces of data, including those just discussed.

The basic notion here is degree of {\em disconnectedness.} Intuitively, the
disconnectedness measure of an analysis of an initial segment of a sentence is
how many semantically unrelated pieces have been introduced so far.  In
`standard' syntactic theory disconnectedness has a straight-forward
implementation:

\startxl{ThetaAttachment}
{\em Theta Attachment:} (Pritchett 1988,
1992)\nocite{Pritchett88,Pritchett92}\\ The theta criterion attempts to be
satisfied at every point during processing given the maximal theta grid.
\stopx

The theta criterion is part of the competence theory which assigns every verb
(and other open-class complement-taking words such as adjectives and nouns) a
collection of thematic `slots', called theta-roles.  For a sentence to be
well-formed, every theta-role must be filled by an argument, and every argument
must fill a particular slot.  It turns out that thematic roles are not rich
enough to capture the necessary semantic relations among words in a sentence,
especially when their semantic content (e.g. {\sc agent, instrument}) is
ignored.  So Pritchett broadens his heuristic to include every principle of
syntax, not just the theta-criterion.  Gibson (1991)\nocite{Gibson91}
operationalizes \ref{ThetaAttachment} in a slightly different way to make it
work with his parsing algorithm and data representation, and notes that any
syntactic theory that mentions thematic relations would give rise to a similar
parsing heuristic.  In this project, I cast the notion of disconnectedness
minimization in purely semantic (i.e.  non-syntactic) terms.  I do not
distinguish the semantic relation of `thematic role' from any other semantic
relation such as `determiner-noun' or `modal-verb' etc.  This notion of
disconnectedness will be made more concrete presently.  But first, I introduce
the semantic representation formalism which I will use in this dissertation.

\subsection{A Representation of Semantics}
\label{sec:SemanticRepresentation}

For the purposes of the present project, the semantic representation which I
choose is borrowed from the work of Hobbs and his colleagues (Hobbs 1985;
Hobbs, Stickel, Appelt and Martin 1993\nocite{Hobbs85}\nocite{HobbsEtal93})
which is in turn an elaboration of work by Davidson (1967)\nocite{Davidson67}.
Davidson argues that events can be talked about, just like physical objects, so
a logical form must include event variables as well as the traditional `thing'
variables.  Hobbs (1985) argues that predications (e.g.\ states) must also be
afforded this treatment as first class members of the ontology.  The semantic
representation which he proposes is not the usual term or logical formula but
rather a set of terms, each comprising a predicate symbol and one or more
arguments which are either variables or constants, but crucially not terms
themselves.  All variables are (implicitly) existentially quantified.  For
example, the semantic representation for \ref{BoyWanted} is
\ref{BoyWantedS}.

\startxl{BoyWanted}
The boy wanted to build a boat quickly.
\stopx

\startxl{BoyWantedS}
\(\exists e_1,e_2,e_3,x,y\ Past(e_1)\wedge want'(e_1,x,e_2)\wedge 
quick'(e_2,e_3)\wedge
build'(e_3,x,y)\wedge boy'(x)\wedge boat(y)\)
\stopx

Which means something like
\begin{quote}
There is an event/state $e_1$ which occurred in the past, in which the entity
$x$ wants the event/state $e_2$.  $e_2$ is an event/state in which the
event/state $e_3$ occurs quickly.  $e_3$ is a building event in which $x$ build
$y$, where $x$ is a boy and $y$ is a boat.
\end{quote}

Hobbs (1985) motivates this `flat' representation on the grounds that it is
simpler, and thus encodes fewer commitments in the level of the semantics.
This is superior to hierarchical, recursively-built representation, he argues,
as semantic representation is difficult enough as it is without additional
requirements that it cleverly account for certain syntactic facts as well
(e.g.\ count nouns vs. mass nouns).  He defends the viability of this approach
by showing that it can cope with traditional semantic challenges such as opaque
adverbials, {\em de dicto/de re\/} belief reports, and identity in belief
contexts.  This notation is used in the {\sc tacitus\/} project (Hobbs
\etal\ 1988, 1990) --- a substantial natural language understanding system,
demonstrating its viability as a meaning representation.  Haddock (1987,
1988)\nocite{Haddock87}\nocite{Haddock88} exploits the simple structure of
each term to perform efficient search of the representation of a prior
discourse in order to resolve definite NPs.

In the current project, a lexicalized grammar is used where each word is
associated with a combinatory potential and a list of terms.  When words
(constituents) are combined, their term lists are simply appended to determine
the term list of the combined constituent.  Details and examples will be given
in \refsec{sec:grammar}.  The semantic analysis, then, develops incrementally
word-by-word.

\subsection{A Formal Definition}
\label{sec:DefDisconnect}

To formally define the degree of disconnectedness of a semantic analysis S, I
first construct an undirected graph whose vertices are the variables (both
ordinary `thing' variables and `event/state' variables) mentioned in S.  Two
vertices are adjacent (have an edge connecting them) just in case they both
appear as arguments of a term in S.  The disconnectedness measure of S is the
number of components of the graph, minus 1.  By {\em number of components\/} I
mean the standard graph-theoretic definition: two vertices are in the same
component if and only if there is a path of edges that connects them.

For example, when the initial segment in \ref{cannibals1} is encountered,

\startxl{cannibals1}
When the cannibals ate the missionaries\ldots
\stopx

there are two analyses corresponding to the transitive and intransitive
readings of `ate', respectively:

\startxl{cannibals2a}
when(e1,e2), eat(e1,e3,e4), definite(e3), cannibals(e3), definite(e4), 
missionaries(e4)\\
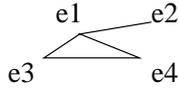
\psfig{figure=cannibals2a.eps,silent=y}
\stopx

\startxl{cannibals2b}
when(e1,e2), eat(e1,e3), definite(e3), cannibals(e3), definite(e4),
missionaries(e4)\\
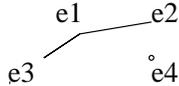
\psfig{figure=cannibals2b.eps,silent=y}
\stopx

Since the intransitive reading carries a higher disconnectedness measure than
the transitive reading (1 vs. 0) the transitive reading is preferred.  A Late
Closure Effect therefore results when the next word is `drank'. I use the
capitalized term {\em Disconnectedness\/} to refer to the theory that the
processor prefers to minimize the measure of disconnectedness.

Disconnectedness similarly accounts for the garden path effects in
\ref{herBis}.  At the word `contributions' there are two analyses corresponding
to the common noun and NP readings, respectively.

\startxl{herCN}
without(e1,e2), feminine(e1), of(e1,e3), contributions(e3)
\stopx

\startxl{herNP}
without(e1,e2), feminine(e1), implicit-quantifier(e3),\footnotemark\ 
contributions(e3)
\stopx
\footnotetext{This is a placeholder for a
semantic theory of bare plurals.}

The common noun reading is thus preferred.

\subsection{Consequences}
\label{sec:DisConsequences}

Disconnectedness predicts difficulty with \ref{realize} -- \ref{stalinBis}.  In
fact, since Disconnectedness is insensitive to lexical or conceptual
preferences, its input to the analysis selection process could conflict with
the input from lexical preferences.  This conflict can account for the puzzling
findings of Stowe's (1989) second experiment above.

The findings of Holmes \etal\ (1987)\nocite{HKM87} and Kennedy \etal\
(1989)\nocite{KMJR89} that in sentences such as \ref{HKM87m} above, repeated
here as \ref{HKM87mBis}, the clausal conditions TC and RC are slower to read
than TR, are also consistent with the additional disconnectedness associated
with the subject reading of `the safe's location within the house'.

\startxl{HKM87mBis}
(TR) The maid disclosed the safe's location within the house to the officer.\\
(TC) The maid disclosed that the safe's location within the house had been 
changed.\\
(RC) The maid disclosed the safe's location within the house had been changed.
\stopx

Additional evidence in support of Disconnectedness comes from experiments with
filling WH-gaps. Boland, Tanenhaus, Carlson, and Garnsey
(1989)\nocite{Boland89} investigated whether plausibility affects gap-filling.
They used materials such as those in \ref{Remind} and a subject-paced
word-by-word cumulative display method where subjects were asked to detect when
the sentence stopped making sense.  They found that the word `them' caused
difficulty in \ref{Remind}a as compared to its control \ref{Remind}b.

\startxl{Remind} 
a. Which child did Mark remind them to watch this evening?\\
b. Which movie did Mark remind them to watch this evening?  
\stopx

Boland \etal\ conclude from this and other experiments that inferential
information such as the argument structure of the verb are used as soon as
logically possible.  Let us examine closely what happens with these two
sentences.  In \ref{Remind}b when the reader comes to the word `remind' s/he
can check whether movies can be reminded.  Since that is implausible, the
remindee spot is not filled, and the next word `them' causes no difficulty.  In
\ref{Remind}b, a child is something that can be reminded so the gap-filling
analysis is pursued.  But the non-filling alternatives is just as plausible!  A
person can remind someone of something having to do with children.
Plausibility alone cannot fully explain why the filling analysis is preferred
in this case. Note that the non-filling analysis has a higher disconnectedness
measure --- the relation between the WH-element and the rest of the material in
the utterance is not established.  Without Disconnectedness, one need a
partially structurally-based theory, such as `first plausible gap' to account
for this gap-filling behavior.

The interpretation of the results of Holmes \etal\ (1987)\nocite{HKM87} and
Kennedy \etal\ (1989)\nocite{KMJR89} as dis\-connected\-ness-related processing
difficulty in the unambiguous TC condition suggest that there might be other
unambiguous, highly disconnected structures which are hard to process.  Indeed,
center embedding \ref{ratcat}, the classical example of an unambiguous
structure that is hard to process, reaches a disconnectedness measure of 2
after the word `dog'.

\startxl{ratcat}
The rat that the cat that the dog$\ldots$\hspace{\fill} \\
definite(e1), rat(e1), definite(e2), cat(e2), definite(e3), dog(e3)\hfill\\
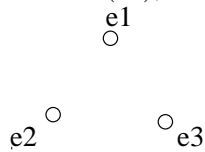
\psfig{figure=ratcat.eps,silent=y}\hfill\\[3mm]

The rat that the cat that the dog bit chased died.\hfill\\ definite(e1),
rat(e1), definite(e2), cat(e2), definite(e3), dog(e3)\hfill\\ bite(e4,e3,e2),
past(e4), chase(e5,e2,e1), past(e5), die(e6,e1), past(e1)\hfill\\
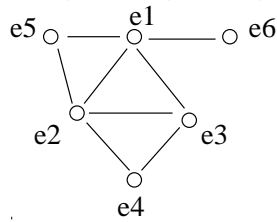
\psfig{figure=ratcatdied.eps,silent=y}
\stopx

It seems quite likely that the computations of processing load which Hawkins
(1990)\nocite{Hawkins90} uses to derive many word-order universals could be
recast in terms of disconnectedness score.  Rambow's
(1992a,b\nocite{Rambow92a}\nocite{Rambow92b}) account of marginally grammatical
scrambled sentences in German in terms of storage requirements is also very
likely to be statable in terms of disconnectedness.  But this awaits further
research.

Given that disconnectedness-related processing difficulties, such as the Late
Closure Effect, are mitigated, and often overridden by inferential preferences,
one would expect processing difficulties with structures such as
\ref{ratcat} to be ameliorated by better semantic `coherence'.  In fact this
prediction is borne out.  Bever (1970)\nocite{Bever70} hypothesized that
\ref{dogdestruction} might be easier to process than \ref{ratcat}.
\startxl{dogdestruction} 
The dog that the destruction that the wild fox produced was scaring will run
away fast.
\stopx

Fodor Bever and Garrett (1974)\nocite{FodorBeverGrarrett74} and Frazier and
Fodor (1978)\nocite{sausage} mention \ref{fish} and \ref{snow}, respectively,
which seem somewhat easier to understand than \ref{ratcat}.

\startxl{fish}
The water the fish the man caught swam in was polluted.
\stopx

\startxl{snow}
The snow that the match that the girl lit heated melted.
\stopx

Frank (1992)\nocite{Frank92} provides \ref{Italian} which seems to do away with
processing difficulty altogether.
\startxl{Italian}
A book that some Italian I've never heard of wrote will be published soon by
MIT press.
\stopx

Inferential and discourse factors are clearly involved in the degree of
difficulty of these sentences.  For example, note that having a deictic as the
most deeply embedded subject (as in \ref{Italian}) seems to improve things
somewhat; and replacing the definite subjects in \ref{ratcat} --- \ref{snow}
with indefinites, in \ref{Italian} seems to make a further improvement.  The
interaction between these interpretive factors and processing difficulty in the
absence of ambiguity remain matters for further research, as do the subtle
effects of the choice of relativizer: {\em that\/} vs. {\em who/whom/which\/}
vs.  zero.\footnote{Whatever manipulations one applies to \ref{ratcat} to make
it as good as \ref{Italian} can be applied `in reverse' to cause
center-embedding-type processing difficulty for structures which are usually
considered unproblematical.  Consider (i) which Gibson (1991)\nocite{Gibson91},
following Cowper (1976)\nocite{Cowper76} takes to be unproblematical.

\startxr{i}
The possibility that the man who I hired is incompetent worries me.
\stopxr

\noindent Replacing the deictic pronouns with a definite NPs renders the 
resulting sentence (ii) harder to understand.

\startxr{ii}
The possibility that the man who the executive hired is incompetent worries the
stockholders.
\stopxr
}

The connection between ambiguity resolution preferences for semantically
better-integrated readings and processing difficulties with center embedding
has been explored by Gibson (1991)\nocite{Gibson91}.  While Gibson's measure of
semantic integration is formulated in terms of the Government and Binding
principles (e.g.\ the $\theta$ Criterion, see Chomsky 1981\nocite{Chomsky81}),
and not graph theoretic notions, his proposed underlying mechanisms are
comparable to the account offered here --- analyses in which semantic relations
among entities are established are preferred to analyses in which they are not.
Gibson opts for a different explanation for the relative improvement of
\ref{snow} over \ref{ratcat}.  He assumes that each of \ref{ratcat} through
\ref{snow} (and presumably
\ref{Italian}) overwhelms the parser's capacity and causes a breakdown of 
ordinary syntactic processing.  In sentences like \ref{snow}, the interpretive
module is still able to piece the uncombined fragments together using
inferential processes such as determinations of plausibility.  This sort of
inference cannot salvage a sentence such as \ref{ratcat}.  Gibson's account 
predicts that in deeply embedded structure where the parser
breaks down, if there is a choice between syntactic ill-formedness and
inferential implausibility, the former will be opted for by the inferential
salvaging process.  For example, the string

\startxl{ScrambledItalian}
Some Italian that a book I've never heard of wrote will be published soon by
MIT press.
\stopx

is predicted by his account to be judged as acceptable\footnote{I assume that
the string in \ref{ScrambledItalian} is somehow derivable by a combination of
scrambling operations which operate in other languages but cannot be ruled out
for this English sentence because the competence grammar is being ignored.}
(or, at least, significantly better than \ref{ratcat}) and construed as meaning
the same thing as \ref{Italian}.

There is no necessary connection between center embedding and disconnectedness.
\ref{spoil1} is just as center embedded as \ref{ratcat} but does not encounter
disconnectedness at any point.

\startxl{spoil1}
John asked the woman that gave the boy that kicked the dog some candy why she's
spoiling him.
\stopx

Intuitively\footnote{To corroborate my intuitions I conducted a miniature
survey of six colleagues. I presented them with sentences 1 through 4.
\begin{enumerate} 
\item John asked the woman that gave the boy that kicked the dog some candy 
why she's
spoiling him.
\item John asked the woman that gave the boy that the dog frightened some 
candy why she's
spoiling him.
\item The woman that the boy that the dog frightened is bothering right now 
is a friend of John's.
\item The woman that the boy that kicked the dog is bothering right now is a 
friend of John's. 
\end{enumerate} 
Their maximum disconnectedness measures are 0 1 2 and 1, respectively.
Everyone I asked initially rated all sentences as equally bad.  After some
begging and coaxing on my part, each informant provided some partial ranking.
All responses were consistent with the ranking 1, 4, 2, 3 from best to worst
(and only this ranking).  This is consistent with the predictions of
Disconnectedness.}
\ref{spoil1} is slightly easier to read than \ref{spoilV} --- 
a variant whose structure
directly mirrors that of \ref{ratcat}.

\startxl{spoilV}
The woman that the boy that the dog frightened is bothering right now is a
friend of John's.
\stopx

Eady and Fodor (1981)\nocite{EadyFodor81} report an experiment in which they
independently manipulated two relative clauses --- one contained in the other
--- for center embedding versus right-branching.  They found that \ref{Eady}a
and \ref{Eady}b were of comparable reading difficulty; \ref{Eady}c was
substantially harder to read than \ref{Eady}b, and \ref{Eady}d was harder yet.
The largest difference was between \ref{Eady}b and \ref{Eady}c.  That is, when
the innermost relative clause is center-embedded, the difficulty is greatest.
Their results argue against an account of processing difficulty which is, in
their words, ``based on inherent properties of center-embedding.''
Descriptively, what matters most is whether or not the filler-gap dependencies
overlap.  Disconnectedness theory captures this finding: the maximum
disconnectedness scores for \ref{Eady}a through \ref{Eady}d are 1, 1, 2, and 2
respectively.

{
\newcommand{\subi}{$_i$}
\newcommand{\subj}{$_j$}
\newcommand{\gap}{$e$}

\startxl{Eady}
\begin{tabbing}
a. Jack met the patient$\underline{\mbox{\subi\ the nurse sent \gap\subi}}$ to 
the doctor$\underline{\mbox{\subj\ the clinic had hired \gap\subj}}$.\\[1mm] 
b. The patient$\underline{\mbox{\subi\ the nurse sent \gap\subi}}$ to the
doctor$\underline{\mbox{\subj\ the clinic had hired \gap\subj}}$ met Jack.
\\[1mm] 
c. Jack met the patient$\underline{\mbox{\subi\ the 
nurse$\underline{\mbox{\subj\ the clinic
had hired \gap\subj}}$\ sent \gap\subi}}$ to the doctor.\\[.5mm]
d. The patient$\underline{\mbox{\subi\ the nurse$\underline{\mbox{\subj\ 
the clinic
had hired \gap\subj}}$\ sent \gap\subi}}$ to the doctor met Jack.\\[.5mm]
\end{tabbing}
(underlining depicts filler-gap dependencies)
\stopx
}

A Disconnectedness-based account of the difficulty of \ref{ratcat} would
predicts that \ref{spoil1} should be completely free of any
center-embedding-type processing difficulty, and that an even more deeply
nested structure \ref{DeepToo} should be as easy to process as its purely
right-branching control in \ref{DeepLarynx}.  This does not seem to be the case
--- more research is needed.

\startxl{DeepToo}
John asked the woman that offered the boy that gave the dog that chased the cat
a big kick some candy why she's spoiling him.
\stopx

\startxl{DeepLarynx}
John met the woman that rewarded the boy that kicked the dog that chased the
cat.
\stopx

In summary, the strategy of minimizing the measure of disconnectedness has a
variety of evidence to support it:

\begin{itemize}
\item residual Late Closure Effects 
\item residual NP preference for NP vs.\ S ambiguities
\item gap-filling 
\item processing difficulty in unambiguous, temporarily disconnected sentences
\end{itemize}

But would adoption of Disconnectedness weaken the overall thesis?  After all,
Disconnectedness is stated over the sense-semantics of a string --- a level of
representation which is on the interface of syntax and interpretation.  It is
quite conceivable that one could propose a notational variant of
disconnectedness theory which is stated solely in terms of structure. (After
all, its theoretical predecessors --- Pritchett and Gibson's proposals --- are
based on thematic role assignment in syntactic structure.) Nevertheless, I
claim that Disconnectedness is a viable candidate for a component of the thesis
of ambiguity resolution from interpretation.  It is stated over the domain of
meaning, not syntactic structure. As is suggested by the susceptibility of
disconnectedness to discourse factors (e.g.\ \ref{Italian}) the locus of
disconnectedness might not be the sense-semantics as I defined it but a level
of meaning representation which is `deeper', more pragmatic.

Another potential objection is why should a temporarily high disconnectedness
measure matter to the processor?  Given that no complete grammatical sentence
has any disconnectedness, the processor can just patiently wait until the
connecting words arrive.  There are two responses to this objection.  First, a
processor that waits for additional information before making its decision
might require large computational resources when faced with compounding
ambiguity, i.e.\ waiting might be too expensive.  Second, a processor might
well be closely attuned to disconnectedness since the very task of a
sentence-understanding system is to determine the logical connection among the
words in the sentence --- the better the connection, the more preferred the
analysis. It would follow that some connection is preferable to no connection.

I now turn to a drastically different account for most of the data in this
section.

\section{Avoid New Subjects}

An examination of the syntactic structures that disconnectedness accounts for
reveals that with one exception, they all involve a preference not to analyze
an NP as a subject.

\startxl{DList}
\begin{tabbing}
a. {\bf late closure effects} \\
\hspace{10mm} \= When the cannibals ate the missionaries drank. \\
      \> Without her contributions failed to come in.  \\
   \> When they were on the verge of winning the war against Hitler, Stalin, \\
   \> \ \  Churchill and Roosevelt met in Yalta to divide up postwar Europe. \\
b. {\bf NP Preference for NP vs.\ S complement ambiguity} \\
      \> John has heard the joke is offensive.\\
c. {\bf subject relative clause center embedding} \\
      \> The rat that the cat that the dog bit chased died.\\
d. {\bf gap-filling } \\
      \> Which child did Mark remind them to watch this evening
\end{tabbing}
\stopx

The one exception is gap-filling.  The so-called {\em filled gap effect\/}
(Crain and Fodor 1985\nocite{CrainFodor85}) which readers experience in
\ref{DList}d at the word `them', tends to be less severe than the other garden
path effects discussed in \refsec{sec:Disconnectedness}.

In a second set of experiments, Boland \etal\ (1989)\nocite{Boland89} present
intriguing evidence that the processing difficulty in \ref{DList}d is not of
the same sort as the other garden-path effects in
\refsec{sec:Disconnectedness}.  Using the same subject-paced word-by-word
cumulative display stop-making-sense task that they used in their first
experiment described on \refpage{Remind} above, they investigated the effect of
the plausibility of the WH-filler on reading time. For materials as in
\ref{Granted}

\startxl{Granted}
a. Bob wondered which bachelor Ann granted a maternity leave to this 
afternoon.\\ 
b. Bob wondered which secretary Ann granted a maternity leave to this 
afternoon.
\stopx

they found that subjects were able to detect the anomaly in \ref{Granted}a
starting with the word `leave', that is, before the preposition `to' could
trigger the construction of the phrase which contains the gap position.  This
suggests that certain pragmatic integration processes occur before bottom-up
syntactic evidence is available to tell the processor that a gap is present.

It follows from this finding that encountering the unexpected NP `them' in
\ref{DList}d is odd not just syntactically but also pragmatically.  Indeed
Boland (p.c.\ 1992) reports varying strengths of filled-gap effects for
different lexical realizations of the `surprising' NP (e.g.\ pronoun, proper
name, indefinite NP, definite NP) suggesting that inference and accommodation
might be involved.  The difference between a filled-gap effect and a garden
path effect is then in the processing component in which they are detected: a
garden path is detected when the syntactic processor discovers that none of the
analysis that it is currently maintaining can be extended with the current
word. This condition results because the necessary analysis was {\em
discarded\/} earlier.  A filled-gap effect, on the other hand is initially
detected in the interpreter, not the syntactic processor.  When the surprising
NP appears, the interpreter has not yet told the syntactic processor to commit
to the filled-gap analysis.

With filled-gap effects now eliminated from the collection of garden-path data
that Disconnectedness is relevant for, Disconnectedness is indistinguishable,
on the remaining examples, from a preference for avoiding treating an NP as a
subject.  This is a very strange preference to have in a processor whose
purpose it is to understand sentences, given that every sentence has a subject!
Perhaps all of the subjects in the examples in \refsec{sec:Disconnectedness}
are somehow special, and the prohibition is not on all subjects, only on this
special sort of subject.  In this section I argue that this is indeed the case.
All of the sentences were presented out of context, and it is {\em
subjects\footnote{By `subject' I refer solely to canonical, pre-verbal
subjects, and not to the broader class of grammatical subject which may include
existential `there' sentences, and V2 constructions such as `Outside stood a
little angel.' {\em inter alia.}} that are new to the discourse\/} that the
processor seeks to avoid.  It must be emphasized that Avoid New Subjects makes
no primary distinctions between definite and indefinite NPs.  Out of context,
both are new to the discourse. In context, definites tend to be given more
frequently, but definiteness is not a defining characteristic of givenness.

\subsection{Given and New}
\label{sec:GivenNew}

Prince (1981)\nocite{Prince81} proposes a classification of occurrences of NPs
in terms of assumed familiarity.  When a speaker refers to an entity which s/he
assumes salient/familiar to the hearer, s/he tends to use a brief form, such as
a definite NP or a pronoun.  Otherwise the speaker is obliged to provide the
hearer with enough information to construct this entity in the hearer's mind.
Prince classifies the forms of NPs and ranks them from given to new:

\begin{description}
\item[evoked]  An expression used to refer to one of the conversation's 
participants or
an entity which is already under discussion.  (usually a definite NP or
pronoun)
\item[unused] A proper name which refers to an entity known to the speaker 
and hearer, but
not already in the present discourse.
\item[inferable] A phrase which introduces an entity not already in the
discourse, but which is easily inferred from another entity currently under
discussion. (c.f.  bridging inference of Haviland and Clark
(1973)\nocite{HavilandClark74})
\item[containing inferable] An expression that introduces a new entity and 
contains a reference to the extant discourse entity from which the inference is
to proceed. (e.g.\ `One of the people that work with me bought a Toyota.')
\item[brand new] An expression that introduces a new entity which cannot not be
inferentially related or predicted from entities already in the discourse. 
\end{description}

Prince constructs this scale on the basis of scale-based implicatures that can
be drawn if a speaker uses a form which is either too high or too low --- such
a speaker would be sounding uncooperative/cryptic or needlessly verbose,
respectively.

Using this classification, Prince analyzed two texts --- the first is an
informal chat and the second, formal scholarly prose.  Her findings are
summarized in the following table.

\mytable{
\begin{tabular}{|l|cc|cc|}
\hline & \multicolumn{2}{c|}{spoken} & \multicolumn{2}{c|}{written} \\ 
                           & subject                & non-subject 
& subject & non-subject\\  \hline
Evoked                     & 93.4\%                 &   48.8\%    
& 50.0\%  & 12.5\% \\
(containing) Inferable     & \rule{.8em}{0ex}6.6\%  &   30.2\%    
& 41.7\%  & 62.5\% \\
New (unused and brand new) & \rule{0.8em}{0ex}0.0\% &   20.9\%    
& \rule{0.8em}{0ex}8.3\% 
& 25.0\% \\
\hline
\end{tabular}}

In both genres there is a clear tendency to make subjects more given.  If we
construe this tendency as resulting directly from a principle of the linguistic
competence which calls for using subject position to encode given information,
we would indeed expect a reader to prefer to treat out-of-context NPs as
something other than subjects.  I refer to this principle as {\em Avoid New
Subjects.\/}

\subsection{Consequences}
\label{sec:ANSConsequences}

The theory of Avoid New Subjects predicts that for ordinary text (spoken or
written) the Late Closure Effect and the residual NP preference for NP vs.\ S
ambiguities should disappear.  I now present corpus-based investigations of
these two predictions in turn.

\subsubsection{Late Closure and Avoid New Subjects}
\label{sec:LCANS}

To test the prediction that Late Closure Effects should disappear when the
subject is given, I conducted a survey of the bracketed Brown and Wall Street
Journal corpora for the following configuration: a VP which ends with a verb
and is immediately followed by an NP.  Crucially, no punctuation was allowed
between the VP and the NP.  I then removed by hand all matches where there was
no ambiguity, e.g.\ the clause was in the passive or the verb could not take
the NP as argument for some reason.  Here are the remaining matches, preceded
by a bit of context, and followed by illustration/discussion of the ambiguity.

\begin{enumerate}
\item{} 
[An article about a movie describes how its composer approached one of the
singers.]  When you approach a singer and tell her you don't want her to sing
you always run the risk of offending.

[`You don't want her to sing you a song.']
\item 
{}From the way she sang in those early sessions, it seemed clear that Michelle
(Pfeiffer) had been listening not to Ella but to Bob Dylan. ``There was a
pronunciation and approach that seemed Dylan-influenced,'' recalled Ms.
Stevens.  Vowels were swallowed, word endings were given short or no shrift.
``When we worked it almost became a joke with us that I was constantly
reminding her to say the consonants as well as the vowels.''

[`When we worked it out$\ldots$']
\item 
After the 1987 crash, and as a result of the recommendations of many studies,
``circuit breakers'' were devised to allow market participants to regroup and
restore orderly market conditions.  It's doubtful, though, whether circuit
breakers do any real good.  In the additional time they provide even more order
imbalances might pile up, as would-be sellers finally get their broker on the
phone.

[Even though this example involves gap-filling, the fact remains that the NP
`even more order imbalances' could be initially construed as a dative, as in
`In the additional time they provide even the slowest of traders, problems
could$\ldots$']
\item{} 
[article is about the movie ``The Fabulous Baker Boys''.  Preceding paragraphs
describe the actors and movie in generalities.]  When the movie opens the Baker
brothers are doing what they've done for 15 years professionally, and twice as
long as that for themselves: They're playing proficient piano, face-to-face, on
twin pianos

[`The movie opens the Baker brothers to criticism from$\ldots$']
\item 
\label{default}
Jonathan Lloyd, executive vice president and chief financial officer of Qintex
Entertainment, said Qintex Entertainment was forced to file for protection to
avoid going into default under its agreement with MCA.  The \$5.9 million
payment was due Oct. 1 and the deadline for default was Oct. 19.  Mr. Lloyd
said if Qintex had defaulted it could have been required to repay \$92 million
in debt under its loan agreements.

[Both Webster's and American Heritage Dictionary classify the verb `default' as
both transitive and intransitive.  None of the 145 occurrences of `default' in
a larger corpus of Wall Street Journal text take an NP complement.]
\item 
What's more, the U.S. has suspended \$2.5 million in military aid and \$1
million in economic aid (to Somalia.) But this is not enough.  Because the
U.S. is still perceived to be tied to Mr. Barre, when he goes the runway could
go too.

[There are many transitive uses of `go' in the corpus: go a long way, a step
further, a full seven games, golfing, `town watching', home, nuts, hand in
hand.]
\item 
Butch McCarty, who sells oil-field equipment for Davis Tool Co., is
also busy.  A native of the area, he is back now after riding the
oil-field boom to the top, then surviving the bust running an Oklahoma
City convenience store. ``First year I came back there wasn't any
work,'' he says. ``I think it's on the way back now.

[First year I came back there I nearly...]
\item{} 
[Story about the winning company in a competition for teenage-run businesses,
its president, Tim Larson, and the organizing entity, Junior Achievement.]  For
winning Larson will receive a \$100 U.S. Savings Bond from the Junior
Achievement national organization.

[$\ldots$winning Larson over to their camp$\ldots$]
\item{} 
Why did the Belgians grant independence to [the Congo,] a colony so
manifestly unprepared to accept it? $\ldots$ Yet there were other
motivations$\ldots$ for which history may not find them guiltless.
[paragraph-break]

As the time for independence approached there were in the Congo no fewer than
120 political parties, or approximately eight for each university graduate.

[As the time for independence approached there, the people$\ldots$]

\item 

Science has simply left us helpless and powerless in this important sector of
our lives [spirituality].  \\ {} [paragraph-break] \\ The situation in which we
find ourselves is brought out with dramatic force in Arthur Miller's play The
Crucible, which deals with the Salem witch trials.  As the play opens the
audience is introduced to the community of Salem in Puritan America at the end
of the eighteenth century.

[the play opens the audience up to new$\ldots$]
\item{}  
[bodybuilding advice --- experimenting with a particular technique]
Oh, you'll wobble and weave quite a bit at first.  But don't
worry.  Before your first training experiment has ended there will be
a big improvement and almost before you know it you'll be raising and
lowering yourself just like a veteran! 

[Before your first training experiment has ended there in the room, you'll
know...]
\end{enumerate}

The givenness status of the ambiguous NPs is as follows:

\mytable{\begin{tabular}{cll}
match \# & NP 		& givenness status \\ \hline
1  &you 			& evoked	\\
2  &it 				& pleonastic	\\
3  &even more order imbalances 	& brand new	\\
4  &The Baker brothers 		& evoked	\\
5  &it 				& evoked	\\
6  &the runway 			& evoked	\\
7  &there                       & pleonastic	\\
8  &Larson                      & evoked	\\
9  &there                       & pleonastic	\\
10 &the audience                & inferable	\\
11 &there                       & pleonastic
\end{tabular}\hspace{2cm}
\begin{tabular}{lr}
\multicolumn{2}{c}{Summary:}\\ \hline
pleonastic	&	4\\
evoked		&	5\\
inferable	&	1\\
brand new	&	1
\end{tabular}
}

Prince's givenness scale does not include pleonastic NPs, since they do not
refer.  For the present purpose, it suffices to note that Avoid New Subjects
does not rule out pleonastics.\footnote{If one had to guess the perceived
givenness status of pleonastic, considering their tendency, cross
linguistically to be homophonous with pronouns and deictics, one would guess
that they are treated as given.} While the numbers here are too small for
statistical inference,\footnote{ Given the high frequency of given subjects,
optionally transitive verbs and fronted adverbials, one might expect more
matches in a two million word corpus.  But examination of the Wall Street
Journal corpus reveals that most fronted adverbials are set off by comma,
regardless of potential ambiguity.  Of 7256 sentence initial adverbials, only
8.14\% (591) are not delimited by comma.  Of these 7256 adverbials 1698 have
the category SBAR, of which only 4.18\% (71) are not delimited by comma.  The
great majority of fronted adverbials (4515) have category PP, of which 8.75\%
(433) are not delimited by comma.  The high frequency of the comma, therefore,
has the effect of significantly shrinking the available corpus of relevant
examples.} the data suggest that the prediction of Avoid New Subjects is
maintained.\footnote{It must be emphasized that these findings are just
suggestive.  Just because a particular sentence appears in a newspaper it does
not mean that that it did not cause the proofreader to garden path.  (This is
especially true of sentence 3 above, which causes some readers to garden-path.)
The only way to really test the current hypothesis is using carefully
constructed minimal pairs.}

Avoid New Subjects also provides an account for the ``perplexing'' results of
the second experiment reported by Stowe (1989), as discussed in the beginning
of \refsec{sec:DisConsequences}.  Recall that Stowe used materials such as

\startxl{Stowem2Bis}
\begin{tabbing}
\ \ \ Implausible: \= \kill
Animate: \\
\ \ \ Plausible: \ \> 
When the police stopped the driver became very frightened.\\
\ \ \ Implausible: \> 
When the police stopped the silence became very frightening.\\
Inanimate:\\
\ \ \ Plausible:   \> 
When the truck stopped the driver became very frightened.\\
\ \ \ Implausible: \> 
When the truck stopped the silence became very frightening.
\end{tabbing}
\stopx

For the animate condition, one expects an effect of implausibility at the
critical NP `the silence' because the reader is using the causative analysis of
`stopped'.  Given the evidence from her first experiment (using sentences like
the inanimate plausible in \ref{Stowem2Bis}) that inanimate subjects cause
readers to adopt the ergative analysis, one would not expect the reader to
consider the object analysis of the critical NP for inanimate conditions.  But
this is exactly what Stowe found --- implausibility effects for the inanimate
condition which mirrored those for the animate condition.

To resolve this paradoxical findings, one must make two observations.  First,
while the inanimate subject (`truck') indeed rules out a causative analysis for
the verb (`stopped'), it does not necessarily rule out all other transitive
analyses.  In particular, `stopped' allows a third subcategorization frame ---
the so-called instrumental.

\startxl{CausativeErgativeInstrumental}
\begin{tabbing}
Instrumental: \= \kill
Causative:    \> John moved the pencil.\\
Ergative:     \> The pencil moved.\\
Instrumental: \> The pencil moved the paper.
\end{tabbing}
\stopx

Unlike the ergative, the subject of an instrumental is not the patient
(affected object).  The name is somewhat of a misnomer because in examples such
as \ref{sleet}, the `instrumental' subject might not be serving as an
instrument of any causal agent\footnote{Theological arguments to the contrary
notwithstanding.}.

\startxl{sleet}
The sleet stopped the parade short.
\stopx

The second observation is that in the inanimate plausible condition, the
givenness status of the critical NP, `the truck' is inferable.  In the
inanimate implausible condition it is brand new --- more new in Prince's terms
than an inferable.

Both of these observation are true of the great majority of the experimental
materials in Stowe (1989).  Given the availability of a transitive verb
analysis of the inanimate conditions and given the tendency to avoid new
subject --- a tendency which is likely to be sensitive to the degree of
newness, it is no longer surprising that readers chose the object analysis for
the critical NP in the inanimate implausible condition.  The presence of the
instrumental analysis did not matter for the first experiment, where the
critical NPs were plausible, and, crucially, inferable --- not so new as to
drive the processor to the object NP analysis.

Disconnectedness theory is not conditioned on discourse status and cannot
simultaneously account for the plausibility effects in the inanimate conditions
of experiment 2 and the lack of garden path effects in the ambiguous conditions
of experiment 1.  It would have to be restated over representations which
distinguish unrelated entities from those which can be related by means of
bridging inferences.

In order to decide between Disconnectedness and Avoid New Subjects, we may be
able to combine results from two very different experiments, using the
following reasoning: While Disconnectedness theory makes predictions for
gap-filling, Avoid New Subjects does not.  To falsify Disconnectedness, one
could show that Disconnectedness acts irreconcilably differently when driving
gap-filling than it does when driving late-closure-effects.  One way of
characterizing a preference is how strong it is compared to another one, in
this case, plausibility.  Recall the experiment of Boland, Tanenhaus, Carlson,
and Garnsey (1989)\nocite{Boland89} discussed on \refpage{Remind}. Using
examples such as \ref{RemindX}, Boland \etal\ argued that gaps are filled
unless implausibility results.

\startxl{RemindX} 
a. Which child did Mark remind them to watch this evening?\\
b. Which movie did Mark remind them to watch this evening?  
\stopx

It follows that for gap-filling decisions, Disconnectedness is not as strong a
factor as Plausibility.  Stowe's second experiment, on the other hand, suggests
that Disconnectedness (or Avoid New Subjects) is sufficiently strong so as to
override Plausibility.  Of course, to be convincing, Stowe's second experiment
must be repeated with materials which completely rule out transitive
(instrumental) readings in the inanimate conditions. For example

\startx
While the cake was baking the oven caught fire.\\
As the plot unfolds the reader is ushered into a world...
\stopx

\subsubsection{Complement Clauses}
\label{sec:CC}

In order to be relevant for the ambiguity in \ref{DList}b, Avoid New Subjects
must be applicable not just to subjects of root clauses but also to embedded
subjects.  It is widely believed that constituents in a sentence tend to be
ordered from given to new.  The statistical tendency to avoid new subjects may
be arising solely as a consequence of the tendency to place new information
toward the end of a sentence and the grammatically-imposed early placement of
subjects.  If this were the case, that is, Avoid New Subjects is a corollary of
Given Before New, then Avoid New Subjects would make no predictions about
subjects of complement clauses, as these are neither at the beginning nor at
the end of sentence/utterance.  In this section, I argue that it is the
grammatical function of subjects, not just their linear placement in the
sentence, that is involved with the avoidance of new information.

When a speaker/writer wishes to express a proposition which involves reference
to an entity not already mentioned in the discourse, s/he must use a {\em
new\/} NP. S/he is quite likely to avoid placing this NP in subject position.
To this end, s/he may use constructions such as passivization, there-insertion,
and clefts.  It is often observed that speakers tend to use structures like
\ref{Mercedes}b in order to avoid structures like
\ref{Mercedes}a.

\startxl{Mercedes}
a. A friend of mine drives a Mercedes.\\
b. I have a friend who drives a Mercedes.
\stopx

The theory of Avoid New Subjects predicts that this sort of effort on behalf of
writers should be evident in both root clauses and complement clauses.  To test
this prediction I conducted another survey of the Penn Treebank.  I compared
the informational status of NPs in subject and non-subject positions in both
root and embedded clauses, as follows.  I defined subject position as `an NP
immediately dominated by S and followed (not necessarily immediately, to allow
for auxiliaries, punctuation, etc.) by a VP.'  I defined non-subject position
as an `an NP either immediately dominated by VP or immediately dominated by S
an not followed (not necessarily immediately) by VP.'  To determine givenness
status, I used a simple heuristic procedure\footnote{I am grateful to Robert
Frank for helpful suggestions regarding this procedure.} to classify an NP into
one of the following categories:{\sc empty-category, pronoun, proper-name,
definite, indefinite, not-classified.} The observed frequencies for the
bracketed Brown corpus are as follows.\footnote{For clarity I only give results
{}from the Brown corpus, but all assertions I make also hold of the Wall Street
Journal corpus.
\refapp{SubjAppendix} contains data for both corpora.}

\mytable{
 \begin{tabular}{|l|rr|rr|}
 \hline
  & \multicolumn{2}{c|}{root clause} &  \multicolumn{2}{c|}{embedded clause}\\
 		          &  subj  & non-subj &  subj & non-subj\\ \hline
 {\sc empty-category}     &      0 &     0    &    50 &    47 \\
 {\sc pronoun}		  &   7580 &   956    &  1800 &   213 \\
 {\sc proper-name}	  &   2838 &   539    &   282 &    53 \\
 {\sc definite}	          &   6686 &  3399    &  1156 &   533 \\
 {\sc indefinite}	  &   4157 &  5269    &   736 &   899 \\
 {\sc not-classified}	  &   3301 &  1516    &   366 &   246 \\ \hline
 {\sc total}              &  24562 & 22679    &  4390 &  1991 \\ \hline
 \end{tabular}}

All {\sc pronoun}s are either pleonastic or evoked --- they are thus fairly
reliable indicators of given (at least non-new) NPs.  The category {\sc
indefinite} contains largely brand-new or inferable NPs, thus being a good
indicator of new information.  Considering {\sc pronoun}s and {\sc indefinite}s
there is a clear effect on grammatical function for both root clauses and
embedded clauses.

\mytable{
\begin{tabular}{|l|rr|rr|}
\hline
 & \multicolumn{2}{c|}{root clause} & \multicolumn{2}{c|}{embedded clause}\\
		  &  subj  & non-subj &  	subj & non-subj\\ \hline
{\sc pronoun}     &   7580 &   956    &  	1800 &   213 \\
{\sc indefinite}  &   4157 &  5269    &  	 736 &   899 \\ \hline
                  & \multicolumn{2}{c|}{$\chi^2 = 3952.2,\; p<0.001$}  
                       & \multicolumn{2}{c|}{$\chi^2 = 839.5,\; p<0.001$} \\ 
\hline 
\end{tabular}}

The prediction of Avoid New Subjects is therefore verified.  

As remarked earlier in this section, when a hearer/reader is faced with an
initial-segment such as

\startxl{HeardAgain}
John has heard the joke...
\stopx

the ambiguity is not exactly between an NP complement analysis versus an
S-complement analysis, but rather between an TR (transitive verb) analysis and
an RC (reduced S-comp\-lement).  It is therefore necessary to verify that Avoid
New Subjects is indeed operating in this RC sub-class of sentential
complements.  A further analysis reveals that this is indeed the case.

\mytable{
\begin{tabular}{|l|rr|rr|}
\hline
 & \multicolumn{2}{c|}{TC} &  \multicolumn{2}{c|}{RC}\\
                        &  subj  & non-subj &  subj & non-subj\\ \hline
{\sc empty-category}    &      0 &     6   &     50 &     41\\
{\sc pronoun}           &    773 &    79   &   1027 &    134\\
{\sc proper-name}       &    201 &    32   &     81 &     21\\
{\sc definite}          &    890 &   351   &    266 &    182\\
{\sc indefinite}        &    617 &   555   &    119 &    344\\
{\sc not-classified}    &    259 &   167   &    107 &     79\\ \hline
{\sc total}             &   2740 &  1190   &   1650 &    801\\ \hline
\end{tabular}}

\mytable{
\begin{tabular}{|l|rr|rr|}
\hline
 & \multicolumn{2}{c|}{TC} &  \multicolumn{2}{c|}{RC}\\
		  &  subj  & non-subj &  subj & non-subj\\ \hline
{\sc pronoun}	  &    773 &    79   &   1027 &    134\\
{\sc indefinite}  &    617 &   555   &    119 &    344\\ \hline
                  & \multicolumn{2}{c|}{$\chi^2 = 332.6,\; p<0.001$} 
                            & \multicolumn{2}{c|}{$\chi^2 = 627.6,\; p<0.001$}
\\ \hline
\end{tabular}}

If anything, Avoid New Subjects has a stronger effect after a zero
complementizer.\footnote{ This is in fact demonstrable: when a writer must
place a new NP in an embedded subject position, s/he tends not to omit the
complementizer.

\mytable{
\begin{tabular}{|l|rr|} \hline
embedded subject  &  TC 		    &   RC \\ \hline
{\sc pronoun}	  & 773                     & 1027 \\
{\sc indefinite}  & 617                     &  119 \\ \hline
                  & \multicolumn{2}{c|}{$\chi^2 = 352.6,\; p<0.001$}  \\ \hline
\end{tabular}}

\vspace{1ex}

This observation provides a tantalizing suggestion that the that-trace effect,
exemplified by \ref{ThatTrace} may in fact have a functional explanation ---
the overt complementizer tends to signal new subjects, and a WH-gap can be
thought of as the most given NP possible.

\startxl{ThatTrace}
\begin{tabbing}
d. * \= \kill
a. \> Who did John say Mary likes?\\
b. \> Who did John say that Mary likes?\\
c. \> Who did John say likes Mary?\\
d. *\> Who did John say that likes Mary?
\end{tabbing}
\stopx

\noindent As given here, the mere tendency for new subjects to be
associated with an overt complementizer falls short of completely accounting
for the categorical that-trace effect.  This issue awaits further research.}

\subsubsection{Unambiguous Structures}
\label{sec:UnambiguousStructures}

The consequences of Avoid New Subjects on unambiguous structures such as
\ref{HKM87bis}

\startxl{HKM87bis}
(TR) The maid disclosed the safe's location within the house to the officer.\\
(TC) The maid disclosed that the safe's location within the house had been 
changed.\\
(RC) The maid disclosed the safe's location within the house had been changed.
\stopx

{}from Holmes \etal\ (1987), and center embedding are remarkably similar to
those of Disconnectedness theory.  When presented out of context,
\ref{HKM87bis} TC requires the reader to accommodate a subject which is new to
the discourse, which the TR form does not require.  The TC form is thus
predicted to present some difficulty.

Avoid New Subjects also predicts a difference between \ref{AnotherRat} and
\ref{AnotherItalian}.  

\startxl{AnotherRat}
The rat that the cat that the dog bit chased died.
\stopx

\startxl{AnotherItalian}
A book that some Italian I've never heard of wrote will be published soon by
MIT press.
\stopx

\ref{AnotherRat} requires the reader to accommodate three new subjects 
simultaneously, whereas \ref{AnotherItalian} requires only two, since `I' is an
evoked entity.  Substituting a new entity for `I' is predicted to render the
sentence harder to process.

\startxl{MoreItalians}
A book that some Italian the teacher has never heard of wrote will be published
soon by MIT press.
\stopx

Also, as with Disconnectedness, changing the locus of the embedding from
subject to complement predicts an amelioration of center embedding difficulty
in \ref{spoil1} (repeated here as \ref{SpoilBis}) as compared with a mixed
subject-object embedding in \ref{SpoilMix} and the doubly subject embedded
\ref{SpoilSubjSubj}.

\startxl{SpoilBis}
John asked the woman that gave the boy that kicked the dog some candy why she's
spoiling him.
\stopx

\startxl{SpoilMix}
John asked the woman that gave the boy that the dog frightened some candy why
she's spoiling him.
\stopx

\startxl{SpoilSubjSubj}
The woman that the boy that the dog frightened is bothering right now is a
friend of John's.
\stopx

Considering Eady and Fodor's results (see \ref{Eady} on \refpage{Eady}), Avoid
New Subjects predicts difficulty when there are many simultaneous new subjects.
The maximum number of simultaneous subjects in \ref{Eady}a through \ref{Eady}d
is 1, 2, 2, and 3, respectively.  This makes the incorrect prediction that the
difference in processing difficulty between \ref{Eady}b and \ref{Eady}c should
be the smaller than the other two differences.

Lastly, both Avoid New Subjects and Disconnectedness fail to account for the
remaining center embedding effects in non-subject embedding.  \ref{GaveKick} is
embedded one level deeper than \ref{SpoilBis}.  The additional level of
embedding exacts a cost in processing difficulty despite its inoffensiveness to
both Disconnectedness and Avoid New Subjects.\footnote{Robert Ladd (p.c.\ 1993)
hypothesizes that difficulties with center-embedded constructions stem from the
unavailability of well-formed prosodic structures.  Consider the following
contrast:\\

\noindent (i) The shirts that the maid Tom can't stand sent to the laundry 
came back in tatters.\\
\noindent (ii) The shirts that the maid, whom Tom can't stand, sent to the 
laundry came back in
             tatters.\\

Ladd argues that the vocabulary of major prosodic breaks (i.e.\ the single item
--- major break, or comma) is not sufficiently rich to indicate nesting of
brackets, or even whether a break denotes a left or right bracket.  In (i), one
could get by with one break, after the entire matrix subject, and the sentence
sounds fine (except, perhaps for an unusually long intonational phrase).  In
(ii), three breaks are necessary (one for each comma, and one at the end of the
matrix subject) so the nesting relations are not properly encoded/recovered.}

\startxl{GaveKick}
John asked the woman that offered the boy that gave the dog that chased the cat
a big kick some candy why she's spoiling him.
\stopx

The residual difficulty of center-embedding constructions is very likely
explained by memory limitations in the syntactic processor.  Bach, Brown and
Marslen-Wilson (1986)\nocite{BachBrownMarslen-Wilson86} compared
center-embedding and crossed-dependency constructions in German and Dutch, as
in \ref{GermanDutch} and found that the center-embedded examples in German were
harder to understand than their crossed-dependency analogs in Dutch.

\startxl{GermanDutch}
German:\\
\ \ \shortex{8}
{\ \ Arnim & hat & Wolfgang & der Lehrerin & die Murmeln & aufr\"aumen 
& helfen & lassen.}
{\ \ Arnim & has & Wolfgang & the teacher  & the marbles & collect up  
& help   & let}
{\ \ Arnim let Wolfgang help the teacher collect up the marbles.} \\[1ex]

Dutch:\\
\ \mbox{\shortex{8}
{\ \ Aad & heeft & Jantje & de lerares  & de knikkers & laten & helpen 
& opruimen.}
{\ \ Aad & has   & Jantje & the teacher & the marbles & let   & help   
& collect up}
{\ \ Aad let Jantje help the teacher collect up the marbles.}}
\stopx

Rambow and Joshi (1993)\nocite{RambowJoshi93} propose a syntactic parsing
automaton based on Tree Adjoining Grammar.  Using the storage mechanism of
their automaton, they define a processing complexity metric based on how many
storage cells a particular parse needs, and how long they are needed for, in
analogy with paying rent for storage space.  This complexity metric is
consistent with the findings of Bach \etal This metric also provides an
interesting predictor of processing difficulty associated with various
word-order variations of complex sentences in German: Rambow and Joshi show
that a range of acceptability judgements can be accounted for.  Applying Joshi
and Rambow's automaton account to pre- and post- verbal center embedding,
\ref{SpoilBis} and \ref{SpoilSubjSubj}, respectively, yields no difference
(Owen Rambow, p.c.\ 1993) --- all that matters is that the dependencies are
nested.  This suggests that center embedding difficulties really do originate
{}from memory limitations in the syntactic processor.  We can conclude that the
difficulties with the classic center embedded sentences such as \ref{RatHole}
is the aggregate of difficulties in two loci: memory requirements in the
syntactic processor, and interpretive effects resulting from subject embedding,
as discussed above. (cf.\ Eady and Fodor 1981).

\startxl{RatHole}
The rat that the cat that the dog bit chased died.
\stopx

\section{Summary}

I have presented two competing theories which account for human performance
patterns on a variety of syntactic constructions.  Disconnectedness theory
assigns a penalty for each constituents which has not been semantically
integrated with the rest of the constituents.  Avoid New Subjects theory
assigns penalty for noun phrases which appear in subject position and introduce
entities which are new to the discourse.  Avoid New Subjects requires no
assumptions about the sentence processing system beyond what is already
necessary for accounting for competence phenomena (namely that people use
subject position to encode given information).  Disconnectedness theory
requires the assumption that the processor prefers to avoid disconnected
analyses, even when the disconnectedness can be eliminated by immediately
forthcoming words in the string.

While these two theories are very different, stated over different domains,
their predictions coincide for much of the available data.  Disconnectedness
theory as defined here is inconsistent with the post-hoc analysis I have
presented for Stowe's second experiment in
\refsec{sec:LCANS} --- i.e.\ it is insensitive to the degree of newness.  
Of course, a direct experiment would be necessary to validate that analysis.
Another area of disagreement between the two theories is in gap-filling.
Disconnectedness theory predicts that other factors being equal, the processor
would prefer to fill a gap if a filler is available.  Avoid New Subjects makes
no predictions with regards to gap-filling.  I have argued at the end of
\refsec{sec:LCANS} that putting together results from different experiments
could provide us with grounds for falsifying Disconnectedness theory.  The
necessary experiments remain for future research.

In the next two chapters, I present a computational framework for modelling the
various aspects of sentence processing.  The aim is to ultimately provide a
means of integrating experimental results and theories regarding a variety of
factors into one consistent picture.

\chapter{Parsing CCG}
\label{ch:parsing}

\begin{figure}[htb]
\centerline{\psfig{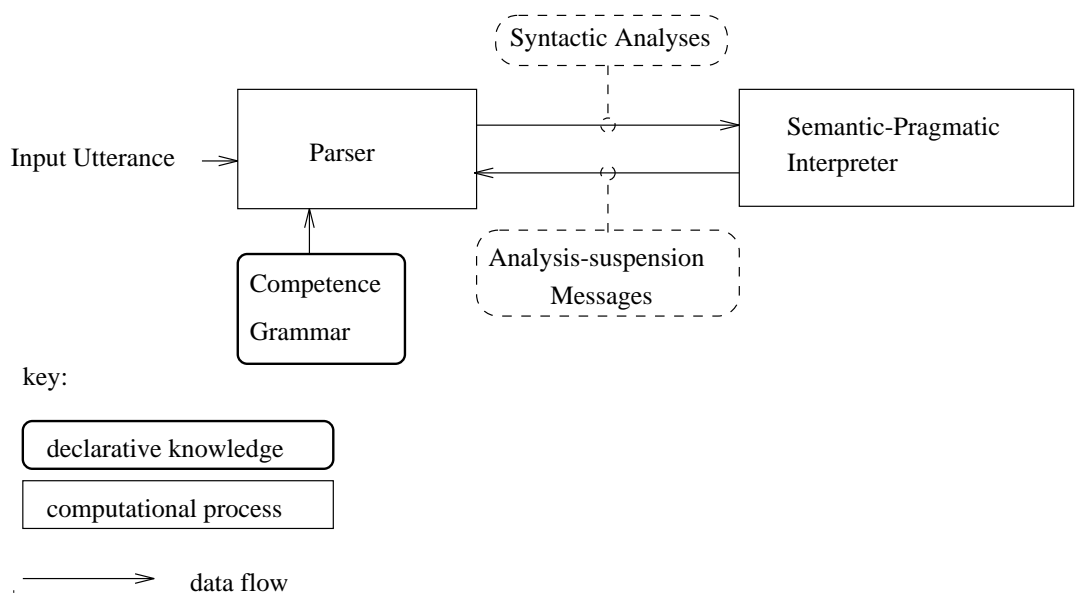}}
\caption{An interactive sentence-processing architecture}
\label{interactiveFig}
\end{figure}

In the preceding chapters I have argued for a view of the sentence processing
architecture where the syntactic processor --- the parser --- proposes
syntactic analyses for the incoming words and the interpreter chooses among
them based on sensibleness.  This is depicted diagrammatically in
\reffig{interactiveFig}.

Having argued for an architecture of this general kind, I now focus on the
specifics of each of the two constituent components, in turn.  In this chapter,
I consider the design of the parsing component, and in the next, I turn to the
interpreter and the integrated system.

\section{Goal}

In the preceding two chapters I have argued that syntactic ambiguity is
resolved according to the interpretations of the available readings.  Given the
virtual immediacy in which a word's contribution to the meaning impinges on
ambiguity resolution decisions, it follows that the parser --- whose task it is
to identify for the interpreter the syntactic relations among the words in the
sentence --- must be performing this task very quickly.  That is, at every
word, the parser identifies {\em all of the grammatically available alternative
analyses\/} and determines for each analysis {\em all of the syntactic
relations which are logically entailed by the competence grammar,} or at least
enough syntactic relations to draw the distinctions necessary for
interpretation.  Crucially, these determinations must not be delayed by the
parser until the end of an utterance or even the end of a clause.

My aim in this chapter is to adhere to the central claim of the dissertation,
that the parser is as simple as logically possible --- all that it encodes is
analyses as defined by the competence grammar.

Steedman (1994)\nocite{GandP} has proposed a processor which he claims is able
to construct sense-semantics in a timely fashion and, in addition, embodies a
very transparent relation to the grammatical competence, the so called {\em
Strict Competence Hypothesis.} In the next section I present Steedman's
architecture.  In the following
\ifprediction five \else six \fi 
sections I consider
\ifprediction five \else six \fi 
different challenges to the simplicity and adequacy of Steedman's proposal and
advocate certain extensions to Steedman's design which promise to address
shortcomings with the original.

\section{Steedman's Proposal}
\label{sec:SteedmansProposal}

How simple can the syntactic processor be?  Steedman (1994)\nocite{GandP}
argues as follows: At the very minimum, the processor needs three components: a
`transparent' representation of the grammar, a method for constructing
constituents by executing steps in the grammar, and a method of resolving
ambiguity.  If the competence grammar is in its traditional form (e.g.\ always
dividing a sentence into a subject and a predicate) then it turns out that this
minimal collection of three components is inadequate to provide the necessary
sense-semantics.  Consider the pair in \ref{FlowerSent}

\startxl{FlowerSent}
a. The doctor sent for the patient arrived.\\ b. The flowers sent for the
patient arrived.
\stopx

While \ref{FlowerSent}a is a garden path, \ref{FlowerSent}b is not.  This is
because the implausibility of the main verb analysis of `the flowers sent' is
detected.  This detection takes place before the sentence is complete.  It
follows that the sense-semantics of the subject is combined the verb before the
entire VP is processed.  If the grammar requires a VP node, however, the
straight-forward interpretation of the minimal model above, wherein the
processor can only combine two constituents when the syntax allows them to
combine, must wait until the VP is finished before the content of the subject
is integrated with the content of the VP.  Steedman argues that the obvious
ways of relaxing this strict rule-to-rule parsing hypothesis, (which he calls
the {\em Strict Competence Hypothesis}) such as adding Earley-style dotted
rules (Earley 1970\nocite{Earley70}) or a top-down parse-stack, complicate the
design of the parser and shifts additional explanatory burden to the theory of
evolution.  Steedman argues that if the grammar does not require an explicit VP
constituent, i.e.\ if it is able to treat `The flowers sent' (where `sent' is
the main verb) as a constituent, strict competence can be restored to the
processor.

Details of Steedman's grammatical theory, Combinatory Categorial Grammar (CCG)
provide an illustration of his claim.  In CCG every constituent has a
grammatical category drawn from a universe of categories as follows.  There is
a finite set of {\em basic categories\/} such as s, np, n, etc. It is given
either as a list of symbols or a space of finite feature structures.  There are
two binary type-forming connectives, / and \? such that if X and Y are
categories then X/Y and X\?Y are also categories.  The set of categories is the
set of basic categories closed under the connectives / and \?.  By convention,
slashes associate to the left, so (s\?np)/np is usually written
s\?np/np. Intuitively, a constituent with category X/Y (or X\?Y) is an X which
is missing a Y to its right (left).  CCG is a lexicalized grammar formalism
which means that the collection of constituent-combination rules is rather
minimal and most of the complexity of the grammar resides in the way individual
words are assigned a category (or a set of categories in case of lexical
ambiguity).  From the description of the `meaning' of the slash connectives,
one expects the combinatory rules in \ref{SimpleCG}:

\newcommand{\rulepair}[2]{\rulepairHelper{\underline{#1}}{#2}}
\newcommand{\rulepairHelper}[2]{
X/Y & Y#1 & \GoesTo & X#1 & \forw #2 & Y#1 & X\?Y & \GoesTo & X#1 & \back #2 }
\newcommand{\MovedLeft}[1]{\(\!\!\!\!\! \mbox{#1}\)}
\startxl{SimpleCG}
\begin{tabular}{|lllll|lllll|} \hline
\multicolumn{5}{|c}{Forward Functional Application} & 
\multicolumn{5}{|c|}{BackwardFunctional Application} \\ \hline
\rulepair{}{} \\ \hline
\end{tabular}
\stopx

By convention, the arrow is in the direction of parsing, not generation.  These
are actually rule schemata, where X and Y are variables which range over
categories.  A rule combines two adjacent constituents whose categories match
its left hand side and creates a new constituent with the category on its
right-hand side.  A particular CCG can stipulate restrictions over the
categories that the variables may take as values.  In addition to the two
so-called {\em functional application\/} rules above, CCGs also includes {\em
functional composition\/} rules such as \mbox{X/Y \ \ Y/Z \ \ \GoesTo \ \ X/Z}.
In the rest of this document, I use the following unified notation for
application and generalized functional composition:

\Center{%
\begin{tabular}{|lllll|lllll|}
\hline
\multicolumn{4}{|c}{\hspace{\fill} Forward Combination \hspace{\fill} rule}  
& \MovedLeft{name} & 
\multicolumn{4}{c}{\hspace{\fill} Backward Combination \hspace{\fill} rule} 
& \MovedLeft{name} \\ 
\hline
\rulepair{}{0} \\
\rulepair{\up Z}{1} \\
\rulepair{\up Z\SUB{1}\up Z\SUB{2}}{2} \\[-1mm]
 & & \multicolumn{1}{c}{$\vdots$} & & & & & \multicolumn{1}{c}{$\vdots$} 
& & \\[-1mm]
\rulepair{\up Z\SUB{1}\ldots\up Z\SUB{n}}{n} \\[-1mm]
 & & \multicolumn{1}{c}{$\vdots$} & & & & & \multicolumn{1}{c}{$\vdots$} & & \\
\hline
\end{tabular}}

In the table above, \up Z stands for either /Z or \?Z.  Underlined regions in a
rule must match.

Aside from the combination rules above, CCG systems often include two other
kinds of rules.  Type raising and Substitution.  Type raising, schematized as

\Center{%
\begin{tabular}{|l|l|}
\hline
Forward Type Raising & Backward Type Raising \\ \hline 
X \ \ \GoesTo \ \ Y/(Y\?X) \ \ \ T\forw & X \ \ \GoesTo \ \ Y\?(Y/X) \ \ \ 
T\back \\ 
\hline
\end{tabular}}

is assumed to apply in the lexicon and is therefore not included as a rule in
the grammar.  The Substitution rule, posited in order to handle parasitic gaps
(Steedman 1987\nocite{CGPG}; Szabolsci 1983\nocite{Szabolsci83}) is

\Center{%
\begin{tabular}{|c|}
\hline
Substitution \\ \hline Y/Z \ \ X\?Y/Z \ \ \GoesTo \ \ X/Z \ \ \ S\back \\\hline
\end{tabular}}

and is also included in the universe of rules.

In addition to the above rules, There is a special rule for coordination which
combines three subconstituents:

\Center{X \ \ coord \ \ X \ \ \GoesTo \ \ X}

A {\em derivation\/} is a tree whose leaves are categories, and whose internal
nodes are valid rule applications.  A string is {\em grammatical\/} just in
case there is a derivation whose frontier is a sequence of categories which are
each in the lexical entry of the corresponding word in the string.  Aside from
determining the syntactic category of a string, a derivation can also assign it
semantics.  One way of achieving this is using combinators (Curry and Feys
1958\nocite{CurryFeys58}; Quine 1966\nocite{Quine66}; Steedman
1990\nocite{GACC}).  A combinatory semantics consists of augmenting each
lexical entry with a semantic object, and each combinatory rule with a semantic
combination recipe.  The lexicon, then, maps a word to a set of pairs
\mbox{$\langle$ syntactic-category : semantic object $\rangle$}.  The semantic
combinations recipes are as follows

\Center{%
\begin{tabular}{|llll|}
\hline
& & & \\[-4mm] X:a \ \ Y:b 		 & \GoesTo & Z:(\cB\SUP{i} a b) & 
\forw i \\ Y:b \ \
X:a 		 & \GoesTo & Z:(\cB\SUP{i} a b) & \back i \\ X:a
& \GoesTo &
Y/(Y\?X):(\cT\ a) & T\forw \\ X:a 		 & \GoesTo &
Y\bkslf(Y/X):(\cT\ a) & T\back
\\ Y/Z:b \ \ X\?Y/Z:a & \GoesTo & X/Z:(\cS\ a b) & S\back \\
\hline 
\multicolumn{4}{c}{\rule{0pt}{.1ex}} \\
\multicolumn{4}{c}{\parbox{238pt}{(Juxtaposition denotes term application.  By
convention, terms associate to the left, so $(x y) z$ is written as $x y z$.)}}
\end{tabular}}

The semantic terms \cB\SUP{i}, \cT, and \cS\ are special symbols, called {\em
combinators.}  They do not carry any semantic content themselves, rather they
encode combinatorial recipes of their arguments, according to the following
equations.

\Center{%
\begin{tabular}{lcl}
\cB\SUP{i} x y\SUB{1} $\cdots$ y\SUB{i+1} & = & x (y\SUB{1} $\cdots$ 
y\SUB{i+1}) \\
\cT\ x y & = & y x \\
\cS\ x y z & = & x z (y z)
\end{tabular}}

By way of an illustration, consider the following unambiguous lexicon

\Center{%
\begin{tabular}{ll}
John & s/(s\?np) : \cT\ {\em j} \hspace{2cm} (notice the application of 
T\forw\ in the lexicon)\\
has & s\?np/(s\?np) : {\em has}\\ 
met & s\?np/np : {\em met}\\ 
Susan & np : {\em s}
\end{tabular}}

The string `John has met Susan' is grammatical since it is possible to derive a
single constituent from it as follows:

\startxl{JohnAndSusan}
\Center{%
\BN{\vspace{-9mm}
   \begin{eqnarray} 
     \mbox{s} & : & \cB^1(\cB^1 (\cT j) h) m\; s \nonumber\\ 
              & = & \cB^1 (\cT j) h (m\; s) \nonumber\\ 
              & = & (\cT j) (h (m\; s)) \nonumber\\ 
              & = & h (m\; s) j\nonumber \end{eqnarray}} 
   {\BN{s/np : $\cB^1 (\cB^1 (\cT j) h) m$} 
       {\BN{s/(s\?np) : $\cB^1(\cT j) h$}
           {\LN{John}{s/(s\?np) : \cT$j$}} 
           {\LN{has}{s\?np/(s\?np) : $h$}}
           {\forw 1}}
       {\LN{met}{s\?np/np : $m$}}
       {\forw 1}} 
  {\LN{Susan}{np:s}}
  {\forw 0}}
\stopx

Notice, however, that there are other derivations for this string, which yield
the same semantic result.  For example,

\startxl{JohnAndMoreSusan}
\Center{%
\BN{\vspace{-9mm}\begin{eqnarray} \mbox{s} & : & \cB^1 (\cT j) (\cB^1 h m) s
\nonumber\\ & = & \cT j (\cB^1 h\; m\; s) \nonumber\\ & = & \cT j (h (m\; s))
\nonumber\\ & = & h (m\; s) j \nonumber \end{eqnarray}} {\BN{s/np : $\cB^1
(\cT j) (\cB^1 h m)$} {\LN{John}{s/(s\?np) : \cT$j$}} {\BN{(s\?np)/np : $\cB^1
h m$} {\LN{has}{s\?np/(s\?np) : $h$}} {\LN{met}{s\?np/np : $m$}} {\forw 1}}
{\forw 1}} {\LN{Susan}{np:s}} {\forw 0}}
\stopx

These analyses makes use of the functional composition rule \forw 1 to
construct the non-traditional constituent `John has met'.  It has been argued
(e.g.\ Dowty 1988\nocite{Dowty88}; Steedman 1990, 1991,
1992\nocite{GACC}\nocite{Steedman91}\nocite{SurfaceStructure}) that such
constituents are necessary for a proper treatment of the syntax of
coordination, WH-dependencies, and sentence-level prosodic structure.  The
reader is referred to these papers for details of the theory of competence.

Steedman's point, then, is that a processor for CCG uses the slash mechanism of
the {\em competence grammar\/} --- the same mechanism which is responsible for
constructing the material between a WH filler and its `gap' and `non-standard'
constituents for coordination --- in order to produces a grammatical
constituent for `the flowers sent' in \ref{FlowerSent} (which repeated here as
\ref{FlowerSentAgain}) whereas a processor for a traditional phrase-structure
grammar would have to use grammar-external devices such as dotted rules to
achieve the same effect. 

\startxl{FlowerSentAgain}
a. The doctor sent for the patient arrived.\\
b. The flowers sent for the patient arrived.
\stopx

\section{Strict Competence and Asynchronous Computation}

Shieber and Johnson (1993)\nocite{ShieberJohnson93} claim that Steedman's
argument that a standard right-branching grammar requires a more complicated
parser rests on an incorrect assumption.  They distinguish two sorts of
computational architectures, synchronous and asynchronous and argue that the
assumption of a synchronous architecture, necessary for Steedman's argument, is
no more likely {\em a priori\/} than that of an asynchronous architecture and
may, in fact, be less likely.  In this section I present Shieber and Johnson's
argument and assess its force.

\subsection{Synchronous and Asynchronous Computation}

Suppose one had to construct a machine to compute the following function of two
numeric arguments $x$ and $y$ \[ f(x,y) = xy + (-y) \] out of components which
perform primitive arithmetic operations.

\begin{figure}
\centerline{\psfig{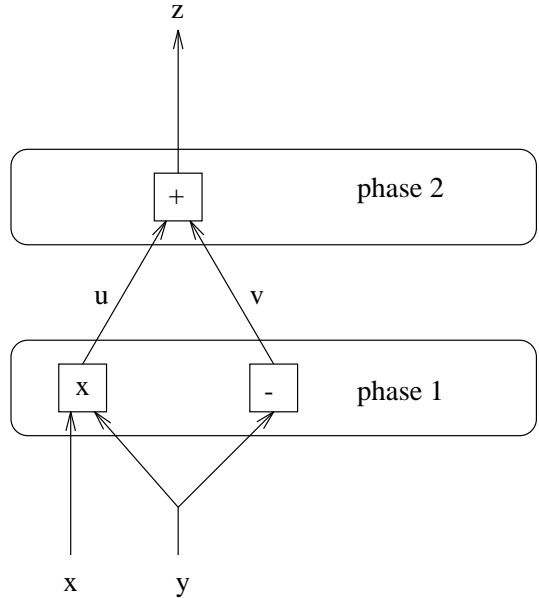}}
\caption{A circuit for computing \protect{$z=xy + (-y)$} 
(from Shieber and Johnson 1993)}
\label{ShieberJohnsonFig}
\end{figure}

One way to do this is to use a two-phase circuit as in
\reffig{ShieberJohnsonFig}.  The first phase computes the intermediate results
$u = x y$ and $v = -y$ and the second phase computes the sum of $u$ and $v$.
The multiplication unit could come in one of two varieties --- synchronous or
asynchronous.  The synchronous variety requires that both of its inputs be
specified before the output is computed.  The asynchronous variety emits an
output as soon as it can --- whenever one of its inputs is zero, it does not
wait for data on the other input before emitting the answer zero on its output.
If the circuit is indeed built from asynchronous components, then a $y$ input
of zero would cause it to emit a $z$ answer of zero without waiting for the
value of $x$.  If one were using asynchronous components but wanted phase-level
synchronization, i.e.\ all of a phase's inputs must be specified before its
output is emitted, one would have to build in additional `restraints' into the
circuit.  Shieber and Johnson argue that Steedman's strict competence
hypothesis precisely imposes phase-level synchronization at the level of the
module --- Steedman requires that the syntactic module make available to the
interpretive module analyses of only {\em complete\/} constituents.  That is,
the interpreter may not {\em see\/} the results of combinations of incomplete
syntactic constituents.  They argue that this phase-level synchronization is
not necessary, and could, in fact make the design of the processor more
complicated, as is the case with the design of phase-synchronous digital
electronic devices.

To illustrate the viability of asynchronous computation they propose a
grammatical formalism which pairs partial parse-trees with partial LF-like
representations (May 1985)\nocite{May85}.  The apparatus uses the same
structure-combining operation for both the construction of grammatical
constituents (including the residue of WH-movement) as well as the construction
of `partial' constituents such as \ref{PartialTree}.

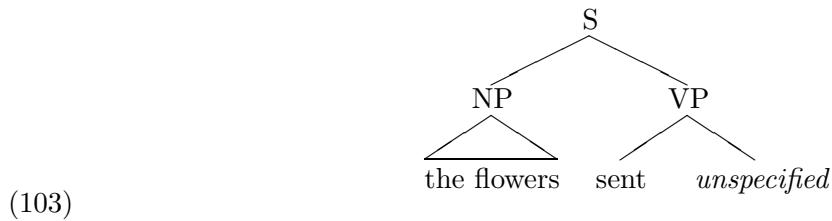
\startxl{PartialTree}
\centerline{
  \begin{picture}(152.15,84.0)(0,0)
   \put(62.175,71.0){\makebox(1,1){\rule[-1ex]{0ex}{3ex}{\normalsize S}}}
   \put(62.175,66.72){\line(-2,-1){37.15}}
   \put(62.175,66.72){\line(2,-1){37.15}}
   \put(25.025,41.0){\makebox(1,1){\rule[-1ex]{0ex}{3ex}{\normalsize NP}}}
   \put(25.025,36.72){\line(-3,-2){25.025}}
   \put(25.025,36.72){\line(3,-2){25.025}}
   \put(0.0,20.037){\line(1,0){50.05}}
   \put(25.025,11.0){\makebox(1,1){\rule[-1ex]{0ex}{3ex}{\normalsize the 
flowers}}}
   \put(99.325,41.0){\makebox(1,1){\rule[-1ex]{0ex}{3ex}{\normalsize VP}}}
   \put(99.325,36.72){\line(-3,-2){25.525}}
   \put(99.325,36.72){\line(3,-2){25.525}}
   \put(73.8,11.0){\makebox(1,1){\rule[-1ex]{0ex}{3ex}{\normalsize sent}}}
   \put(124.85,11.0){\makebox(1,1){\rule[-1ex]{0ex}{3ex}{\normalsize 
		{\em~unspecified}}}}
   \put(0.0,0.0){\makebox(1,1)[l]{\rule[-1ex]{0ex}{3ex}{\normalsize}}}
  \end{picture}
}
\stopx

Such a tree is paired by the formalism with an underspecified logical-form
representation similar to \ref{PartialLF}.

\startxl{PartialLF}
\mbox{({\em unspecified-op.} ... ({\em unspecified-op.} 
(send($\langle$the-flowers$\rangle$,unspecified-object))) ... )}
\stopx

This representation anticipates zero or more sentence-level operators (which
may appear syntactically adjoined to the VP node but move to S at the logical
form level).  The subject of the sending is specified, but the object is not.
An interpreter may look at this structure and opportunistically draw whatever
conclusions it can from the parts that are specified.

The formalism that Shieber and Johnson use is that of Synchronous Tree
Adjoining Grammars (Shieber and Schabes 1990)\nocite{ShieberSchabes90}.  I now
sketch the idea briefly.  The reader is referred to the original papers for
details.  

(Lexicalized) Tree Adjoining Grammar (TAG, Joshi 1985\nocite{Joshi85}; Joshi
and Schabes 1991\nocite{JoshiSchabes91}) is a grammatical formalism where a
grammar associates a finite set of trees with each lexical item and trees can
combine using one of two operations: substitution and adjunction.  One tree
$\beta$ is {\em substituted\/} into another $\alpha$ by simply replacing one
non-terminal symbol at the frontier of $\alpha$ with $\beta$.  Adjunction is
slightly more complex --- a tree $\beta$ is {\em adjoined\/} into $\alpha$ by
excising a subtree of $\alpha$ that is rooted at some nonterminal $X$,
substituting the subtree into $\beta$ at some occurrence of $X$, and then
substituting the new $\beta$ into the original $X$ site in $\alpha$.
Synchronous TAG (no relation to synchronous computation) is a grammatical
formalism for {\em transduction\/} --- the idea is that two TAGs are {\em
synchronized\/} (or coupled) such that operations in one member are reflected
in the other.  Given two TAGs, a synchronous TAG can be defined as a set of
ordered pairs of trees, from the respective grammars.  Within an ordered pair,
nodes in one tree can be paired with corresponding nodes in the other.
Whenever an operation (substitution or adjunction) happens to one node in a
tree, a corresponding operation must happen to the node it is linked to.

\subsection{Evaluation}

Steedman claimed that a processor for a right-branching grammar needs to have a
grammar-external operation for partial combination.  Given that the interpreter
must be able to take advantage of applications of this operations, it must
follow that it too is able to `see' the operation as well.  Shieber and Johnson
have shown that using an asynchronous computational paradigm, it is not
necessary to augment the parser or interpreter with any operations beyond those
allowed by the competence grammar.

The ultimate question --- of whose system is simpler: Steedman's CCG or Shieber
and Johnson's asynchronously computed partial-structure paradigm --- can only
be resolved when they are both extended to provide wide coverage of the
linguistic phenomena, and their precise implementation details are given.

One important aspect which Shieber and Johnson do not address in their paper is
coordination.  CCG provides a uniform mechanism for incremental interpretation
and the constituency necessary for coordination.  For example, CCG assigns the
string `John loves' the grammatical category S/NP. This category can be
coordinated with another of the same type, giving rise to constructions such as
Right Node Raising:

\startxl{RNR}
John loves and Bill hates London.
\stopx

Such an analysis of Right Node Raising is not readily available in Shieber and
Johnson's mechanism, which assigns `John loves' the grammatical category S,
which the {\em grammar\/} cannot distinguish from the category for `John loves
London'.  While there are various approaches possible for extending Shieber and
Johnson's account (e.g.\ by adopting a proposal by Joshi 1992\nocite{Joshi92},
or elaborating on the work of Sag \etal\ 1985\nocite{SagEtal85}) more research
is needed to determine whether the elegance and simplicity of their account
would remain once it is extended to cover coordination.

One potential source of difficulty for CCG is the need for interpretation of
constituents in which not all combinators have been rewritten.  For example,
the interpretation for the main-verb analysis `the flowers sent' is

\startxl{BTflowers}
\cB (\cT(the' flowers')) send'
\stopx

One might suppose that interpreter contains special strategies to cope with
such expressions, but such a move introduced serious complexity.

Another possibility is to redefine the notion of combinator.  Instead of
treating it as a primitive symbol, one could treat is as standing for a
$\lambda$-term.  Given their definitions above, this is straightforward:

\Center{%
\begin{tabular}{lcl}
\cB\SUP{i} & = & $\lambda$ x y\SUB{1} $\cdots$ y\SUB{i+1} \ . 
\ x (y\SUB{1} $\cdots$ y\SUB{i+1}) \\
\cT\ & = & $\lambda$ x y \ . \ y x \\
\cS\ & = & $\lambda$ x y z \ . \ x z (y z)
\end{tabular}}

Given this interpretation, \ref{BTflowers} rewrites and $\beta$-reduces to

\startxl{LambdaFlowers}
$\lambda$ x \ . \ send' x (the' flowers')
\stopx

which is very similar to Shieber and Johnson's representation in
\ref{PartialLF}.

I conclude that Shieber and Johnson have made a compelling case against
Steedman's claim that CCG clearly gives rise to a simpler processor than any
system based on a right-branching grammar.  This debate, therefore, is far from
settled.  For now either approach is viable, so I use one, CCG or the remainder
of this document.

A note is in order about the choice of a formalism for semantic representation.
The obvious formalism is that of an applicative system such as combinators or
$\lambda$-terms, described above.  Such formalisms require the application of
zero or more reduction rules after each syntactic combination.  It is possible
to eliminate the necessity for reduction rules using `pre-compilation'.  The
idea, described in Pereira and Shieber (1987)\nocite{PereiraShieber87} is to
replace the simple category symbol with a Prolog term which, in addition to
encoding the usual syntactic features such as number and gender, encodes the
predicate argument structure as well.  See Pareschi and Steedman
(1987)\nocite{lazy} and Park (1992)\nocite{Park92} for discussions of
applications of this approach to CCG.  Moore (1989)\nocite{Moore89} argues that
reduction rules cannot be eliminated altogether --- the problem is that
unification-based approximations of the $\lambda$-calculus do not treat
separate $\lambda$-bindings of the same variable as distinct.  A clear
illustration of this problem arises in subject coordination, as in

\startxl{VPcoord}
John and Bill walk.
\stopx

If the predicate `walk' is treated as in\footnote{Pereira and Shieber (1987)
introduce the infix operator \WhatsUp\ as a notation to encode
$\lambda$-terms. So a term such as
\[\lambda x . \lambda y . fxy\]
would be encoded in Prolog as
\[ \mbox{{\tt X\WhatsUp(Y\WhatsUp f(X,Y))}.} \]}

\startxl{WalkPred}
\verb+X^walk(X)+
\stopx

and the coordinate subject `John and Bill' is given a generalized-quantifier
treatment (i.e.  typeraised) as in

\startxl{JandB}
\verb+(S^P)^((john^P)&(bill^P))+
\stopx

then the two `copies' of the predicate (bound by the variable \verb+P+) will
fail to serve as independent $\lambda$-binders of X.  The unification step
necessary to give the result

\startxl{JBResult}
\verb+(john^walk(john)) & (bill^walk(bill))+
\stopx

will be blocked.  Park (1992) considered a collection of coordinate structures
and argued that for each, it is possible to construct a coordination rule (in
his case, a separate CCG lexical entry for the word `and') which provides the
correct logical form using only unification.  But in some cases, Park's
resulting logical form is not the intuitively obvious, simplest one, but
rather, a more complicated form which is truth-conditionally equivalent.  For
example, the logical form assigned to \ref{ParkS}a is
\ref{ParkS}b, not \ref{ParkS}c.  

\startxl{ParkS}
\begin{tabbing}
a. \=A farmer and every senator talk.\\
b. \>\((\exists x \: .\: (\mbox{\verb+Farmer+}(x) \wedge (\exists y \: 
.\: y = x \wedge 
      \mbox{\verb+Talk+}(y)))) \wedge\)\\
   \>\((\forall z \: .\: \mbox{\verb+Senator+}(z) \Rightarrow 
              (\exists y \: .\: y = z \wedge \mbox{\verb+Talk+}(y))) \) \\
c.  \( \exists x \: .\: (\mbox{\verb+Farmer+}(x) \wedge \mbox{\verb+Talk+}(x)) 
\wedge
             (\forall z \: .\: \mbox{\verb+Senator+}(z) \Rightarrow 
\mbox{\verb+Talk+}(z)) \) 
\end{tabbing}
\stopx

Park suggests that a post-semantic process (within the interpreter) could
massage forms like \ref{ParkS}b into forms like \ref{ParkS}c.  But these
manipulations of logical form, while they do preserve the entailments of the
meaning, might be too heavy-handed for other, more pragmatic aspects of
meaning.  I conclude from Park's results that Moore's observations are quite
accurate: attempting to simulate $\beta$-reduction using term unification
results in rather contorted and unnatural semantic representations.  Phenomena
such as coordination indeed do require interleaving applications of the
combinatory rules of grammar with applications of semantic reduction rules.

The Davidsonian approach to semantic representation is somewhat similar to a
unification-based approach to semantics in CCG.  It too is incapable of an
elegant treatment of many coordinate structures (e.g.\ \ref{ParkS}a).  But it
is quite straight-forward to extend it to allow one.  The idea is to move to a
representation which separately enumerates each argument to a predicate.  Thus
\ref{NeoDav}a.  would have \ref{NeoDav}c as its representation, instead of 
\ref{NeoDav}b.

\startxl{NeoDav}
a. Most students prefer denim.\\
b. [most(X), student(X), tns(E, present), prefer(E,X,Y), denim(Y)]\\
c. [most(X), student(X), tns(E,present), subj(E,X), obj(E,Y), denim(Y)]
\stopx

A coordinate structure such as \ref{StudProf}a would have the semantic analysis
in \ref{StudProf}b.

\startxl{StudProf}
\begin{tabbing}
a. Most students and some professors prefer denim.\\
b. [\=most(U), student(U), some(V), professor(V), and(U,V,X), tns(E,present),\\
    \> subj(E,X), obj(E,Y), denim(Y)]
\end{tabbing}
\stopx

In the current implementation, I use the traditional Davidsonian approach
mostly for ease of readability.  If coordination were to become important to
the work, it would be straight-forward to map the system to the newer
representation.

\section{Identifying Ungrammaticality}
\label{sec:Ungrammaticality}

The minimum conceivable sentence processor contains a representation of the
grammar, a nondeterministic algorithm for applying the rules of the grammar,
and an ambiguity resolution mechanism which it is claimed here is solely based
on sensibleness of the available analyses.  Consider how such a processor would
cope with the unremarkable sentence in \ref{Insults}.

\startxl{Insults}
The insults the new students shouted at the teacher were appalling.
\stopx

The word `insults' has two categories, a plural noun and a finite transitive
verb.  The noun analysis can combine with the determiner to its left. The verb
analysis cannot.  Should the processor abandon the verb reading?  As external
observers we can examine the grammar of English and conclude that the verb
analysis is doomed --- we know that salvation cannot arrive later in the string
in the form of a category such as s\?(np/n)\?(s\?np/np).  But the processor
cannot know this, since it does not contain pre-compiled knowledge about the
grammar.  The processor cannot automatically prefer a combined analysis to an
uncombined analysis, as this would constitute a structural preference strategy
which would have wide ranging and bizarre predictions. For example, in
\ref{SexLess} when the word `Chris' is encountered, the processor would prefer
to coordinate the two NPs Sandy and Chris because that would yield a single
constituent for the whole string.\footnote{Another problem with preferring a
combined to an uncombined analysis incorrectly predicts difficulties with the
string `Which house did John paint a picture of?' This will be discussed in
\refsec{sec:PictureNouns}.}

\startxl{SexLess}
Kim likes Sandy and Chris likes Dana.
\stopx

The only available recourse given the minimal processor is an account by which
the interpreter is able to discard the verb analysis of `insults'.  But this is
rather unlikely: There is no {\em a priori} reason to expect that an uncombined
determiner should present a problem for an interpreter, thus imposing a penalty
on the verb analysis.  In fact, there are languages where determiners (e.g.\
deictics, quantifiers) are routinely kept uncombined for many words until their
head nouns are processed. For example, in Korean the structure of a noun-phrase
is

\begin{quote}
Determiner Relative-Clause Noun
\end{quote}

The noun analysis of `insults', on the other hand, does incur penalties when
the subsequent words arrive and require a restrictive relative-clause analysis
which entails a complex process accommodating the resulting noun-phrase out of
context.  These same words pose no problems for the verb analysis --- the verb
phrase ``insults the new students'' is constructed.  Given the preference for
avoiding complex accommodation processes, one would expect the interpreter to
discard the noun analysis, leading to a garden path effect at the
disambiguating word `shouted'.  This is clearly wrong.  The sentence
\ref{Insults} causes no conscious processing difficulty.

Whatever solution is provided for eliminating inappropriate analyses (e.g.\ the
verb analysis for `insults' in \ref{Insults}) it must operate rather quickly
and ruthlessly, otherwise the number of surviving ungrammatical analyses
becomes unmanageable.

The problem here is by no means a new one.  The minimal design is a classic
bottom up parser which, online, can determine which analyses are possible, but
not which are impossible.  Ungrammaticality information is only available at
the end of the string when no analysis contains one category which spans entire
input.  Many parsing techniques have been proposed that address this problem:
LR tables (see \refsec{sec:ShieberPereira}) are pre-compiled `guides' to a
parse stack which identify viable sequences of stack elements and implicitly
encode how these elements will be ultimately combined.  The Earley parser
(Earley 1970\nocite{Earley70}) constructs this sort of information online using
annotations on the rules of the grammar to encode which constituents have been
seen and which are expected.  Marcus's parser (Marcus 1980\nocite{Marcus80},
see \refsec{sec:parsifal}) contains rules which explicitly {\em diagnose\/}
which available syntactic analysis should be followed.

To address this problem I propose to augment the syntactic parsing module and
interpretation module with a third module --- an unviable state filter.  There
are three issues pertaining to the design of this filter.

\begin{enumerate}
\item Should it operate as a categorical filter ruling out most
(or all) ungrammatical analyses, or should it have graded judgement, rating
certain analyses better than others?
\item Should it be conceived of as innate, of biological standing equal
to the other two modules, or should it be conceived of as a `skill' which an
experienced language user acquires for discriminating grammatically viable
analyses from ones that are doomed.
\item Should this module be placed before the syntactic processor, mediating
lexical access by performing a first-cut disambiguation process over the
available grammatical categories, or should this module operate on the output
of the syntactic processor, discarding unviable category buffer
configurations?  
\end{enumerate}

Implementing the filter as rating among available analyses can be thought of as
a way of importing structurally/lexically based ambiguity resolution
preferences.  For example, suppose the processor rates a complementizer
analysis of the word `that' when it follows a noun higher than it rates the
relativizer analysis.  The expectation then is indistinguishable from that of
Minimal Attachment in examples like

\startxl{Psycho2}
The psychologist told the wife that he was having trouble with...
\stopx

Similarly for ambiguous words like `raced' in 

\startxl{RaceyExample}
The horse raced past the barn...
\stopx

As has been argued in the preceding chapters, there is little evidence for
structurally based preferences, so a categorical filter that evaluates each
analysis on its own without regard to its competitors is preferable.

As for the second choice, it is clear that an innate account of this filter is
evolutionarily unparsimonious --- if this element is necessary for language
communication then a grammar could not have evolved without it, nor could the
filter have evolved without the grammar.  An empiricist account of the filter
as a skill is rather plausible: When a child begins acquiring language, the
filter is totally permissive, allowing all analyses, even those of a determiner
followed by a verb.  At this stage, the proliferating candidate analyses
usually quickly overwhelm the processor's ability to keep track of them.
Consequently only short utterances are properly understood.  The child observes
that a buffer such as \mbox{[the:DET insults:VERB]} {\em never\/} gives rise to
valid utterances and learns to filter it out.  Gradually this filter is refined
to its observed sophistication in adult listeners.

The third choice concerns the placement of the filter in the rest of the
system.  Placing the filter, as proposed by Steedman (1994)\nocite{GandP},
between the lexicon and the syntactic processor allows one to exploit much
recent work in automatic part-of-speech labeling of words. Church
(1988)\nocite{Church88} has shown that is very easy to train a part-of-speech
tagger on a tagged corpus to achieve accuracy better than 90\% on unseen text.
There has been much recent work on improving the accuracy of such taggers,
and/or reducing the volume of training materials necessary. (see Brill
1992\nocite{Brill92} and references therein.)  Such taggers are sensitive to
only a small portion of the syntactic context in which a word appears ---
usually a window of a few words to either side of it.  In many taggers it is
possible to adjust a parameter called the precision-recall tradeoff.  When {\em
precision\/} is high, the tagger is likely to find few incorrect categories for
a word.  When {\em recall\/} is high, the tagger is likely to miss few correct
categories (but it may increase the number of incorrect parts-of-speech it
guesses for each word).  It is quite plausible that an excellent-recall
moderate-precision part-of-speech tagger mediates lexical access.  I am aware
of no comparable existing work for automatically training a filter which
discriminates viable bottom-up buffer states.  But given the fact that an
unviable buffer state {\em never\/} results in a grammatical analysis for a
string (or at least an analysis which does not require correction on the part
of the hearer) whereas every viable buffer state does eventually give rise to a
grammatical sentence, and given the fact that the space of viable buffer states
is quite small and regularly structured, it is plausible that such a filter can
be trained by observation of successful and unsuccessful buffer states.

Either placement of the filter is therefore viable.  In the next two sections I
consider a two further problems to the filterless minimal architecture. These
problem are resolved using a filter placed between the syntactic processor and
the interpreter.

Here is a sketch of an algorithm which could be used to carry out the
acquisition of this skill of identifying viable buffers.  After each word, for
each parser state, record the sequence of categories in that state's buffer.
At the end of a grammatical string, for each state in the correct analysis, go
back and add a $+$ mark to that state's buffer.  For each state in each
analysis which did not turn out to be the correct one, add a $-$ mark.  After
some training, the resulting collection of marks can be used to implement the
viable buffer criterion as follows:

\begin{itemize}
\item If a particular buffer configuration contains at least one $+$ mark then
it is viable.  
\item  If the buffer configuration contains no $+$ marks, and more $-$ marks 
than some threshold, then it is unviable.
\item  If the buffer configuration contains no $+$ marks, and fewer $-$ 
marks than the threshold then not enough information is available.  In the
absence of definitive information, accept the buffer, thus trading efficiency
for completeness.
\end{itemize}

Note that this algorithm considers each buffer state individually --- the
viability of a buffer is independent of other competing analyses.  As given
above, the algorithm is inefficient, maybe even impractical in that it requires
potentially unboundedly long buffer configurations be stored and retrieved.
But as will be seen in the next three sections, the parser, in practice will
construct very short buffers, rarely exceeding three constituents.
Furthermore, it may well turn out, I suspect, that it is sufficient to consider
only the right-most two or three constituents in a long buffer for the purposes
of buffer viability.  Finally, there is the issue of how many distinct
categories must be kept track of.  Considering current CCG analyses of English,
the collection of relevant categories is likely to turn out to be quite small.
For Dutch, which allows verb clusters to form constituents (see Steedman
1985\nocite{Steedman85}), some additional bit of cleverness may be necessary to
make the theoretically infinite space of categories manageable.  Empirical
investigation of particular induction strategies for the viable buffer
criterion await a broad coverage CCG for English.

\section{Shift-Reduce Conflicts}
\label{sec:PictureNouns}

A bottom-up parser for CCG encounters three kinds of nondeterminism.

\begin{description} 

\item[categorial ambiguity] A word may have more than one part of speech (e.g.\
`rose' is either n or s\?np) or even for the same part of speech, a word may
have more than one combinatory potential, (e.g.\ `raced' is either s\?np or
n\?n/pp).  In LR parsing parlance this is can be thought of as a shift-shift
conflict.

\item[how to combine constituents] constituents in the buffer may combine in
more than one way.  One example is PP attachment: `Chris tickled the dog with
the feather'. This is a reduce-reduce conflict.

\item[whether to combine constituents]  Consider the string
\startxl{WhichHouse}
Which house did you paint a picture of?
\stopx
After the word `paint' the relevant buffer state is

\startxl{WhichHouseBuff}
\BN[0ex]
   {}
   {\LN{Which house}{q/(s$_{+inv}$/np)}}
   {\LN{did you paint}{s$_{+inv}$/np}}
   {}
\stopx

Combining the two constituents is a valid move.  It yields an analysis wherein
it was a house that was painted, not something else.  The combined analysis
cannot be continued grammatically by another NP (a picture).  This is obviously
not the appropriate move here.  The two constituents must remain uncombined
until the end of the string, as in

\startxl{NiceHouse}
\BN{q}
   {\LN{Which house}{q/(s$_{+inv}$/np)}}
   {\BN{s$_{+inv}$/np}
       {\BN{s$_{+inv}$/pp}
           {\BN{s$_{+inv}$/n}
               {\LN{did you paint}{s$_{+inv}$/np}}
               {\LN{a}{np/n}}
               {\forw 1}}
           {\LN{picture}{n/pp}}
           {\forw 1}}
       {\LN{of}{pp/np}}
       {\forw 1}}            
   {\forw 0}
\stopx

This is a shift-reduce conflict.
\end{description}

The first two ambiguities were the topic of the preceding chapters.
Shift-reduce ambiguity is a subject of serious concern since it potentially
applies to {\em every\/} combination.

One may opt to treat this latter ambiguity as any other --- pursue both
analyses in parallel and let the interpreter work it out.  While this sort of
solution might work for ordinary phrase structure grammars, it is impractical
for CCG because of CCG's associativity of derivation.  Recall that CCG's rule
of functional composition, \forw 1 can give rise to multiple equivalent
analyses, as in \ref{JohnAndSusan} and \ref{JohnAndMoreSusan} on
\refpage{JohnAndSusan}.  This derivational ambiguity\footnote{This has been
called `spurious ambiguity' (Wittenburg 1986)\nocite{Wittenburg86}.  Although
it has been pointed out that this ambiguity of CCGs is necessary on linguistic
grounds (see
\refsec{sec:SteedmansProposal}).} proliferates very quickly.  For example the
string in \ref{Catalan7} has 132 truth-conditionally equivalent CCG analyses.

\startxl{Catalan7}
\TN[0ex]{}
  {\BN[0ex]{}
   {\LN{John}{s/(s\?np)}}
   {\LN{was}{s\?np/(s\?np)}}
   {}}
  {\BN[0ex]{}
   {\LN{thinking}{s\?np/s'}}
   {\LN{that}{s'/s}}
   {}}
  {\TN[0ex]{}
   {\LN{Bill}{s/(s\?np)}}
   {\LN{had}{s\?np/(s\?np)}}
   {\LN{left}{s\?np}}
   {}}
  {}
\stopx

In general, for sequences of functional compositions (\forw 1) the degree of
this ambiguity grows as the Catalan series, that is, roughly exponentially.

\begin{eqnarray*}
\mbox{Catalan}(1) & = & 1 \\
\mbox{Catalan}(n) & = &  \sum_{0<i<n}^{} \mbox{Catalan}(i)\mbox{Catalan}(n-i) 
\end{eqnarray*}

In \refsec{sec:Revealing} I describe a way for the parser to cope with this
proliferation of equivalent analyses by keeping track of one representative
{}from each (truth-conditional) equivalence class of analyses.  The processor
can therefore pursue only the maximally left-branching analysis, ignoring the
possibility that two constituents may remain uncombined.  But the local
ambiguity in \ref{WhichHouseBuff} affects truth conditions --- it is either a
house that was painted, or a picture.  So that example requires special
treatment.  The processor must know to distinguish the uncombined analysis and
pursue it in this case.  One may argue that the question of whether to leave
`which house' and `did you paint' uncombined can be easily resolved by waiting
for the very next word for disambiguating information.  But this is not always
possible --- sometimes syntactic disambiguating information is delayed for many
words, as in
\ref{HeavyPictures}.

\startxl{HeavyPictures}
\begin{tabbing}
a. \=Here is the cathedral that John drew, and Bill bought, three beautiful\\
   \> charcoal sketches of.\\
b. \>Which of his daughters was Percival planning to donate to the university\\
   \>an extravagant portrait of?
\end{tabbing}
\stopx

In these examples, it is clear that interpretation determines how the local
ambiguity is resolved.  The parser therefore must present the interpreter with
both analyses.  On what basis, then, can the parser know to make the
interpreter aware of the uncombined analysis in this case, but not to bother
the interpreter with the many other truth-conditionally irrelevant uncombined
analyses?  The viable-buffer filter discussed in
\refsec{sec:Ungrammaticality} offers a solution.  Placing this filter between 
the parser and the interpreter allows the possibility for distinguishing
relevant non-reductions, which, in the case of picture nouns (e.g.\
\ref{WhichHouseBuff}) are identifiable by a sequence of categories of the form
\mbox{[X/(s/np), s/np]}.  Placing the filter between the lexicon and the parser
does not immediately propose such a solution.

Notte that allowing the WH-filler and the gap-containing constituent not to
combine precisely implements the idea argued for in
\refsec{sec:DisConsequences} that the locus of filled gap effects is in the
interpreter, not the parser.

\section{Heavy Shift and Incremental Interpretation}
\label{sec:HeavyIncremental}

Another challenge to the filterless architecture arises from the interaction of
heavy NP shift and referential processes. The Strict Competence Hypothesis
(\refsec{sec:SteedmansProposal}) taken together with the usual assumption of
Compositionality --- that combinations in the syntactic domain are are mapped
to combination in the semantic domain --- predicts that the interpreter may not
become aware of combination of semantic constituents before parser performs the
corresponding syntactic combination. Steedman uses this reasoning
(\refsec{sec:SteedmansProposal}) to argue against a grammar which requires a VP
node.  But does CCG provide sufficiently incremental analyses so as to overcome
every instance of this problem?  The places to look for an answer is where CCG
does not provide a word-by-word left-branching analysis.  One such place is
around the `canonical' position of heavy-shifted arguments.

\ref{Heavy1}a, exemplifies heavy NP shift.  Once one of a verb's arguments is 
heavy-shifted, it is ungrammatical to `move' its other arguments, as shown by
the ungrammaticality of `WH-movement' \ref{Heavy1}b. and right-node-raising in
\ref{Heavy1}c.  Note that multiple right-node-raising is not impossible in
general, as \ref{Heavy1}d shows (the latter is from Abbott
1976)\nocite{Abbott76}.

\startxl{Heavy1}
\begin{tabbing}
a. * \= \kill
a.   \> The bird found in its nest a nice juicy worm.\\
b. * \> What did the bird find in \UL\ a nice juicy worm?\\
c. * \> The bird found in, and its mate found near, the nest, 
some nice juicy worms.\\
d.   \> I promised, but you actually gave, a pink Cadillac to 
Billy Schwartz.
\end{tabbing}
\stopx

In order to rule out \ref{Heavy1}b and c, CCG must delay the combination of
`found' and `in' until the entire PP `in its nest' is constructed.  If we can
show that before the PP is fully processed, the processor is nevertheless aware
of the combination of `found' and `in' then we have shown that even CCG fails
to provide sufficiently incremental analyses.

Evidence for the detection of the heavy NP shift before the PP is fully
processed is provided by the lack of garden path in \ref{Bird2}a as compared to
\ref{Bird2}b.

\startxl{Bird2}
a. The bird found in its nest died.\\
b. The horse raced past the barn fell.
\stopx

In \ref{Bird2}a, when the processor encounters `found' it has two analyses ---
reduced relative clause or main verb.  Out of context, there is no mutually
established `background', so the reduced relative analysis requires the
accommodation of a fairly complex set of presuppositions, as discussed in
\refsec{sec:WeaklyInteractive}.  The main verb analysis has no problems so
far. But the next word, a preposition does present a problem for the main verb
analysis: the verb `find' is obligatorily transitive, so heavy NP shift must be
assumed.  The construction of heavy NP shift is felicitous when the material
which mediates the verb and the shifted argument is backgrounded --- given
information.  Out of context, heavy shift is therefore not felicitous, so the
main verb reading also carries a penalty.  Faced with two imperfect analyses,
the processor has no basis for preferring one over the other, so it keeps them
both, leading to the acceptability of either continuation --- \ref{Heavy1}a and
\ref{Bird2}a.  When the ambiguous verb is potentially intransitive, e.g.\
`raced' in \ref{Bird2}b, encountering a preposition does not present any
difficulties for the main verb analysis, so the processor decides to discard
the reduced relative clause analysis in favor of the main verb analysis,
leading to the garden path in
\ref{Bird2}b.

Crucially, the processor is able to detect the inevitability of heavy NP shift
for the main verb analysis of \ref{Heavy1}a before the PP is fully processed.
Were the processor to wait until the end of the PP to resolve the ambiguity, it
would surely be able to avoid the garden path in \ref{Bird2}b.\footnote{One
possible attempt to salvage the minimal account is to argue that the processor
does not actually determine that heavy shift is unavoidable in the main-verb
analysis of `the bird found in...' but rather that the processor merely notices
that there are two constituents which it cannot yet combine.  In such cases,
the processor proceeds cautiously, not discarding competing analyses (i.e.\ the
reduced relative clause analysis).

To counter this argument one could make the following observation.  While both
analyses of `found' are maintained when the sentence is presented out of
context, there are contexts which can cause the processor to make a commitment
before the end of the PP.  The relevant case here is a context which makes
heavy shift felicitous, as in the question (i).

(i)  What did the bird find in the nest?

\noindent The response in (ii) is a garden path. 

(ii) The bird found in the nest died.

A theory in which the processor does not discard infelicitous analyses (e.g.\
an unnecessary restrictive relative clause) would fail to predict the garden
path in (ii).  }

Placing a viable-buffer filter between the parser and the interpreter can
provide the necessary mechanism for identifying the unavoidable heavy NP shift
in \ref{Heavy1}a.  In the same way that such a device would learn the
inevitable failure of certain buffer configurations, it could also learn the
inevitability of the heavy shift construction which the parser will find.  The
current implementation of this mechanism is presented in
\refsec{sec:admissible}.

\ifprediction {} \else
\section{Category and Subcategory}
\label{sec:subcat}

The account presented above assumes penalties for heavy shift that is
infelicitous in context, and for accommodating a complex NP.  While it performs
correctly with obligatorily transitive verbs such as `found' and obligatorily
intransitive verbs such as `race' (when the subject is `the horse') consider
what would happen with a verb like `read' which has both of these
subcategorizations (when the subject is an animate, such as `poet').

\startxl{dung}
a. The poet read in the garden stank.\\
b. The poet read in the garden a lengthy article about Canadian earthworms.
\stopx

In \ref{dung}a, the complex NP accommodation penalty correctly excludes the
reduced relative analysis, resulting in a garden path.  But in \ref{dung}b,
which is presumably not a garden path sentence,\footnote{This claim must, of
course, be empirically verified.} does the processor described so far correctly
find the heavy NP shift analysis?  Given that the reduced-relative analysis is
discarded, one would expect the main-verb analysis to persist.  But CCG has
three separate analyses for `read': a reduced relative, an intransitive main
verb, and a transitive main-verb.  When the preposition `in' is encountered,
the reduced-relative analysis incurs a penalty for requiring accommodation of a
complex NP; the transitive main-verb analysis incurs a penalty for heavy shift
out of context; and the intransitive main-verb analysis incurs no penalties.
The latter therefore remains, predicting a garden path when the unexpected
heavy-shifted object is encountered after the PP `in the garden'.  This problem
arises, of course, because of the lexicalized nature of CCG: every combinatory
potential of a word is treated as a separate lexical entry.  In other words,
CCG does not distinguish small differences between categories (e.g.\
subcategorization) from major differences (e.g.\ main verb versus reduced
relative clause).  What is necessary is a process for treating the two
main-verb analyses as essentially the same analysis, delaying the distinction
between the intransitive and transitive subcategories until the last moment ---
i.e.\ until either an NP argument is encountered, or the end of the sentence is
reached.

One way of implementing this is to admit a certain form of disjunction into the
representation of a CCG derivation, thereby `sharing structure' between two
similar analyses.  An optionally transitive verb would therefore be represented
as

\startxl{OptionallyTransitive}
$\{1,2\}$ : s\?np(/$_1$np)$_2$
\stopx

which means ``This is a disjunction of two possibilities, labeled 1 and 2.  The
category is either s\?np/np (possibility 1) or s\?np (possibility 2) --- where
the material in the subscripted parentheses is omitted.'' Extending this
example from a category to a derivation, the main-verb analysis of `The poet
read' would thus be

\startxl{StructureSharing}
$\{1,2\}$ : 
\BN{s(/$_1$np)$_2$}
   {\LN{the poet}{s/(s\?np)}}
   {\LN{read}{s\?np(/$_1$np)$_2$}}
   {\forw ($1_1 \vee 0_2$)}
\stopx

This move is inconsistent with the strict competence hypothesis
\refsec{sec:SteedmansProposal} wherein the processor represents nothing but
constituents as defined by the grammar, since the combinatory rules say nothing
about disjunction.  One may adopt a somewhat weaker competence hypothesis,
(Bresnan 82)\nocite{MRGR} wherein the only operations that a parser performs
are the rules of the grammar.  A liberal interpretation of this hypothesis
would allow a composite rule such as {\forw ($1_1 \vee 0_2$)} to apply to a
composite category such as \ref{OptionallyTransitive} --- since the composite
rule is nothing more than a pair of original rules.

It may be possible to avoid the introduction of `structure sharing' by shifting
the burden of distinguishing small differences (subcategories) from large
differences (major part of speech) to the analysis filtering module.  The idea
is to instead of discarding any analysis which is assigned more penalties than
some other analysis, the processor would maintain high-penalty analysis if they
are `sufficiently similar' to some other low-penalty analysis.  This move would
complicate the design of the discarding procedure, requiring it not only to
compare penalty levels but also similarity levels among the various competing
analyses.  It is not clear at this time whether such a move is viable: the
interpreter may only inspect the sense semantics of analyses to judge their
similarity, not their syntactic representation.

Suppose it turns out that processing \ref{dung}b is just slightly harder than
processing a sentence which is similar, but whose verb is obligatorily
transitive (e.g.\ `found' in \ref{Heavy1}a). Then it would be possible to
maintain CCG's distinctions, calling both \ref{dung}a and \ref{dung}b garden
paths.  The way to account for the relatively benign nature of the garden path
in \ref{dung}b is by appeal to the recovery process which follows detection of
a garden-path.  Current theories of recovery (Pritchett
1987\nocite{Pritchett87}, 1992\nocite{Pritchett92}; Gorrell
1993\nocite{Gorrell93}; Inoue and Fodor 1993 \nocite{InoueFodor93}) account for
the ease of recovery by the similarity of the processor's analysis and the
ultimately correct one.  While they diverge on how to define similarity of
analyses, all theories mentioned above converge in entailing that \ref{dung}a
is less similar to the intransitive main-verb analysis than \ref{dung}b is.  In
the current project, I do not consider the process of recovery.  I consider any
ambiguity-related processing difficulty to be a garden path.

In the current implementation, described in \refch{ch:implementation}, none of
the three options above is incorporated, so the system does not predict the
acceptability of
\ref{dung}b. 

\fi 

\section{Coping with Equivalent Derivations}
\label{sec:Revealing}

In this section I address the problem of proliferating equivalent analyses
stemming from CCG's associativity of combination, as introduced in
\refsec{sec:PictureNouns}.  I first examine existing proposals for coping with
this sort of ambiguity.  Combining ingredients from two of the proposals ---
Pareschi and Steedman's (1987)\nocite{lazy} idea of lazy parsing, with Hepple's
(1991)\nocite{Hepple91} normal form construction --- I then introduce a new
parsing system which addresses shortcomings in its predecessors.

\subsection{Evaluation Criteria for a Parser}

In light of the discussion earlier in this chapter, any algorithm which is to
serve as an adequate parser, must satisfy the following desiderata.

\begin{description}

\item[soundness] All parser outputs must be consistent with the grammar and
the input string.

\item[completeness] Given a grammar and a string, every grammatical analysis
for the string should be constructible by the parser.  That is, the parser is
free of structural ambiguity resolution tendencies.

\item[incrementality] Given an initial segment of a sentence, the parser must
be able to identify all the semantic relations which necessarily hold among all
of the constituents seen thus far.  For example, having encountered a subject
NP followed by a transitive main verb, the parser must identify (or merely
narrow down, depending on one's theory of thematic relations) the semantic role
which the subject NP plays in the main sentence.

\item[feasibility] The computational resources needed to run the algorithm
must plausibly be provided by the human brain.  Given our current understanding
of the brain, this criterion is unavoidably fuzzy.  Clearly algorithms which
are exponential in the length of the string are infeasible; but should we brand
infeasible any algorithm which does not bound the processing time of each word
to a constant?  The answer is less clear: issues of implementation of
parallelism and the brevity of most utterances complicate matters.  In the case
of parsing CCG, the associativity of derivations must not impact the parser's
performance adversely.

\item[transparency] The parser uses the competence grammar directly, not a
specially transformed or compiled form.

\end{description}

\subsection{Previous Attempts}

There has been a variety of proposals for parsing CCG.  Wittenburg
(1986)\nocite{Wittenburg86}, Wall and Wittenburg (1989)\nocite{WW89} propose
that the grammar be compiled into a different one in which each semantically
distinct parse has a unique derivation (or, in some cases a few, but much fewer
than the Catalan series.)  Their proposal addresses only the rules \forw 0,
\forw 1, \back 0, and \back 1.  It does not seem to generalize obviously to
higher-order combinations, especially when so called mixed composition, in
which the slashes are not all of the same direction
\footnote{
\begin{minipage}[t]{2cm}For example\end{minipage} 
\BN{s}
   {\NLN{s/(a/e)}}
   {\BN{a/e} 
       {\BN{b/d/e} 
           {\NLN{b/c}} 
           {\NLN{c/d/e}} 
           {\forw 2}}
       {\NLN{a\bkslf(b/d)}}
       {\back 1 (crossing)}}
   {\forw 0}}.
This compilation process comes at the cost of substantially changing the
constituency structure defined by the linguist's original, source grammar,
hence compromising transparency.  Furthermore, the complexity of the operations
required to perform this compilation renders such a scheme a rather unlikely
account of human's representation of grammar.

Following up on the work of Lambek (1958)\nocite{Lambek58} who proposed that
the process of deriving the grammaticality of a string of categories be viewed
as a proof, there have been quite a few proposals put forth for computing only
normal forms of derivations or proofs.  (Moortgat 1988, K\"onig 1989, Hepple
and Morrill 1989, Hepple 1991) 
\nocite{Moortgat88}\nocite{Koenig89}\nocite{HepMor89}\nocite{Hepple91} The
basic idea with all of these works is to define `normal forms' ---
distinguished members of each equivalence class of derivations, and to require
the parser to search this smaller space of possible derivations.  These
proposals enjoy the advantage of transparency. Unfortunately, most of them
cannot result in parsing systems which proceed incrementally through the
string.  This results either from an intrinsically non-string-based
Gentzen-like proof system (Moortgat 1988, K\"onig 1989) or from a
right-branching normal form (Hepple and Morrill 1989).  A possible exception to
this criticism is the work of Hepple (1991).  Hepple considers Meta Categorial
Grammars, a close relative of CCG proposed by Morrill (1988)\nocite{Morrill88}.
Hepple's normal form derivations are as left-branching as the grammar allows
--- just the sort of incrementality necessary for our parser.  But Hepple does
not provide a computational implementation for the elegant normal form
construction which he presents.  Unfortunately, Hepple's claims that his system
can be parsed sufficiently incrementally are not tenable.  The problem is with
the timing: moving left-to-right through the input, the parser cannot know what
is ahead before it must commit to a normal form parse for the input so far.
For example, in

\startxl{jlm}
\BN[0in]
   {} 
   {\BN[0in]
       {}
       {\LN{John}{s/vp}}
       {\LN{loves}{vp/np}}
       {}} 
   {\LN{Mary}{np}} 
   {}
\stopx

Of the two possible derivations, the left-branching one (the one which treats
`John loves' as a constituent of type s/np) is the normal form.  However in

\startxl{madly}
\BN[0in]
   {} {\BN[0in]{}{\LN{John}{s/vp}}{\LN{loves}{vp/np}}{}}
{\BN[0in]{}{\LN{Mary}{np}}{\LN{madly}{vp\?vp}}{}} {}
\stopx

There is only one derivation.  This derivation treats `loves Mary' as a
constituent of type vp.  So what is a parser to do after having encountered
`John loves Mary'?  It is not allowed to construct the non-normal-form
derivation.  If it commits to the normal form derivation for these three words
then it would be stuck if an adverb were to come next.  It is also not allowed
to simply wait and not decide, because that would violate incrementality.
Stated differently, the problem is the inability to `extend' a left-branching
normal form by adding the next word.  When the next word in the input is
encountered, the processor computes distinct normal forms representative for
each distinct analysis; it has no general way of excluding those analyses which
are `extensions' of analyses which have already been discarded.

Karttunen (1989)\nocite{Karttunen89} proposes a very simple solution to the
problem of associativity of derivation.  He uses a bottom-up chart parser and
simply avoids adding duplicate arcs into the chart.  Since he uses a
unification-based system, he checks for subsumption, rather than simple
equality or unifiability of terms.  It follows that for a string of $n$
applications of the rule of forward composition, \forw 1, $O(n^2)$ arcs are
added to the chart, instead of $O(\mbox{Catalan}(n))$.  Karttunen's parser is
clearly sound, complete, and transparent.  But it doesn't construct {\em
derivations,} or {\em analyses.}  Instead, it constructs {\em arcs.} The
difference may appear insignificant --- at the end of the parse, those arcs
that span the whole string are exactly the analyses.  The difficulty arises in
the interaction with the interpreter.  The interpreter cannot simply check each
arc against every other arc: a constituent must be evaluated in the context of
its preceding constituents in the analysis.  The (syntactic)
context-independence assumption which dynamic programming algorithms (such as
Karttunen's chart parser) rely upon is not compatible with the context
necessary for interpretation.  The process of computing all valid
constituent-sequences which span the input so far is quite complex, especially
if one wishes to consider only maximally long constituents and not their
subconstituents (i.e.\ avoid truly spurious ambiguity.)  The cost of
integrating this chart parser with the rest of the current system thus renders
it infeasible.

Pareschi and Steedman (1987)\nocite{lazy} have made a third sort of proposal:
construct only maximally left-branching derivations, but allow a limited form
of backtracking when a locally non-maximally-left-branching analysis turns out
to have been necessary.  For example, when parsing \ref{madly}, Pareschi and
Steedman's algorithm constructs the left branching analysis for `John loves
Mary'.  When it encounters `madly', it applies \forw 0 in reverse to {\em
solve} for the hidden constituent `loves Mary' by {\em subtracting\/} the s/vp
category `John' from the s category `John loves Mary'.

\newcommand{\DoMadlyR}{ 
\SetPivot{\LN{John}{s/vp}}
\UN{s}
   {\BNp{vp} 
        {\UNp[\RevealingThickness] 
             {vp} 
             {\BN{s}
                 {\BN{s/np}
                     {\UsePivot}
                     {\LN{loves}{vp/np}}
                     {\forw 1}} 
                 {\LN{Mary}{np}} 
                 {\forw 0}} 
             {reveal \forw 0}} 
        {\LN{madly}{vp\?vp}} 
        {\back 0}} 
    {\forw 0}}
\startxl{madly-r}
\DoMadlyR
\stopx

The idea with this `revealing' operation is to exploit the fact that the rules
\forw$n$ and \back$n$, when viewed as three-place relations, are functional in
all three arguments.  That is to say, knowing any two of \{left constituent,
right constituent, result\}, uniquely determines the third.  There are some
problems with Pareschi and Steedman's proposal.

The first class of problems is the incompleteness of the parsing algorithm
which they give: a chart parser (Hepple 1987)\nocite{Hepple87}. The essence of
these problems is that in a chart parser, common sub-pieces are shared across
different analyses.  In Pareschi and Steedman's lazy chart parser, the presence
in one analysis of a certain arc can lead to the omission in another analysis
of a crucial arc.  Pareschi and Steedman use a scheme (`right-generator
marking') wherein if an arc has been combined with another arc to its left then
it is prevented from combining with any arcs on its right.  In \ref{madly-r}
the vp/np arc for `loves' is such an arc, and is therefore prevented from
combining with the np arc `Mary' to yield a vp arc.  In the presence of
ambiguity, this could lead to incompleteness.  For example, in
\ref{ps-incomplete} (from Hepple 1987) one category of the word `that' composes
with `he'.  This renders `he' unable to combine with `liked'.  It follows that
the parser cannot find an analysis for the whole string, which is, of course,
grammatical.

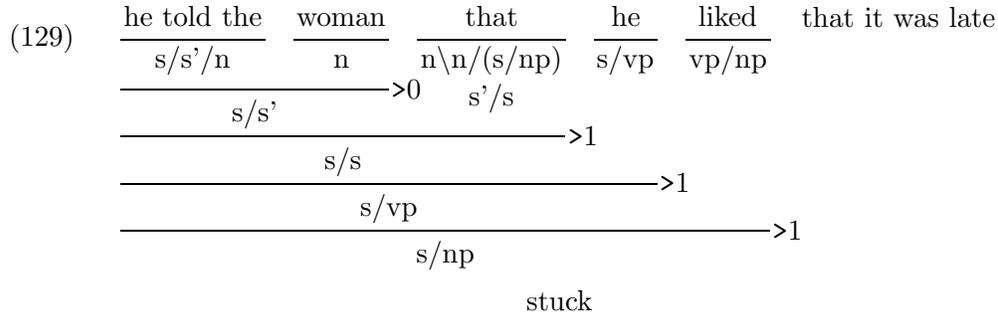
\startxl{ps-incomplete}
\BN[0mm]
   {stuck} 
   {\BN{s/np} 
       {\BN{s/vp} 
           {\BN{s/s} 
               {\BN{s/s'} 
                   {\LN{he told the}{s/s'/n}}
                   {\UN[0mm]{}{\LN{woman}{n}}{}}
                   {\forw 0}}
               {\UN[0mm]{\raisebox{1ex}{s'/s}}{\LN{that}{n\?n/(s/np)}}{}}
               {\forw 1}}
           {\LN{he}{s/vp}}
           {\forw 1}} 
       {\LN{liked}{vp/np}}
       {\forw 1}}
    {\LN[0mm]{that it was late}{}}
    {}
\stopx

This problem of separate analyses contaminating one another through shared
chart cells can be eliminated if one replaces the chart-parsing framework with
one that does not factor sub-results, as I do, below.

A second class of problems with Pareschi and Steedman's revealing computation
is the unsoundness which results from the assumption that the combinatory rules
are invertible.  In \ref{aac}, the category a/b is subtracted from a to reveal
the category b as the result of combining b/c and a\?(a/c).  This is an unsound
inference, regardless of the control algorithm in which it is embedded.  The
consequence is that the parser finds an analysis for
\ref{aac} which is not licensed by the grammar.

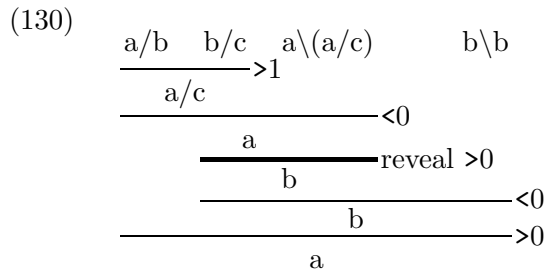
\startxl{aac}
\SetPivot{\NLN{a/b}}
\UN{a}
   {\BNp{b}
        {\UNp[\RevealingThickness]
             {b} 
             {\BN{a} 
                 {\BN{a/c}
                     {\UsePivot}
                     {\NLN{b/c}}
                     {\forw 1}} 
                 {\NLN{a\bkslf(a/c)}} 
                 {\back 0}}
             {reveal \forw 0}}
        {\NLN{\hspace{7mm}b\?b}}
        {\back 0}}
   {\forw 0}
\stopx

This form of unsoundness is not a problem if the grammar happens to be such
that whenever a constituent has a type-raised category, of the form X\?(X/Z)
for some categories X and Z, then it also has the category Y\?(Y/Z) for any
other category Y. While the class of such grammars may be of potential interest
(e.g.\ it would include any reasonable CCG for English), additional arguments
on language-universal grounds would be necessary before one accepts this
theoretical unsoundness as having no practical import.

Hepple (1987) provides another illustration of the unsoundness of the revealing
procedure.  For heavy NP shift, CCG allows a rule of backward crossing
composition, as in
\ref{CrazyPoet}. 

\startxl{CrazyPoet}
\BN{vp}
   {\BN{vp/np}
       {\LN{loves}{vp/np}}
       {\LN{madly}{vp\?vp}}
       {\back 1x}}
   {\LN{the crazy Scottish poet}{np}}
   {\forw 0}
\stopx

Also, the grammar allows coordination of non-traditional constituents, as in
\ref{SusanPassionately}.

\startxl{SusanPassionately}
\BN{vp}
   {\LN{loves}{vp/np}}
   {\TN{vp\?(vp/np)}
       {\BN{vp\?(vp/np)}
           {\LN{Mary}{vp\?(vp/np)}}
           {\LN{madly}{vp\?vp}}
           {\back 1}}
       {\LN{and}{conj}}
       {\BN{vp\?(vp/np)}
           {\LN{Susan}{vp\?(vp/np)}}
           {\LN{passionately}{vp\?vp}}
           {\back 1}}
       {coord}}
   {\back 0}
\stopx

Using revealing, the processor construct the following parse for
\ref{SusanPassionately}:

\startxl{SusanPassionatelyBis}
\SetPivot{\LN{loves}{vp/np}}
\UN{vp}
   {\TNp{vp\?(vp/np)}
        {\UNp[\RevealingThickness]
             {vp\?(vp/np)}
             {\BN{vp}
                 {\BN{vp}
                     {\LN{loves}{vp/np}}
                     {\LN{Mary}{np}}
                     {\forw 0}}
                 {\LN{madly}{vp\?vp}}
                 {\back 0}}
             {reveal \back 0}}
        {\LN{and}{conj}}
        {\BN{vp\?(vp/np)}
            {\LN{Susan}{vp\?(vp/np)}}
            {\LN{passionately}{vp\?vp}}
            {\back 1}}
        {coord}}
   {\back 0}
\stopx

The revealing step in \ref{SusanPassionatelyBis} cannot help but also reveal a
vp\?(vp/np) constituent for the string `madly the crazy Scottish poet', thus
allowing the processor to admit \ref{BrokenPoet} which is ruled out by the
grammar.

\startxl{BrokenPoet}
\SetPivot{\LN{loves}{vp/np}} 
* \ 
\UN{vp}
   {\TNp{vp\?(vp/np)}
        {\UNp[\RevealingThickness]
             {vp\?(vp/np)}
             {\BN{vp}
                 {\BN{vp/np}
                     {\LN{loves}{vp/np}}
                     {\LN{madly}{vp\?vp}}
                     {\back 1x}}
                 {\LN{the crazy Scottish poet}{np}}
                 {\forw 0}}
             {reveal \back 0}}
        {\LN{and}{conj}}
        {\BN{vp\?(vp/np)}
            {\LN{Susan}{vp\?(vp/np)}}
            {\LN{passionately}{vp\?vp}}
            {\back 1}}
        {coord}}
   {\back 0}
\stopx

Again, this unsoundness is not a problem if one assumes (as I do in this
project) that the semantic analysis of heavy-shifted constructions such as
`loves madly' bare markings which distinguish them from unshifted
constructions.  The discrepancy in this marking will prevent the coordination
rule from treating the two constituents in \ref{BrokenPoet} as `like
categories'.  But unless one has strong cross-linguistic evidence that the
unsoundness above will never present a problem for any reasonable grammar, it
is best to have parser which works correctly for every grammar in the
formalism.

Aside from introducing unsoundness, the revealing procedure is also incomplete.
In \ref{bcbc}, the category b\?c cannot be revealed after it had participated
in two combinations of mixed direction: \back 0 and \forw 0.

\startxl{bcbc}

\BN[0ex]
   {}
   {\UN[0ex]
       {}
       {\BN{a}
           {\NLN{a/b}}
           {\BN{b}{\NLN{c}}{\NLN{b\?c}}{\back 0}}
           {\forw 0}}
       {stuck}}
   {\NLN{b\?c\?(b\?c)}}
   {}
\stopx

\subsection{A Proposal}
\label{sec:Proposal}

Pareschi and Steedman's proposal embodies an appealing idea: construct the
maximally left branching analysis, revising this commitment only when it
becomes necessary.  The chart parser implementation of this lazy parsing idea
is clearly unacceptable.  Given the difficulty of incrementally computing
partial derivations from intermediate chart-parser states, and given the
interactive nature of the current system, the traditional advantages of a chart
parser (i.e.\ its reuse of analyzed substring across divergent analyses) are
eclipsed by its disadvantages.  Replacing the chart parser with a shift-reduce
parser which simulates nondeterminism using explicit parallelism eliminates the
problems associated with the system of right-generator marking.

But there are still problems with the subtraction-style operation of revealing.
The unsoundness and incompleteness in \ref{aac} and \ref{bcbc}, respectively,
still remain. One way out of these problems is to reparse the constituent
instead of revealing it.\footnote{This proposal is sketched in Hepple
(1987)\nocite{Hepple87}.} Reparsing the substring need not be performed from
scratch --- if the parser's data structure maintains links from a each
constituent to its subconstituents, then this derivation history can be reused
when constructing the revealed constituent.  For example, to reparse the vp
`saw three birds', the derivation history tells us to use the \forw 1 rule to
combine `saw' and `three' and then use the \forw 0 rule to combine `saw three'
with `birds'.

\startxl{three-birds}
\SetPivot{\LN{Fred}{s/vp}}
\UN{s}
   {\BNp{vp} {\UNp[\RevealingThickness] {vp} {\BN{s} {\BN{s/n} {\BN{s/np}
{\UsePivot} {\LN{saw}{vp/np}} {\forw 1}} {\LN{three}{np/n}} {\forw 1}}
{\LN{birds}{n}} {\forw 0}} {reparse}} {\LN{yesterday}{vp\?vp}} {\back 0}}
{\forw 0}
\stopx

Unfortunately, the presence of categorial applicability conditions on
combinatory rules presents the following problem to this `recipe-reparsing'
approach.  Suppose one wanted to rule out the following derivation from the
competence grammar
\startxl{bought-a} * 
\BN{s}
   {\BN{s/n} {\LN{Dan}{s/vp}} {\TN{vp/n} {\BN{vp/n} {\LN{bought}{vp/np}}
{\LN{a}{np/n}} {\forw 1}} {\LN{and}{coord}} {\BN{vp/n} {\LN{ate}{vp/np}}
{\LN{the}{np/n}} {\forw 1}} {coord}} {\forw 1}} {\LN{potato}{n}} {\forw 0}
\stopx

One could stipulate the following restriction on the rule \forw 0:

\startxl{silly-stipulation}
X/Y \ \ Y \ \ \GoesTo \ \ X \ \ \ \ unless X/Y = vp/n.
\stopx

Granted, this is not the only way of capturing this fact, nor is it a
particularly appealing one.  But this is a substantive grammatical question and
should not be resolved arbitrarily by the parsing algorithm.  Could
recipe-reparsing handle the following example?

\startxl{ate-spud-quickly}

\BN[0pt]
   {} {\BN{s} {\BN{s/n} {\BN{s/np} {\LN{Dan}{s/vp}} {\LN{ate}{vp/np}} {\forw
1}} {\LN{the}{np/n}} {\forw 1}} {\LN{potato}{n}} {\forw 0}}
{\LN{quickly}{vp\?vp}} {}
\stopx
Recipe reparsing done the obvious way (i.e.\ mirroring the derivation) would
first combine `ate' + `the' to make vp/n, and then attempt to combine that with
`potato'.  This derivation is ruled out.  But there does exist a derivation for
the above string:

\startxl{good-spud}

\BN{s}
   {\LN{Dan}{s/vp}} {\BN{vp} {\BN{vp} {\LN{ate}{vp/np}} {\BN{np}
{\LN{the}{np/n}} {\LN{potato}{n}} {\forw 0}} {\forw 0}} {\LN{quickly}{vp\?vp}}
{\back 0}} {\forw 0}
\stopx          

Recipe-reparsing therefore results in incompleteness.  Note that if one were to
change recipe-reparsing so as to work the other way around (i.e.\ build the
right-branching structure) then it would be possible to construct a
counter-example which required the left-branching analysis.

The move from revealing by subtraction to recipe-reparsing is one which trades
efficiency for accuracy.  Given that recipe-reparsing is not sufficiently
accurate, is it necessary to give up Pareschi and Steedman's intuition of lazy
parsing altogether and use full-fledged parsing on the substring to be
recovered?  I now argue that the answer is No.

The way I propose to exploit the information implicit in the derivation history
is by rewriting the derivation into another derivation which preserves all the
semantic relations encoded in the original derivation but makes possible
syntactic combinations which the original did not.  For example, the derivation

\startxl{jlmLB}
\BN{s \ : \ $\cB (\cT j) l\; m$}
   {\BN{s/np \ : \ $\cB (\cT j) l$}
       {\LN{John}{s/vp \ : \ $\cT j$}}
       {\LN{loves}{vp/np \ : \ $l$}}
       {\forw 1}}
   {\LN{Mary}{np \ : \ $m$}}
   {\forw 0}
\stopx

can be rewritten, using one step, to the equivalent right-branching derivation

\startxl{jlmRB}
\BN{s \ : \ $(\cT j) (l\; m)$}
   {\LN{John}{s/vp \ : \ $\cT j$}}
   {\BN{vp \ : \ $l\; m$}
       {\LN{loves}{vp/np \ : \ $l$}}
       {\LN{Mary}{np \ : \ $m$}}
       {\forw 0}}
   {\forw 0}
\stopx

which has the vp constituent necessary for combining with `madly', whose
category is vp\?vp.  I use the technique of term rewrite systems and normal
forms (Hepple and Morrill 1989, Hepple 1991\nocite{HepMor89}\nocite{Hepple91}).
Intuitively, semantics-preserving derivation-rewrite rules, such as the one
mapping \ref{jlmLB} to \ref{jlmRB} can be applied repeatedly to correctly
compute the right-branching equivalent of any derivation.  This computation can
be performed quite efficiently --- in time proportional to the size of the
derivation. In \refapp{DRSappendix} I provide a formal definition of the
rewrite operation, an effective procedure for applying this operation to
compute right-branching derivations and a proof of the correctness and
efficiency of this procedure.

\subsection{Using the Recovered Constituent}

\begin{figure}
\centerline{\psfig{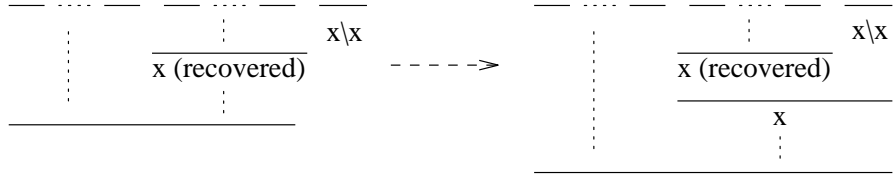}}
\caption{Recombining a recovered constituent with a rightward looking modifier}
\label{recombine}
\end{figure}

Given the rightmost subconstituent recovered using the normal form technique
above, how should parsing proceed?  Obviously, if the leftward looking category
which precipitated the normal form computation is a modifier, i.e.\ of the form
X\?X, then it ought to be combined with the recovered constituent in a form
analogous to Chomsky adjunction, as in \reffig{recombine}. As an illustration,
\ref{ill1} shows a state of the parser when it encounters a backward looking
category.  Normal form computation results in the state shown in \ref{ill2}.
{}From here, two states are possible, corresponding to the two ways of Chomsky
adjoining the modifier --- low and high attachment respectively.  These are
given in
\ref{ill3} and \ref{ill4}.

\startxl{ill1}
\BN[0mm]
   {}
   {\BN{s}
       {\BN{s/np}
           {\BN{s/vp}
               {\BN{s/s'}
                   {\BN{s/vp}
                       {\LN{John}{s/vp}}
                       {\LN{said}{vp/s'}}
                       {\forw 1}}
                   {\LN{that}{s'/s}}
                   {\forw 1}}
               {\LN{Bill}{s/vp}}
               {\forw 1}}
           {\LN{saw}{vp/np}}
           {\forw 1}}
       {\LN{Mary}{np}}
       {\forw 0}}
   {\LN{yesterday}{vp\?vp}}
   {}
\stopx
\startxl{ill2}
\BN[0mm]
   {}
   {\BN{s}
       {\LN{John}{s/vp}}
       {\BN[\RevealingThickness]
           {vp}
           {\LN{said}{vp/s'}}
           {\BN{s'}
               {\LN{that}{s'/s}}
               {\BN{s}
                   {\LN{Bill}{s/vp}}
                   {\BN[\RevealingThickness]
                       {vp}
                       {\LN{saw}{vp/np}}    
                       {\LN{Mary}{np}}
                       {\forw 0}}
                   {\forw 0}}
               {\forw 0}}
           {\forw 0}}
       {\forw 0}}
   {\LN{yesterday}{vp\?vp}}
   {}
\stopx
\startxl{ill3}   
\BN{s}
   {\LN{John}{s/vp}}
   {\BN{vp}
       {\LN{said}{vp/s'}}
       {\BN{s'}
           {\LN{that}{s'/s}}
           {\BN{s}
               {\LN{Bill}{s/vp}}
               {\BN{vp}
                   {\BN{vp}
                       {\LN{saw}{vp/np}}    
                       {\LN{Mary}{np}}
                       {\forw 0}}
                   {\LN{yesterday}{vp\?vp}}
                   {\back 0}}
               {\forw 0}}
           {\forw 0}}
       {\forw 0}}
   {\forw 0}
\stopx
\startxl{ill4}
\BN{s}
   {\LN{John}{s/vp}}
   {\BN{vp}
       {\BN{vp}
           {\LN{said}{vp/s'}}
           {\BN{s'}
               {\LN{that}{s'/s}}
               {\BN{s}
                   {\LN{Bill}{s/vp}}
                   {\BN{vp}
                       {\LN{saw}{vp/np}}    
                       {\LN{Mary}{np}}
                       {\forw 0}}
                   {\forw 0}}
               {\forw 0}}
           {\forw 0}}
       {\LN{yesterday}{vp\?vp}}
       {\back 0}}
   {\forw 0}
\stopx

But what if this category is not of the form X\?X?  Should the
parser compute the reanalysis in \ref{nonmonotonic-reanalysis}?

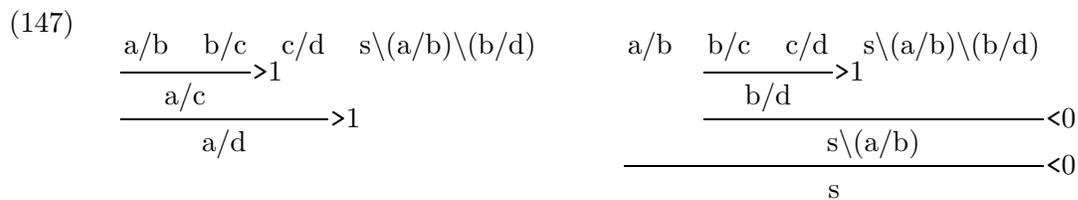
\startxl{nonmonotonic-reanalysis}
\BN[0pt]
   {}
   {\BN{a/d}
       {\BN{a/c}
           {\NLN{a/b}}
           {\NLN{b/c}}
           {\forw 1}}
       {\NLN{c/d}}
       {\forw 1}}
   {\NLN{s\bkslf(a/b)\bkslf(b/d)}}
   {}\hspace{1cm}
\BN{s}
   {\NLN{a/b}}
   {\BN{s\bkslf(a/b)}
       {\BN{b/d}
           {\NLN{b/c}}
           {\NLN{c/d}}
           {\forw 1}}
       {\NLN{s\bkslf(a/b)\bkslf(b/d)}}
       {\back 0}}
   {\back 0}
\stopx

Such a move would constitute a very odd form of cost-free backtracking.  Before
reanalysis, the derivation encoded the commitment that the /b of the first
category is satisfied by the b of the b/c in the second category.  This
commitment is undone in the reanalysis.  This is an undersirable property to
have in a computational model of parsing commitment, as it renders certain
revisions of commitments easier than others, without any empirical
justification.  Furthermore, given the possibility that the parser change its
mind about what serves as argument to what, the interpreter must be able to
cope with such non-monotonic updates to what it knows about the derivation so
far --- this would surely complicate the design of the interpreter.\footnote{I
am indebted to Henry Thompson for a discussion of this issue of monotonicity.}

\section{Summary}

This chapter began by reviewing a very bold proposal of Steedman's: The
internal representation used by the human syntactic parser consists only of
grammatical analyses.  The proposal is bold on two counts:

\begin{enumerate}
\item This processing model is unusually impoverished.
\item On the basis of the parsimony of the grammar + parser package, 
Steedman attempted to argue for a certain theory of competence.
\end{enumerate}

The primary thrust of the argument (point 2) --- that in principle, a processor
for CCG avoids design complexity which is necessary for other grammatical
frameworks --- was challenged by Shieber and Johnson's argument that
asynchronous computation could capture the same computational simplicity for
rather traditional-looking phrase structure grammars.  Resolution of this issue
awaits refinement and elaborations of each of these theories to allow their
evaluation as adequate characterization of how the brain actually represents
and processes grammars.

Returning to point 1 above, I considered whether an impoverished pure bottom up
CCG parser can serve as an adequate parsing module for the language processing
system.  I considered three problems which would traditionally have received
some sort of `precompilation of the grammar' or `top down prediction' (in the
parsing sense of top-down)
\begin{description}
\item[Timely detection of ungrammaticality] e.g.\ the ability to quickly detect
that an
adjacent pair of categories (e.g.\ determiner verb) has no chance of ever
leading to a grammatical analysis
\item[Shift reduce conflicts] identifying the rare set of cases where a CCG 
rule should be
allowed not to apply (e.g.\ picture-noun extractions)
\item[Timely detection of crossing composition] detecting the inevitability 
of certain rule applications before they actually happen (e.g.\ detecting heavy
shift when an obligatorily transitive verb is immediately followed by a
preposition)
\end{description}
Not surprisingly, the pure bottom up processor cannot handle these cases
correctly. More interestingly, however, I have argued that one's theory of the
innate processor can remain as parsimonious as Steedman's if one makes the
rather plausible assumption that while the ability to parse is innate, the
ability to parse {\em efficiently\/} is not.  The skill which the language
learner acquires by attending to intermediate parser configurations and their
eventual outcomes can serve to perform the `predictive' functions necessary for
the three cases above.  The acquisition process is similar in some ways to the
training of n-gram models for part-of-speech taggers.

\ifprediction
  
In the last section of this chapter I discussed a problem which is quite
specific to CCG: CCG distinguishes left-branching and right-branching analyses
which are often truth-conditionally equivalent.\footnote{This property has been
called `spurious ambiguity' (Wittenburg 1986\nocite{Wittenburg86}).  Steedman
(1991)\nocite{Steedman91} has argued that this ambiguity is not spurious,
rather different constituencies correspond to different ways of breaking the
string into a theme and a rheme --- prosodic constituents which are used to
encode information status.  But CCG provides more ambiguity than what is
necessary for prosodic constituency --- the theme and the rheme may, in turn,
receive many truth-conditionally equivalent derivations. }  To cope with the
additional ambiguity brought about by CCG's associativity of derivation, I
proposed that only the maximally left-branching analysis (as allowed by the
grammar) be maintained, and, whenever this analysis turns out not be the
correct one, the necessary right-branching analysis is computed from the
derivation history.

Steedman's proposal of a parser which only represents grammatical analyses has
therefore survived the challenges which it had been put to.  In the next
chapter, I show how the resulting parsing algorithm is used in the broader
sentence processing system.

\else  

\fi 

\chapter{A Computer Implementation}
\label{ch:implementation}

In this chapter I instantiate the parsing mechanism described in
\refch{ch:parsing} and the meaning-based ambiguity resolution mechanisms
presented in chapters 2 through 4. I do so by presenting a computer program
which simulates human sentence processing performance.  The aim of this chapter
and the implementation it describes is to show the consistency of the
collection of subtheories developed thus far to account for the limited data
that has been collected, and to test whether these ingredients can indeed be
combined in a straight-forward and non {\em ad hoc\/} way.

The program accepts words as input, one at a time, developing a set of partial
analyses as it progresses through the sentence.  If at any time, this set
becomes empty, the processor is said to have failed --- the analog of a garden
path.  In this project, I do not address recovery from a garden path.  This
model is successful just in case two goals are achieved:
\begin{enumerate}
\item It correctly predicts garden path effects in the range of examples 
discussed in the earlier chapters.
\item The implementation is `straight-forward', that is, it is a simple 
procedure which applies linguistic competence to the input representation,
without having to resort to specialized algorithms.
\end{enumerate}

\section{Desiderata}
\label{sec:desiderata}

\begin{figure}
\centerline{\psfig{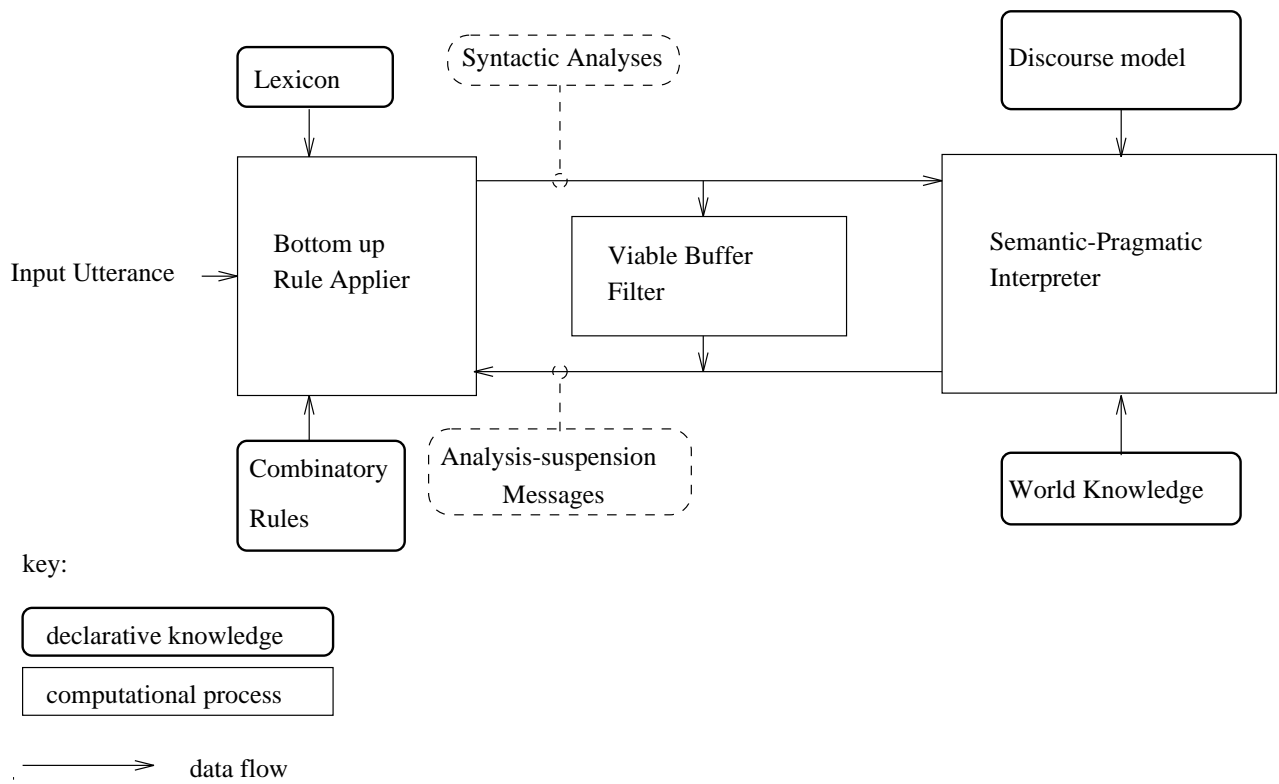}}
\caption{System Diagram}
\label{sysdiag}
\end{figure}

Let us begin by stating the desiderata for the computational model in detail.
The system is divided into the modules shown in \reffig{sysdiag}.  The
bottom-up syntactic rule applier (i.e.\ the parser) constructs in parallel all
possible analyses for the initial segment seen so far.  The buffer-viability
filter detects unviable analyses and immediately signals the parser to discard
these.  The semantic-pragmatic interpreter examines only the sense-semantics
which the parser constructs, and not other, more superficial aspects of the
syntactic analyses.  The parser, in turn, may not `look inside' the interpreter
--- the only information flowing from the interpreter to the parser is whether
to maintain or discard current analyses.  The actual program does not literally
separate the different procedural modules into informationally encapsulated
modules (e.g.\ using asynchronous communicating processes) but nevertheless
obeys these restrictions on data flow.

To avoid the inferential complexities associated with accommodation, the
repositories of knowledge about the world and knowledge of the preceding
discourse are not updated by the interpreter, i.e.\ they are treated as
read-only storage.

The following phenomena are covered:
\begin{enumerate}
\item {\bf Referential Felicity:} Crain and Steedman 
(1985)\nocite{CrainSteedman85}
show context sensitivity in pairs such as \ref{Wife} 
(see \refsec{sec:WeaklyInteractive};
Altmann \etal\ 1992).
\startxl{Wife}
\begin{tabbing}
a. \= The psychologist told the wife that he was having trouble with to leave\\
   \> her husband.\\
b. The psychologist told the wife that he was having trouble with her \\
   \>husband.
\end{tabbing}
\stopx
In a context with just one wife, \ref{Wife}a is a garden path, whereas
\ref{Wife}b. is not.  The opposite is true if the context mentions two wives.

\item {\bf Complexity of Accommodation:} Crain and Steedman's (1985) 
Principle of Parsimony
\ref{ParsimonyAgain} (see \refsec{sec:WeaklyInteractive}) entails that out of 
context, the simplex NP reading of `the wife', compatible with \ref{Wife}b,
would be preferred to the restrictively modified NP reading of
\ref{Wife}a.\footnote{I do not consider non-restrictive relative clauses in
this project.}

\startxl{ParsimonyAgain}
\parbox[t]{4.5in}{
{\em Principle of Parsimony:} (Crain and Steedman 1985)\\ If there is a reading
that carries fewer unsatisfied but consistent presuppositions or entailments
than any other, then, other criteria of plausibility being equal, that reading
will be adopted as most plausible by the hearer, and the presuppositions in
question will be incorporated in his or her [mental] model [of the discourse].}
\stopx

In the current project, the complex process of determining the number and
plausibility of presuppositions carried by an NP will be approximated by a very
simple and crude method: accommodating a simple NP incurs no cost; while
accommodating an NP which is restrictively modified carries some fixed cost.
It must be emphasized that this approximation is not based on the {\em
syntactic\/} complexity of complex NPs, but on the presupposition encoded by
the use of a restrictive relative clause: there is more than just one entity
matching the description of the head noun, so a restrictive modifier is
necessary to individuate the referent intended by the speaker.

\item {\bf Plausibility and Garden Paths: }  Bever (1970)\nocite{Bever70} 
noticed that garden path
effects, as in
\startxl{HorseRacedX}
The horse raced past the barn fell.
\stopx
can be minimized when the plausibility of the main verb analysis of the first
verb is decreased. (see \refsec{sec:WeaklyInteractive}) Bever's example is
\startx
The light airplane pushed past the barn crashed into the post.
\stopx
In this project, I use a slight variant:
\startxl{PoemPoet}
a. The poet read in the garden stank.\\ 
b. The poem read in the garden stank.
\stopx

\item {\bf Heavy Shift and Garden Paths: }
Pritchett (1988)\nocite{Pritchett88} points out the garden path effect 
in \ref{HorseRacedX} is
absent in \ref{BirdFound}.
\startxl{BirdFound}
The bird found in the store died.
\stopx
Clearly the fact that `find' is an obligatorily transitive verb plays a role
here.  Given that
\startxl{BirdFoundA}
The bird found in the store a corner in which to nest.
\stopx
is also not a garden path sentence, it follows that both reduced relative and
main verb analyses of `found' are pursued in parallel.  It is possible to force
one or the other reading using an appropriate context:
\startxl{FoundContext}
\begin{tabbing}
Q: What did the bird find in the store?\\ 
A: The bird found in the store died.\\ 
A: The bird found in the store a corner in which to nest.
\end{tabbing}
\stopx

\startxl{DiedContext}
In the pet store, two exotic birds escaped from their cages.  One was located
in a nearby tree and the other was found hiding inside the store.
\begin{tabbing}
The bird found in the store died.\\
The bird found in the store a corner in which to nest.
\end{tabbing}
\stopx

\item {\bf Adverbial Attachment: }  In \refch{ch:informational} it was argued 
that considerations of information volume were responsible for the low
attachment preference of the adverbial in
\startxl{JSTBLY}
The poet said that the psychologist fell yesterday.
\stopx
But that no such considerations apply to the attachment of the adverbial in
\ref{JSTTDF}
\startxl{JSTTDF}
The poet said that the psychologist fell because he disliked us.
\stopx
Since the inference required for determining correct attachment decisions in
\ref{JSTTDF} is open ended and non-linguistic, the current program leaves this
ambiguity unresolved and reports both readings.

\item {\bf So-called Late Closure Effects: } Out of context, the examples 
in \ref{LCEL} are
garden paths.
\startxl{LCEL}
\begin{tabbing}
a. When the cannibals ate the missionaries drank. \\
b. Without her contributions failed to come in.  \\
c. \=When they were on the verge of winning the war against Hitler, Stalin, \\
   \>Churchill and Roosevelt met in Yalta to divide up postwar Europe.
\end{tabbing}
\stopx

I have implemented both a new-subject detector and a disconnectedness
determining procedure in order to experiment with the two theories presented in
\refch{ch:leftovers}.

\item {\bf Center Embedding Effects: } While \ref{WormBird} does not give rise 
to garden path effects, the system does represents the fact that it is `harder'
than other sentences.
\startxl{WormBird}
The worm that the bird that the poet watched found died.
\stopx
This measure of difficulty is lower when some of the subjects are given in the
discourse.

\end{enumerate}

\section{Syntax}
\label{sec:grammar}

The competence grammar in this system is an instantiation of Steedman's
(1990)\nocite{GACC} Combinatory \mbox{Categorial} Grammar which is capable of
constructing left branching analyses, as discussed in
\refsec{sec:SteedmansProposal}.  A proper linguistic investigation of
grammatical competence being outside the scope of this work, the aim of the
grammar here is to provide at least one analysis for each reading relevant for
the examples in
\refsec{sec:desiderata}.  To this end, the following grammar will do.  

A Basic category is represented as an ordered pair: a Prolog
term\footnote{\label{foot:PrologNames} I use Prolog notation throughout:
symbols beginning with a lowercase letter are constants; symbols beginning with
an uppercase letter are variables; an underscore (\UL) denotes an `anonymous
variable'; different occurrences of \UL\ denote different anonymous variables.
See Pereira and Shieber (1987)\nocite{PereiraShieber87} for an introduction to
the Prolog programming language.

There is an exception to this naming scheme, however: Prolog is usually unable
to keep track of names of variable names after unification takes place.  When
it must print a variable, it prints something tedious such as \UL{\tt 83754}.
To make terms easier to read, I use a printing procedure which gives semantic
indices names such as e1, e2$\ldots$, syntactic variables names such as s1,
s2$\ldots$, and category variables names such as c1, c2$\ldots$} and a semantic
variable, separated by a colon. The Prolog term is a major category symbol with
zero or more arguments --- its features.  The basic categories are

\centerline{
\begin{tabular}{|ll|}
\hline
\multicolumn{2}{|c|}{Basic categories} \\ \hline
n(NUM)          & common noun \\
np(PERS,NUM)    & noun phrase \\
s(TNS,FIN,COMP) & sentence, or SBAR \\
part(PART)      & particle \\
pp(PREP)        & prepositional phrase \\ 
eop             & end of phrase marker (zero morpheme) \\
\hline
\end{tabular}}

A feature may be unspecified, or have a value from the following domains:

\centerline{
\begin{tabular}{|ll|}
\hline
Feature & Possible values \\ \hline
NUM     & sg, pl \\
PERS    & 1, 2, 3 \\
TNS     & to, en, ing, plup, ed, s, fut, --ed, --s , --fut\\
FIN     & +, -- \ (depends on TNS:  plup, ed, s are +; --ed, --s, --fut, to, 
	  en, ing are --)\\
PART    & away, down, up, over\ldots \\
COMP    & 0, that, q \ (q is special, it means that the s is a WH question) \\
PREP    & in, to, without\ldots\\ \hline
\end{tabular}}

For example, the basic category s(ed,+,that):X stands for a sentence in the
past tense whose complementizer is `that'. (e.g.\ `that Mary loves John.') In
the accompanying semantic term list, the variable X represents the main
situation in the sentence.

The lexicon is stored as a collection of words and their associated part of
speech label.  When the system is started, a process generates lexical entries
{}from these part of speech labels.  A lexical entry is a triple of a word, a
syntactic category, and a semantic term list.  Examples of the different parts
of speech labels are as follows:

\rotate[l]{\mbox{
\begin{minipage}{7.8in}
\begin{tabular}{|l@{\hspace{3mm}}l@{\hspace{3mm}}l@{\hspace{
		2mm}}l@{\hspace{2mm}}l|}
\hline 
\multicolumn{2}{|l}{P.O.S. label}&
 example &
 syntactic category &
 semantic term list\\ \hline
v        &
 intransitive V       &
 walk          &
 \VEEPEE\footnotemark  
                                                                &
 walk(S,X,\UL), tns(S,s)\\
vo       &
 transitive V         &
 call          &
 \VEEPEE/np(\UL,\UL):Y 
                                                                &
 call(S,X,Y), tns(S,s)\\
vpr      &
 V + part             &
 go (away)     &
 \VEEPEE/part(away):\UL 
                                                                &
 go\UL away(S,X), tns(S,s)\\
vi       &
 V + infinitival S &
 try&
 \VEEPEE/eop:S/(s(to,--,0):Y\?np(\UL,\UL):X)
                                                                &
 try(S,X,Y), tns(S,s)\\
voi      &
 V + Obj + Sinf       &
 reminds 
                &
 \VEEPEE/eop:S/(s(to,--,0):Y\?np(\UL,\UL):Z)/np(\UL,\UL):Z
                                                &
 remind(S,X,Z,Y), tns(S,s)\\
vc       &
 V + S complement     &
 say           &
 \VEEPEE/eop:S/s(\UL,+,0):Y
                                                                &
 say(S,X,Y), tns(S,s)\\
voc      &
 V + Obj + Scomp      &
 tell  &
 \VEEPEE/eop:S/s(\UL,+,0):Y/np(\UL,\UL):Z
                                                                &
 tell(S,X,Z,Y), tns(S,s)\\
vop      &
 V + Obj + PP         &
 grant &
 \VEEPEE/pp(to):Y/np(\UL,\UL):Z
                                                                &
 grant(S,X,Z,Y), tns(S,s)\\ 
cn       &
 common noun          &
 bird          &
 n(sg):X       &
 bird(X)\\
mn       &
 mass noun            &
 wax           &
 n(sg):X       &
 wax(X)\\      
         &
 0 det mass np        &
               &
 np(3,sg):X    &
 exist(X), wax(X)\\
pn       &
 proper name          &
 John          &
 np(3,sg):X    &
                                the(X), name\UL of(X,john), closed(X)\\ 
nom\UL pro &
 nominative pron.   &
 we          &
 s(T,F,0):S/(s(T,F,0):S\?np(1,pl):X) 
                                                                &
 the(X), 1st\UL prs(X),
                                                             pl(X), closed(X)\\
obj\UL pro &
 object pronoun     &
 us          &
 np(1,pl):X      &
 the(X), 1st\UL prs(X),
                                                             pl(X), closed(X)\\
poss\UL pro &
 possessive pron.  &
 our         &
 np(3,N):X/eop:X/n(N):X  
                                                        &
 the(Y), 1st\UL prs(Y), pl(Y),\ \ldots \\
            &
                   &
             &
         &
 \ \ \ closed(Y), the(X), of(X,Y)\\
det      &
 determiner           &
 the           &
 np(3,N):X/eop:X/n(N):X 
                                                                &
 the(X)\\
part     &
 particle             &
 away        &
 part(away) &
\\
adj      &
 adjective            &
 juicy       &
 n(N):X/n(N):X &
 juicy(X)\\
\multicolumn{2}{|l}{post\UL vp\UL adv} &
 passionately  &
 s(T,F,0):S\?np(P,N):X\?\ \ldots &
 \\
         &
                      &
             &
 \ \ \  (s(T,F,0):S\?np(P,N):X) 
                                                    &
 passionately(S), swa(S)\\
\multicolumn{2}{|l}{post\UL s\UL adv}  &
 yesterday &
 s(T,F,0):S\?s(T,F,0):S&
 yesterday(S), swa(S)\\
prep     &
 preposition          &
 in          &
 pp(in):X/np(\UL,\UL):X &
 \\
         &
 N postmodifier       &
             &
 n(N):X\?n(N):X/np(\UL,\UL):Y &
 in(X,Y), npmod(X)\\
         &
 S postmodifier       &
             &
 s(T,F,0):S\?s(T,F,0):S/np(\UL,\UL):Y &
 in(S,Y)\\
         &
 S premodifier        &
             &
 s(T,F,0):S/s(T,F,0):S/np(\UL,\UL):Y  &
 in(S,Y)\\
sconj    &
 subordinating conj   &
 whenever X Y&
 s(T,F,0):Y/s(T,F,0):Y/eop:X/s(\UL,\UL,0):X &
                                                                whenever(X,Y)\\
         &
                      &
 Y whenever X&
 s(T,F,0):Y\?s(T,F,0):Y/eop:X/s(\UL,\UL,0):X &
                                                                whenever(X,Y)\\
\hline
\end{tabular}
\footnotetext{$^{\thefootnote}$\VEEPEE\ stands for s(s,+,0):S\?np(\UL,pl):X}   
\end{minipage}
}}
\newpage

In addition to the above lexical entry generators, `idiosyncratic' (i.e.\
closed-class) words have the lexical entries listed in table~\ref{ClosedClass}.

\newcommand{\ZPEZ}{\hspace{2mm}}
\begin{table}
\centerline{
\begin{tabular}{|l@{\ZPEZ}l@{\ZPEZ}l@{\ZPEZ}l|} \hline
word  &
 category                                        &
 term list &
 comment \\ \hline
that  &
 s(T,+,that):E/s(T,+,0):E                         &
          &
 complementizer \\
that  &
 n(N):E\?n(N):E/eop:S/(s(\UL,+,0):S\?np(3,N):E)   &
 npmod(E) &
  subject relativizer \\
that  &
 n(N):E\?n(N):E/eop:S/(s(\UL,+,0):S/np(3,N):E)    &
 npmod(E) &
  object relativizer \\
which &
 n(N):E\?n(N):E/eop:S/(s(\UL,+,0):\UL/np(3,N):E)  &
 npmod(E) &
  object relativizer\\
which &
 s(T,+,q):E/(s(T,+,q):\UL/np(3,N):E)/n(N):E       &
 wh(E)    &
 question\\
did   &
 s(ed,+,q):E/s(--ed,--,0):E                       &
          &
 for subj-aux inversion\\
to    &
 s(to,--,0):E\?np(P,N):X/(s(--to,--,0):E\?np(P,N):X) &
       &
 infinitive to\\
will  &
 s(fut,+,0):E\?np(P,N):X/(s(--fut,-,0):E\?np(P,N):X) &
 future(E) &
 aux will\\
was   &
 s(ed,+,0):E\?np(P,N):X/(s(ing,--,0):E\?np(P,N):X)   &
           &
 past progressive \\
is    &
 s(s,+,0):E\?np(P,N):X/(s(ing,--,0):E\?np(P,N):X)    &
           &
 pres. progressive\\
had   &
 s(plup,+,0):E\?np(P,N):S/(s(en,--,0):E\?np(P,N):S)  &
 pluperfect(E) &
\\
been  &
 s(en,--,0):E\?np(P,N):S/(s(ing,--,0):E\?np(P,N):S)  &
         &
 past perf. progressive \\ 
$\epsilon$ &
 eop:X       &
  closed(X)  & end of phrase marker \\
\hline
\end{tabular}}
\caption{Lexical entries for closed class items.}
\label{ClosedClass}
\end{table}

The annotation swa(X) is associated with all single word adverbials.  The
annotation npmod(X) is associated with all nominal modifiers.  These
annotations allow interpreter to approximate the detection of a low information
volume adverbial preceded by a high information volume argument, as discussed
in \refch{ch:informational}.

The zero morpheme eop:X, and the semantic terms the(X) name\UL of(X), of(X,Y),
and phrase\UL closed(X), are part of the reference resolution system.  They
will be described in \refsec{sec:DefiniteReference}.  The latter annotation,
phrase\UL closed(X), is abbreviated in the table above as simply closed(X) for
reasons of space.

There are lexical entries for each verb in all of its inflected forms as
follows:\\

\centerline{
\begin{tabular}{|l|l|l|} \hline
form    & category & semantics \\ \hline
walked  & s(ed,+,0):E\?np(\UL,\UL):X     & [walk(E,X),tns(E,ed)]\\
walked  & s(en,--,0):E\?np(\UL,\UL):X    & [walk(E,X),tns(E,en)]\\
walking & s(ing,--,0):E\?np(\UL,\UL):X   & [walk(E,X),tns(E,ing)]\\
walks   & s(s,+,0):E\?np(3,sg):X         & [walk(E,X),tns(E,s)] \\
walk    & s(--T,--,0):E\?np(\UL,\UL):X   & [walk(E,X),tns(E,T)] (untensed)\\
walk    & s(s,+,0):E\?np(\UL,pl):X       & [walk(E,X),tns(E,s)] (plural 
present)\\
walk    & s(s,+,0):E\?np(1,\UL):X        & [walk(E,X),tns(E,s)] (1st pers 
present)\\
walk    & s(s,+,0):E\?np(2,\UL):X        & [walk(E,X),tns(E,s)] (2nd pers 
present)\\
\hline
\end{tabular}}

The tense system implemented in the current grammar is rather crude, but it
suffices to construct the analyses necessary for the examples.

Subject-Aux inversion is handled as follows:\footnotemark

{\small
\BN{s(ed,+,q):e1/np(3,sg):e2}
   {\BN{s(ed,+,q):e1/(s(--ed,--,0):e1\?np(3,sg):e3)}
       {\LN{did}{s(ed,+,q):e1/s(--ed,--,0):e1}}
       {\LN{Mary}{s(X,Y,0):e1/(s(X,Y,0):e1\?np(3,sg):e3)}}
       {\verb+>+1}}
   {\LN{find}{s(--T,--,0):e1\?np(3,sg):e3/np:e2}}
   {\verb+>+1}\\[2mm]}
[name\UL of(e3,mary),find(e1,e3,e2),tns(e1,ed)]
\footnotetext{Recall (footnote \ref{foot:PrologNames}) that symbols like e1
stand for semantic variables.}

Notice that the identity of the tense, `ed' in this case, is passed from `did'
through `Mary' (where the tense variable X is unified with --ed), to `find'
whose lexical semantics include a variable, T, which is unified with `ed'.

The s(ed,+,q)/np constituent `did Mary find' can then combine to form a WH
question: 

\BN{s(ed,+,q):e1}
   {\LN{which bird}{s(T,+,q):e1/(s(T,+,q):e3/np(\UL,\UL):e1)}}
   {\LN{did Mary find}{s(ed,+,q):e3/np(3,sg):e1}}
   {\forw 0}\\[2mm]
[wh(e1),bird(e1),name\UL of(e2,mary),find(e3,e2,e1),tns(e3,ed)]

A subject type raising rule applies to all NPs with the exception of objective
case pronoun:

\startx
np(P,N):X, sem:S \ \ $\longrightarrow$ \ \ s(T,F,0):S(s(T,F,0)\?np(P,N):X),
sem:[subj(X)\Kons S]\footnotemark
\stopx
\footnotetext{The Prolog notation [H\Kons T] stands for a list whose first 
element is H and
the rest of whose elements are T.  The subject type-raising rule, therefore
adds the notation that the NP appears as subject to the semantic term list
associated with the NP.}

A variant of this rule applies to all determiners:

\startx
np(Pers,Num):X/eop:X/n(Num):X, sem:S \ \ $\longrightarrow$ \\
\ \ \ s(T,F,0):S(s(T,F,0)\?np(Pers,Num):X)/eop:X/n(Num):X, sem:[subj(X)\Kons S]
\stopx

Words that create non-subject WH-dependencies (relativizers, wh-question words)
each have, in addition to the categories listed in table~\ref{ClosedClass} two
additional categories which reflect one and two applications of a non-direction
preserving version of Geach's division rule (Geach 1971)\nocite{Geach71}.

\IndentABit{X/Y \ \ \GoesTo \ \ (X/Z)/(Y/Z)}

For example, the relativizing pronoun `which' has, in addition to the category 

\IndentABit{n(N):E\?n(N):E/eop:S/(s(\UL,+,0):\UL/np(3,N):E)}

listed in table~\ref{ClosedClass}, the following two categories:

\IndentABit{n(N):E\?n(N):E/X/eop:S/(s(\UL,+,0):\UL/X/np(3,N):E)}\\
\IndentABit{n(N):E\?n(N):E/X/Y/eop:S/(s(\UL,+,0):\UL/X/Y/np(3,N):E)}

The latter two categories are included in order to allow for `non-peripheral
extraction',\footnote{See Steedman (1992)\nocite{SurfaceStructure} for a
different way to capture non-peripheral extraction --- interaction between
crossing composition and object-type-raising. } for example

\startxl{NonPeripheral}
Mark reminded the babysitter to watch the movie.\\
the babysitter that Mark reminded to watch the movie
\stopx

The combinatory rules are as follows:\\

\centerline{
\begin{tabular}{|llll|}
\hline
left-child & right-child & result & rule name \\ \hline
A/B     & B             & A       & \forw 0\\
A/B     & B/C           & A/C     & \forw 1\\
A/B     & B/C/D         & A/C/D   & \forw 2\\
A/B     & B/C/D/E       & A/C/D/E & \forw 3\\
B       & A\?B          & A       & \back 0\\
A/B     & C\?A          & C/B     & \back 1\\ \hline
\end{tabular}}

The capital letters in the rules are Prolog variables, and these rules operate
by unification.

Along lines suggested by Aone and Wittenburg (1990)\nocite{AoneWittenburg90}
there is a rule for positing a zero morpheme adjacent to a category which
expect it.  The processor blocks excessive applications of this rule.  For
example, given a determiner and a noun, the rule \forw 0 applies to combines
them and yield the category np/eop.\footnote{Inessential details inside the
categories are omitted for clarity.}

\startxl{DetNoun}
\BN{np/eop}
   {\LN{the}{np/eop/n}}
   {\LN{bird}{n}}
   {\forw 0}
\stopx

The zero morpheme eop (end of phrase) is then posited to the right of the noun,
and immediately combined to yield an np.

\startxl{DetNounEop}
\BN{np}
   {\BN{np/eop}
       {\LN{the}{np/eop/n}}
       {\LN{bird}{n}}
       {\forw 0}}
   {\LN{$\epsilon$}{eop}}
   {\forw 0}
\stopx

When a rule is applied to combine two constituents, the semantic term list of
the result is simply the concatenation of the term lists of the two
constituents, with one exception: the last combinatory rule, so called
`backward crossing composition' introduces an additional term, \mbox{h\UL
shifted(X,Y),} which designates that argument X of Y was heavy shifted.  For
example, in

\startxl{FoundYesterday}
\BN{s:e1}
   {\BN{s:e1/eop:e3}
       {\BN[\RevealingThickness]
           {s:e1/np:e3}
           {\BN{s:e1/np:e3}
               {\LN{john}{s:e1/(s:e1\?np:e2)}}
               {\LN{found}{s:e1\?np:e2/np:e3}}
               {\forw 1}}
           {\LN{yesterday}{s:e1\?s:e1}}
           {\back 1}}
       {\LN{a nice car}{np:e3/eop:e3}}
       {\forw 1}}
   {\LN{$\epsilon$}{eop:e3}}
   {\forw 0}
\stopx

the semantic term lists associated with the marked derivation step are as
follows

\begin{tabbing}
John found\\
\ \ \ \= the(e2), name\UL of(e2,john), closed(e2), find(e1,e2,e3), tns(e1,ed)\\
yesterday\\
      \> yesterday(e1), swa(e1)\\
john found yesterday\\
      \> the(e2), name\UL of(e2,john), closed(e2), find(e1,e2,e3), tns(e1,ed),
yesterday(e1), swa(e1), h\UL shifted(e3,e1)
\end{tabbing}

\section{Data Structure}
\label{Sec:DataStructure}

The processor maintains one or more analyses in parallel.  Each analysis has
data components on two levels: Syntax/Semantics, and Interpretation/Evaluation.
There are four components altogether:

\[
\begin{array}{rl}
  \mbox{Syntax/Semantics} & \left\{ \begin{array}{l}
     \mbox{Buffer}\\ \mbox{Semantic Term List}
  \end{array} \right. \\
\rule{0pt}{-0.1ex}\\
  \mbox{Interpretation/Evaluation} & \left\{ \begin{array}{ll}
     \mbox{Interpreter Annotations}\\\mbox{Penalties}
  \end{array} \right.
\end{array}
\]

The Buffer is a sequence of constituents.  Adjacent constituents may be
combined using the combinatory rules or `revealing' (see \refch{ch:parsing}.  I
use the term `revealing' for the process of recovering the implicit constituent
using derivation rewriting).  A constituent is a 4-tuple:
\[\mbox{\tuple{Category, Rule, LeftChild, RightChild}}\] where LeftChild and
RightChild are normally constituents, and Rule is the name of a combinatory as
listed in \refsec{sec:grammar}.  When a constituent is a single word, Rule is
{\em lex}, LeftChild holds the actual word, and RightChild holds the
place-holder --.  There is a special rule, {\em init\/} which is used in the
initial state of the parser.  It is discussed in
\refsec{sec:ControlStructure}. The Semantic Term List holds the list of
semantic terms associated with a constituent.  In case the Buffer contains more
than one constituent, the Semantic Term List is the concatenation of the term
lists of those constituents.\footnote{Given this representational system, it is
logically possible that there be two terms in the term list which originate
{}from different constituents, thus having no semantic indices in common.
Subsequently, when the two constituents are combined, unification could cause
two such distinct indices to become identical.  Curiously, such a phenomenon
does not arise in the grammar and semantics of the current system.  That is,
whenever two constituents do not combine, it is never the case that they both
introduce semantic terms over semantic indices which will subsequently be
unified.  If this property remains in more comprehensive grammars it provides
opportunities for certain monotonicity-related inferences whose consequences
require further research.}

The interpreter may read the Semantic Term List, but not modify it.  It records
its results (e.g.\ pronoun resolution) in the Interpreter Annotations
component.  The interpreter records its assessment of the sensibleness of the
analysis in the Penalties component.  This component has two parts: the penalty
list which enumerates the particular penalties associated with the state, and a
score which is determined from the penalty list and is used for comparing the
current analyses.

\section{Control Structure}
\label{sec:ControlStructure}

When the system encounters a string, the following top-level control algorithm
is executed.

\vbox{
\begin{tabbing}
\SetMyTabs
Start with one initial state S\SUB{init} where\\
\> S\SUB{init}'s buffer is the single constituent 
\tuple{tls(T,+,C):X/eop:X/s(T,+,C):X,init,--,--}\\
\> S\SUB{init}'s semantic term list, interpretations, and penalties are all 
empty\\
For each word $W$ in the input\\
\> For each lexical entry \tuple{$W$,$Cat$,$Sem$}\\
\> \> For each current state $S$\\
\> \> \> Make a copy $S'$ of $S$\\
\> \> \> Add the constituent \tuple{$Cat$,$lex$,$W$,--} to the Buffer of 
$S'$\\
\> \> \> Append Sem to the Semantic Term List of $S'$\\
\> \> \> For each way $S''$ of nondeterministically applying the rules of 
                                grammar to $S'$ (\refsec{sec:BUPAlg})\\
\> \> \> \> If the resulting buffer is an admissible one 
(\refsec{sec:admissible}) then\\
\> \> \> \> \> For each way of interpreting $S''$ 
(\refsec{sec:interpretation})\\
\> \> \> \> \> \> Compute the penalty of the interpretation\\
\> \> \> \> \> \> Save $S''$ unless subsumed by an extant state\\
\> \> \> Remove $S$\\
\> Perform discarding procedure on the current set of states 
(\refsec{sec:Penalties})\\ Continue with the next word\\ Of the states whose
buffer has the singleton constituent whose category is a tls(\UL,\UL,\UL),\\
\>  display the most sensible state or states (i.e.\ the one(s) 
with the least penalty).
\end{tabbing}
}

The category in the initial state has as its result the special symbol tls,
{\em top level sentence,} which is not mentioned elsewhere in the grammar.
This symbol is introduced mostly for convenience and should be thought of as
identical to the symbol s.  The difference will be ignored in the exposition
whenever possible.  The category has as its first argument the basic category
\mbox{s(\UL,+,\UL):X}, which creates the `expectation' for a tensed sentence.

\section{Bottom-Up Reduce Algorithm}
\label{sec:BUPAlg}

The nondeterministic reduce computation is as follows:

\begin{tabbing}
\SetMyTabs
reduce(state S) = \\
\> either\\
\> \> S as is\\
\> or\\
\> \> if there is a reduce step that can be applied to the buffer of S\\
\> \> then perform this step and recursively call reduce on the resulting 
state.\\
\> or \\
\> \> let RC be the rightmost constituent of the buffer of S \\
\> \> if \=the category of RC is of the form \UL/Z\\
\> \> \> where Z matches a zero morpheme (e.g.\ eop:X)\\
\> \> then \\
\> \> \> append the constituent \tuple{Z,lex,$\epsilon$,--} to the right of 
the buffer\\
\> \> \> append the semantic term list associated with that zero morpheme to
S's 
semantics\\
\> \> \> recursively call reduce on the resulting state\\
\> \> end if\\
\> end
\end{tabbing}

There are two ways of performing a single reduce step: (as discussed in
\refch{ch:parsing})

\begin{tabbing}
\SetMyTabs
let X, Y be the two rightmost constituents in the buffer of S\\
let XC and YC be the syntactic categories of X and Y, respectively \\
method 1:\\
\> if there is a combinatory rule R of the form XC + YC \ $\longrightarrow$
 \ Z\\
\> then \\
\> \> replace X and Y in the buffer with the constituent \tuple{Z,R,X,Y}\\
\> \> if rule R has any semantic terms \\
\> \> then append these terms to the semantic term list of S\\
method 2:\\
\> If YC is of the form W\?W\\
\> then\\
\> \> let XNF be the right normal form of X\\ 
\> \> if there exists a right subconstituent RS of XNF such that\\
\> \> \> the syntactic category of RS is RC and\\
\> \> \> there is a combinatory rule R of the form RC + YC \
 $\longrightarrow$ \ Z\\
\> \> then \\
\> \> \> replace RS by \tuple{Z,R,RS,Y}\\
\> \> \> if rule R has a nonempty semantic term list \\
\> \> \> then append this list to the semantic term list of S
\end{tabbing}

\section{Buffer Admissibility Condition}
\label{sec:admissible}

As discussed in \refch{ch:parsing} (especially sections
\ref{sec:Ungrammaticality} -- \ref{sec:HeavyIncremental}) the adult
listener/reader has access to a procedure which identifies and discards
unviable buffers such as \mbox{[the:DET insults:VERB]}.  For the purposes of
this project, I circumvent the step of acquiring this procedure by stipulating
the condition in \ref{AdmissibilityCondition}, which is adequate for the
grammar I use.

\startxl{AdmissibilityCondition}
Buffer Admissibility Condition\\ For every pair of adjacent constituents whose
categories are X and Y \\ 1.  No obligatory combinatory rule exists which can
combine X and Y, and\\ 2.  the categories of X and Y are ultimately combinable
\stopx

All combinatory rules are {\em obligatory\/} except those forward rules
(\forw$n$) where the left category is \mbox{\UL/(\UL/np:\UL)}, i.e.\ those
rules which determine filler-gap relations as discussed in
\refsec{sec:PictureNouns}.

X and Y are {\em ultimately combinable\/} in case either \ref{UHMatch} or
\ref{BackDollar} or
\ref{RevealDollar} holds.

\startxl{UHMatch}
X is of the form \UL/(Z/\UL\SSUB{1}\ldots\UL\SSUB{m}) and\\
Y is of the form Z/\UL\SSUB{1}\ldots\UL\SSUB{n}\\
for some m, n $\geq 0$ and some category Z.
\stopx

\startxl{BackDollar}
Y is of the form A\?B/\UL\SSUB{1}\ldots\UL\SSUB{n} and\\
there is a combinatory rule which can combine X and A\?B
\stopx

\startxl{RevealDollar}
Y is of the form A\?B/\UL\SSUB{1}\ldots\UL\SSUB{n} and\\
the right normal form of X has a right subconstituent RS\\
such that there is a combinatory rule which which can combine RS with A\?B.
\stopx

Conditions \ref{BackDollar} and \ref{RevealDollar} anticipate applications of
certain backward combinatory rules.  In \refsec{sec:HeavyIncremental} I argued
that semantic terms (in particular a term for marking crossing composition ---
a signal for heavy-NP shift) which would be introduced by the anticipated rule
application should be detected immediately and not delayed until the rule is
actually applied.  This is realized in the implementation.

\section{Interpretation}
\label{sec:interpretation}

The interpretive component in the current system performs only two of the many
interpretive functions of its human counterpart.  It performs a simplistic
database-lookup operation for resolving definite noun phrases against the prior
discourse (without any so-called bridging inferences, see Haviland and Clark
1974\nocite{HavilandClark74}.) It also implements a trivial form of
plausibility/implausibility inference --- relying on a hand-coded database of
implausible scenarios.  These `inferences' are of interest, of course, only
insofar as their contribution to the evaluation of competing analyses.

\subsection{Real World Implausibility}
\label{sec:Implausibility}

Minimal pairs such as 

\startxl{PoemPoetP}
a. The poet read in the garden stank.\\
b. The poem read in the garden stank.
\stopx

(where \ref{PoemPoetP}a.\ is a garden path but not b.) demonstrate the reliance
of the processor on world-knowledge inferences (see
\refsec{sec:WeaklyInteractive}).  It does not follow, of course that {\em
all\/} ambiguities which can be resolved by inference are indeed thus resolved
online.  One could set up arbitrarily complex puzzles the solutions of which
are crucial for resolving a particular ambiguity.  An account of which
inferences are sufficiently fast so as to direct online ambiguity resolution is
far outside the scope of the current work.\footnote{See Shastri and Ajjanagadde
(1992)\nocite{ShastriAjjanagadde92} for one view of `fast' inference.} For the
purposes of the current project, I assume that such an inferential device
exists and is able to quickly notice certain `obvious' semantic incongruities
and alert the interpreter.  One could think of the N400 signal in
electroencephalograms (Garnsey \etal\ 1989\nocite{GarnseyEtal89}) as a
correlate of the human analog of this incongruity alert.  I simulate the
behavior of such an anomaly detector by anticipating each anomalous situation
which will be encountered by the system and encoding that situation by hand.  A
partial list of these situations is as follows: (S is the semantic variable of
the implausible scenario)

\centerline{
\begin{tabular}{ll}
scenario description             & explanation \\ \hline$$
[read(S,X), poem(X)]             & Poems can't read.\\$$
[read(S,X,\UL), poem(X)]         & Poems can't read anything.\\$$
[warn(S,\UL ,X,\UL ), poem(X)]   & One can't warn poems.\\$$
[stop(S,X,\UL ),poem(X)]         & Poems can't stop anything\\$$
[future(S), yesterday(S)]        & Anything that happened yesterday is 
not in the future
\end{tabular}}

\subsection{Definite Reference}
\label{sec:DefiniteReference}

In the current system, all definite NPs --- pronouns, names, and NPs with
definite determiners --- have uniform semantic representations: A segment of
the semantic term list which begins with the term the(X), ends with the term
phrase\UL closed(X).  Between these markers lie semantic terms.  Here are some
examples:

\centerline{\begin{tabular}{|l|l|} \hline
phrase & semantic term list \\ \hline
the poem & [the(X), poem(X), phrase\UL closed(X)]\\
she  &  \parbox[t]{4.29in}{[the(Y), third\UL pers(Y), feminine(Y), 
singular(Y),\\
                          \hbox{\ phrase\UL closed(Y)]}}\\
john & [the(Y), name\UL of(Y,john), phrase\UL closed(Y)] \\ 
his poem & \parbox[t]{4.29in}{[the(X), third\UL pers(X), masculine(X), 
singular(X),\\
                             \hbox{\ phrase\UL closed(X), the(Y), of(Y,X), 
poem(Y),
                             phrase\UL closed(Y)}]} \\
\parbox[t]{1in}{the poem that john likes} & 
           \parbox[t]{4.29in}{[the(Y), poem(Y), npmod(Y),
                            the(Z), \hbox{name\UL of(Z,john)},\\ \hbox{\ 
phrase\UL closed(Z)},
                            like(W,Z,Y), tns(W,s), 
\hbox{phrase\UL closed(W)},\\ 
                             \hbox{\ phrase\UL closed(Y)}]} \\
\hline
\end{tabular}}

Terms such as name\UL of(X,Y) and poem(X) are called restrictive.  Others, such
as phrase\UL closed(X) and the(X) are non-restrictive, as they do not serve to
narrow down the set of possible referents.  It is assumed that all modifiers
are restrictive, i.e.\ non-attributive.

The algorithm for resolving definite reference is in \reffig{RefAlg}.

\begin{figure}
\begin{tabbing}
XX \= XX \= XX \= XX \= XX \= XX \= XX \= \kill
Given a database D representing the entities of the prior discourse and 
relations among them\\
and given a state S \\
\> with semantic term list SEM,\\
\> interpreter annotation list IA, and\\
\> penalty list P\\
Scan SEM from right to left \komment{SEM's atoms reflect the order 
of the input string} \\
For each occurrence O of the(X)\\
\> if accom(X,\UL) or resolved(X,\UL) is in IA then do nothing 
\komment{Already processed.}\\
\> else \\
\> \> let SEM$'$ be the final segment of SEM which begins with O\\
\> \> let Q be the query derived by conjoining all the restrictive atoms
 of SEM$'$\\
\> \> if the Q is empty\\
\> \> then do nothing 
      \komment{Don't look for a referent of a phrase still missing its 
lexical head}\\
\> \> else \\
\> \> \> let C be the set of values for X for which Q succeeds on D\\
\> \> \> if C is empty then\\
\> \> \> \> if the term phrase\UL closed(X) appears in SEM\\
\> \> \> \> then add accom(X,Q) to IA\\
\> \> \> \> else add accom\UL complex\UL description(X) to P\\
\> \> \> \> end-if\\
\> \> \> else if $\|$C$\|=1$\\
\> \> \> \> add resolved(X,C$'$) to IA, where C$'$ is the element of C\\
\> \> \> \> if the term phrase\UL closed(X) does not appear in SEM\\
\> \> \> \> then add overspecified\UL ref(X) to P\\
\> \> \> \> end-if\\
\> \> \> else if $\|$C$\|>1$ then\\
\> \> \> \> if the term phrase\UL closed(X) appears in SEM \\
\> \> \> \> then \\
\> \> \> \> \> let C$'$ be an arbitrary member of C\\
\> \> \> \> \> add resolved(X,C$'$) to IA\\
\> \> \> \> \> add underspecified\UL ref(X) to P\\
\> \> \> \> end-if\\
\> \> \> end-if\\
\> \> end-if\\
\> end-if\\
end for
\end{tabbing}
\caption{Definite Reference Resolution Algorithm}
\label{RefAlg}
\end{figure}

Some illustrations will make this algorithm's operations clear.

1. Suppose that \ref{shown} is encountered out of context.
\startxl{shown}
The horse shown to the poet fell.
\stopx

When the first word, `the' is processed the state has the semantics
[subj(e2),the(e2)].  Since there are no restrictive semantic
atoms\footnote{Recall that subj(X) is introduced by the subject type-raising
rule.  It is not a restrictive semantic atom.} the algorithm does nothing.  The
next word, `horse' introduces a syntactic ambiguity --- is the phrase `the
horse' closed or not?\\

\vbox{
state (i)\footnotemark\\
\BN{s:e1/(s:e1\?np:e2)}
   {\BN{s:e1/(s:e1\?np:e2)/eop:e2}
       {\LN{the}{s:e1/(s:e1\?np:e2)/eop:e2/n:e2}}
       {\LN{horse}{n:e2}}
       {\forw 0}}
   {\LN{$\epsilon$}{eop:e2}}
   {\forw 0}\\[2mm]
[subj(e2), the(e2), horse(e2), phrase\UL closed(e2)]\\[3mm]
state (ii)\\
\BN{s:e1/(s:e1\?np:e2)/eop:e2}
   {\LN{the}{s:e1/(s:e1\?np:e2)/eop:e2/n:e2}}
   {\LN{horse}{n:e2}}
   {\forw 0}\\[2mm]
[subj(e2), the(e2), horse(e2)]\\
}
\footnotetext{I number the states solely for ease of reference.}

In state (i), the parser nondeterministically chose to close the NP.  The
discourse representation is queried to find all things X which match the query
horse(X).  Since the discourse representation is empty, the result of this
query is the empty set.  The following annotation is therefore added to the
state's Interpreter Annotations List: accom(e2,[horse(e2)]).  No penalties
apply.  In state (ii), the parser chose not to close the NP.  The penalty
accom\UL complex\UL description(e2) is added to the state's penalty list, since
the state's buffer encodes a commitment to restrictive postmodifiers for the
NP.

The next word, `shown' resolves the closure/nonclosure ambiguity, as it
triggers a restrictive, reduced relative clause.  When the reduced relative
clause is finished, again, there is a closure ambiguity, as follows:\\

\vbox{
state (iii)\\
\BN{s:e1/(s:e1\?np:e2)}
   {\BN{s:e1/(s:e1\?np:e2)/eop:e2}
       {\LN{the}{s:e1/(s:e1\?np:e2)/eop:e2/n:e2}}
       {\BN{n:e2}
           {\LN{horse}{n:e2}}
           {\LN{shown to the poet}{n:e2\?n:e2}}
           {\back 0}}
       {\forw 0}}
   {\LN{$\epsilon$}{eop:e2}}
   {\forw 0}\\[2mm]
[subj(e2), the(e2), horse(e2), show(e3,e4,e2), tns(e3,en), npmod(e2), 
to(e3,e5), the(e5),
poet(e5), phrase\UL closed(e5), phrase\UL closed(e2)]\\[3mm] 
}

\vbox{
state (iv)\\
\BN{s:e1/(s:e1\?np:e2)/eop:e2}
   {\LN{the}{s:e1/(s:e1\?np:e2)/eop:e2/n:e2}}
   {\BN{n:e2}
       {\LN{horse}{n:e2}}
       {\LN{shown to the poet}{n:e2\?n:e2}}
       {\back 0}}
   {\forw 0}\\[2mm]
[subj(e2), the(e2), horse(e2), show(e3,e4,e2), tns(e3,en), npmod(e2), 
to(e3,e5), the(e5),
poet(e5), phrase\UL closed(e5)]
}

State (iii) gets the interpreter annotation 
\[\mbox{accom(e2,[horse(e2), show(e3,e4,e2), tns(e3,en), to(e3,e5), 
poet(e5)])}\] (ignoring the independent processes of resolving the NP `the
poet'.)  State (iv) is not yet closed, so it does not get this accommodation
annotation.  Instead it gets another accom\UL complex\UL description(e2)
penalty, which is subsequently removed by a duplicate removal procedure.

The presence of the main verb `fell' disambiguates the closure question, this
time by selecting the closed state, (iii).

2. Suppose the prior discourse contains two horses, introduced, for example by
the passage

\begin{quote}
There were two horses being shown to a prospective buyer.  One was raced in the
meadow and the other was raced past the barn.
\end{quote}

In this context, the interpretation of \ref{shown} proceeds differently.  After
encountering the first two words, `the horse', the parser constructs states (i)
and (ii) above.  The query of horse(X) now returns two possible candidates,
call them horse1 and horse2.  State (i), in which the NP is marked as closed,
is incapable of acquiring additional information to identify a unique referent
for `the horse'.  The interpreter then chooses one of these arbitrarily, say
horse1, and adds the annotation resolved(e2,horse1) to the interpreter
annotations of state (i).  Noting this premature choice, it adds the penalty
\mbox{underspecified\UL ref(e2)} to the state's penalty list.  State (ii) 
is not closed, so the algorithm decides to wait for additional individuating
information.

The next few words which the processor encounters are `shown to the poet'.
When interpreting state (ii), the interpreter decided to wait for information
to distinguish horse1 from horse2.  But this restrictive clause is
infelicitous.  It refers to a poet which is not in the current discourse and
must be accommodated.  When the algorithm applies the query

\begin{quote}
horse(X) $\wedge$ show(Y,Z,X) $\wedge$ tns(Y,en) $\wedge$ to(Y,P) $\wedge$
poet(P)
\end{quote}

it finds no matching candidates for the variable X.  As in 1.\ above, the
interpreter adds an annotation accommodating the definite description.  Also,
it applies the penalty {accom\UL complex\UL \mbox{description(e2).}}

Had the restrictive relative clause been appropriate, e.g.\ had the sentence
been

\startxl{RacedFell}
The horse raced past the barn fell
\stopx

The set of discourse elements satisfying the query

\begin{quote}
horse(X) $\wedge$ race(Y,Z,X) $\wedge$ tns(Y,en) $\wedge$ past(Y,P) $\wedge$
barn(P)
\end{quote}

would have been the singleton set horse2.  The algorithm would add the
annotation re\-solved(e2, horse2) to the interpreter annotations list and apply
no penalties.

\section{Detecting the End of a Phrase}
\label{sec:eop}

In this section I provide the rationale for the end-of-phrase mechanism used in
the current implementation.

The definite reference resolution algorithm relies on the accurate signaling of
the end of an NP.  Without the ability to identify the boundaries of a noun
phrase, the processor would be unable to distinguish from the various
assertions made of a semantic variable those which are within the scope of the
determiner, from those which are not.

Since this algorithm fits squarely within the interpretation module, it does
not have direct access to the syntactic representation, so identification of
the end of the phrase cannot be simply performed by checking that a particular
node or constituent is no longer on the `right frontier' of the emerging
analysis.  The detection of an end of a noun phrase must therefore be
identified by the syntactic processor and passed to the interpreter using the
only available data-path, namely the sense-semantics.  Given the tremendous
variation of NP structure in the world's languages it is natural to place the
burden of end-of-phrase detection with the language-particular grammar, not
with the processor in general.

How can can phrase-boundary be implemented in a CCG?  In semantics of the usual
sort, where a constituent is assigned a meaning term or a `logical form', the
mechanism of (quantifier) scope is available, and nothing special is required.
However, in the semantic-term-list approach which I have adopted here (see
\refsec{sec:SemanticRepresentation}) scope is rather difficult to express in
the sense-semantics.  One way to implement phrase-closure-detection in CCG is
to disallow recursive postmodification of NPs and simply state in advance, in
the lexical entry for a determiner or a noun, exactly what the constituents of
the NP are.  This is rather awkward, and may well be missing the generalization
that post-head adjectival apply recursively\footnote{It is not clear to me
whether restrictive adjectivals really can recurse, but they are commonly
assumed to do so.}. The other way us to use the narrowly constrained
zero-morpheme scheme as I have presented above.  I use the same zero morpheme
(eop) for clauses as well.  This move is not forced by anything, and is adopted
mostly for uniformity.  It happens to play a convenient role in avoiding
certain shortcomings which would otherwise arise from the way revealing is
implemented in Prolog.

\section{An Example}
\label{sec:example}

The processor consists of the components discussed above --- competence
grammar, control structure, parsing algorithm, and interpreter --- as well as
state-adjudication algorithm.  Before turning to the details of this final
component, it would be best to illustrate the operation of the processor so far
with an example.  In this example, state-adjudication should be thought of as
working out by magic.  In \refsec{sec:Penalties} I present a decision procedure
for it.

Let us begin with the string

\startxl{PoetRead}
The poet read in the garden stank.
\stopx

encountered out of context.

Before any words are processed, the parser starts with one initial state whose
buffer has one constituent:

\tuple{tls(T,+,C):E/eop:E/s(T,+,C):E,lex,init,--}

The first word, `the' is encountered.  It has two lexical entries,
corresponding to the original determiner category, and the subject-type-raised
determiner, respectively.

np(3,s1):e2/eop:e2/n(N):e2 \\
s(s2,+,0):e1/(s(s2,+,0):e1\?np(3,s1):e2)/eop:e2/n(s1):e2

The nondeterministic reduce algorithm results in three states:

\vbox{
state 1: \ (the initial category and the non-type-raised category for the
determiner)\\
\ \ \LN{{\em init}}{tls(T,+,C):E/eop:E/s(T,+,C):E} \ \ 
\LN{the}{np(3,s1):e2/eop:e2/n(N):e2}\\
}

\vbox{
state 2:  \ (the initial category and the type-raised category 
for the determiner)\\
\ \ \LN{{\em init}}{tls(T,+,C):E/eop:E/s(T,+,C):E} \ \ 
    \LN{the}{s(s2,+,0):e1/(s(s2,+,0):e1\?np(3,s1):e2)/eop:e2/n(s1):e2}\\
}

\vbox{
state 3: \ (initial category and type-raised determiner, combined)\\
\ \ \BN{tls(s2,+,0):e1/eop:e1/(s(s2,+,0):e1\?np(3,s1):e2)/eop:e2/n(s1):e2}
       {\LN{{\em init}}{tls(T,+,C):E/eop:E/s(T,+,C):E}}
       {\LN{the}{s(s2,+,0):e1/(s(s2,+,0):e1\?np(3,s1):e2)/eop:e2/n(s1):e2}}
       {\forw 2}
}

State 1 is ruled out by the second clause of the Buffer Admissibility Condition
\ref{AdmissibilityCondition} which requires adjacent constituents to be 
ultimately combinable.  State 2 is ruled out by the first clause of
\ref{AdmissibilityCondition} which requires that adjacent constituents not be
immediately combinable.  State 3 is therefore the only one which the parser
outputs.  Since it does not have its head noun yet, the interpreter does not
add any interpretations or penalties to this state.  For the rest of this
example, I ignore the initial state, and pretend that the current state has the
category

s(s2,+,0):e1/(s(s2,+,0):e1\?np(3,s1):e2)/eop:e2/n(s1):e2.

I also omit parser states which are ruled out by the Buffer Admissibility
Condition.

The next word, `poet' is encountered.  It gives rise to

state 4:  \\
\zp Buffer: s(s1,+,0):e1/(s(s1,+,0):e1\?np(3,sg):e2)/eop:e2\\
\zp Semantics: [subj(e2), the(e2), poet(e2)].

The parser also nondeterministically posits a zero morpheme following `poet'
yielding\\

\vbox{
state 5:  \\
\zp Buffer: s(s1,+,0):e1/(s(s1,+,0):e1\?np(3,sg):e2)\\
\zp Semantics: [subj(e2), the(e2), poet(e2), phrase\UL closed(e2)].
}

In state 5, Because the definite phrase e2 is closed, the interpreter
accommodates a poet.  In state 4, the interpreter anticipates further
restrictive modifiers, so it penalizes the state for having to accommodate a
complex NP.  The results are \\

\vbox{
state 4:  \\
\zp Buffer: s(s1,+,0):e1/(s(s1,+,0):e1\?np(3,sg):e2)/eop:e2\\
\zp Semantics: [subj(e2), the(e2),  poet(e2)]\\
\zp Interpretation: [\ ]\\
\zp Penalties: [accom\UL complex\UL description(e2)]
}

\par

\vbox{
state 5:\\
\zp Buffer: s(s1,+,0):e1/(s(s1,+,0):e1\?np(3,sg):e2)\\
\zp Semantics: [subj(e2), the(e2),poet(e2), phrase\UL closed(e2)]\\
\zp Interpretation: [accom(e2,[poet(e2)])]\\
\zp Penalties: [\ ]}

Despite the penalty in state 4, both states are maintained, for now.  Also,
both states 4 and 5 incur a penalty for having a new NP `the poet' in subject
position.  Because this penalty will apply to every state in the rest of this
example, it will turn out to be irrelevant, so I omit it.

The next word `read' is many-ways ambiguous.  The untensed verb reading and the
present tense non-3rd-person-singular reading are ruled out because their
features conflict with the \mbox{s(\UL,+,0)\?np(3,sg)} requirement of the
subject NP category.  Three readings remain: past-tense intransitive,
past-tense transitive, and past-participle acting as head of a reduced relative
clause.  The first two combine with state 5 to yield states 6 and 7,
respectively.  The third is added to state 4 to yield state 8.

\State{6\footnotemark}
 {[s(ed,+,0):e1]} {[subj(e2), the(e2), poet(e2), phrase\UL closed(e2),
read(e1,e2,e3), tns(e1,ed)]} {[accom(e2,[poet(e2)])]} {[\
]}\footnotetext{Recall that the category of this state is actually
tls(ed,+,0):e1/eop:e1.  So in addition to state 6, the processor constructs
state 6$'$ where it posits an end of phrase morpheme signaling the end of the
main clause.  This state has no continuation and it disappears when the next
word is encountered.}
\State{7}
 {[s(ed,+,0):e1/np(s1,s2):e3]}
 {[subj(e2), the(e2), poet(e2), phrase\UL closed(e2), read(e1,e2,e3), 
	tns(e1,ed)]}
 {[accom(e2,[poet(e2)])]}
 {[\ ]}

\State{8}
 {[s(s1,+,0):e1/(s(s1,+,0):e1\?np(3,sg):e2)/eop:e2,\\ 
  \mbox{\ n(sg):e2\?n(sg):e2/(s(s2,s3,0):e6\?s(s2,s3,0):e6)]}}
 {[subj(e2), the(e2), poet(e2), read(e6,e5,e2), tns(e6,en), npmod(e2)]}
 {[\ ]}
 {[accom\UL complex\UL description(e2)]}

Notice that state 8 satisfies clause \ref{RevealDollar} of the Buffer
Admissibility Condition.  That is, the category n is revealed from the
right-hand edge of the first constituent, `the poet', and this category may be
modified by the second constituent `read', when the latter has received all of
its arguments, namely the adverbial `in the garden'.

The next word, `in' is four-way ambiguous, as shown in the table in
\refsec{sec:grammar}.  Of these only one category, sentential post-modifier, is
not ruled out by the buffer admissibility condition.  States 6, 7, and 8, then,
become states 9, 10, and 11, respectively.

\State{9}
 {[s(ed,+,0):e1,~~s(ed,+,0):e1\?s(ed,+,0):e1/np(s1,s2):e4]}
 {[subj(e2), the(e2), poet(e2), phrase\UL closed(e2), 
read(e1,e2,e3), tns(e1,ed), in(e1,e4)]}
 {[accom(e2, [poet(e2)])]}
 {[\ ]}

\State{10}
 {[s(ed,+,0):e1/np(s1,s2):e3,~~s(ed,+,0):e1\?s(ed,+,0):e1/np(s3,s4):e4]}
 {[subj(e2), the(e2), poet(e2), phrase\UL closed(e2), 
read(e1,e2,e3), tns(e1,ed), in(e1,e4),
   \mbox{h\UL shifted(e3,e1)}]} 
 {[accom(e2, [poet(e2)])]}
 {[shifted\UL past\UL non\UL given(e1)]}

\State{11}
 {[s(s1,+,0):e1/(s(s1,+,0):e1\?np(3,sg):e2)/eop:e2,
	~~n(sg):e2\?n(sg):e2/np(s2,s3):e3]}
 {[subj(e2), the(e2), poet(e2), read(e6,e5,e2), tns(e6,en), 
	npmod(e2), in(e6,e3)]}
 {[\ ]}
 {[accom\UL complex\UL description(e2)]}

State 10 incurs a penalty for heavy NP shift past material which is not given
in the discourse, (see \refsec{sec:HeavyIncremental}.  States 10 and 11 are
discarded because while they each carry a penalty, state 9, does not.  By
discarded state 11, the processor has resolved the main-verb/reduced-relative
ambiguity of `read', selecting the main-verb analysis.  By discarding state 10,
it has further committed to the intransitive use of this verb.
\ifprediction
The consequence of the latter commitment will be discussed in
\refsec{sec:prediction}.
\fi

The word `the' yields state 12 from state 9.

\State{12}
 {[s(ed,+,0):e1,~~s(ed,+,0):e1\?s(ed,+,0):e1/eop:e4/n(s1):e4]}
 {[subj(e2), the(e2),  poet(e2), phrase\UL closed(e2), read(e1,e2,e3), 
	tns(e1,ed), in(e1,e4), the(e4)]}
 {[accom(e3, [poet(e3)])]}
 {[\ ]}
The word `garden' leads to the familiar closure ambiguity in states 13 and 14.

\State{13}
 {[s(ed,+,0):e1,~~s(ed,+,0):e1\?s(ed,+,0):e1/eop:e4]}
 {[subj(e2), the(e2), poet(e2), phrase\UL closed(e2), read(e1,e2,e3), 
	tns(e1,ed), in(e1,e4),
   the(e4), garden(e4)]} 
 {[accom(e2,[poet(e2)])]}
 {[accom\UL complex\UL description(e4)]}

\State{14}
 {[s(ed,+,0):e1]}
 {[subj(e2), the(e2), poet(e2), phrase\UL closed(e2), read(e1,e2,e3), 
	tns(e1,ed), in(e1,e4),
   the(e4), garden(e4), phrase\UL closed(e4)]}
 {[accom(e4,[garden(e4)]), accom(e2,[poet(e2)])]}
 {[\ ]}

But neither state is compatible with the next word, `stank'.  
	The lack of surviving states
indicates the garden path effect of the sentence.

\section{A Consistent Theory of Penalties}
\label{sec:Penalties}

I now construct an adjudication algorithm --- a procedure to evaluate the set
of processor states and based on the kind and number of penalties which they
have, decide which, if any, to discard.  I begin by considering the examples
listed in \refsec{sec:desiderata}, the penalties that the processor assigns
each analysis, and the appropriate action at each moment.  I then present one
of many possible algorithms that fit these data.

\subsection{Desired Behavior}
\label{sec:PenaltyData}

For convenience, I refer to penalties by their number, as follows:

\begin{center}
\begin{tabbing}
1. \    \= implausibility\\
2.      \> underspecified\UL ref\\
3.      \> overspecified\UL ref\\
4.      \> accom\UL complex\UL description\\
5.      \> new\UL subject \\
6.      \> heavy\UL arg\UL light\UL modifier\\
7.      \> shifted\UL past\UL non\UL given\\
\end{tabbing}
\end{center}

{\newcommand{\Chunk}[1]{\vbox{\begin{tabbing}XX \= XXXXXXXXX \= \kill #1 
\end{tabbing}}}
\Chunk{
XX \= XXXXXXXXX \= \kill
One wife context:\\
1. \>             \> The psychologist told the wife\=\ that he disliked 
Florida\=.\\
\> that = relativizer    \>                        \>3                      
      \\
\> that = complementizer \>                        \>                       
  \>\ ok\\
\\
2. \>            \> The psychologist told the wife\=\ that he disliked 
that\=\ he liked Florida.\\
\> that = relativizer     \>                      \>3                    
  \>\ gp \\
\> that = complementizer  \>                                       
}
\Chunk{
two wife context:\\
3. \>            \> The psychologist told the wife\=\ that he disliked that 
he liked Florida\=.\\
\> that = relativizer    \>                        \>			
\>\ ok\\
\> that = complementizer \>                        \>2                     
\\
\\
two wife context:\\
4. \>            \> The psychologist told the wife\=\ that he disliked 
that\=\ he liked Florida.\\
\> that = relativizer    \>                                          
        \\
\> that = complementizer \>                        \>2                 
     \>\ gp
}
\Chunk{
out of context:\\
5. \>            \> The poet\=\ read in the garden\=\ stank.\\
\> main verb \> \\
\> reduced relative  \>      \>  4                 \>\  gp\\
\\
6. \>            \> The poem\=\ read\=\ in the garden\=\ stank.\\
\> main verb        \>      \>   \>   1              \\
\> reduced relative \>      \>  4 \>               \>\   ok\\
\\
7. \>            \> The bird\=\ found in\=\ the nest\=\ a nice juicy worm.\\
\> reduced relative \>     \>4\\
\> main verb        \>     \>            \> 7       \>\  ok\\
\\
8. \>            \> The bird\=\ found in\=\ the nest\=\ died.\\
\> reduced relative \>       \> 4      \>         \> ok\\
\> main verb        \>       \>        \> 7       
}
\Chunk{
context:  what did the bird find in the nest?\\
9. \>            \> The bird\=\ found in the nest\=\ a nice juicy worm.\\
\> reduced relative \>      \> 3                  \\
\> main verb        \>      \>                   \>\   ok\\
\\
10. \>          \>  The bird\=\ found in the nest\=\ died.\\
\> reduced relative   \>     \> 3                 \>\ gp\\
\> main verb         \>                           
}
\Chunk{
context:  Fred found a bird in a nest and Bill found one in the garden.\\
11. \>           \> The bird\=\ found in the nest\=\ a nice juicy worm.\\
\> reduced relative \>                            \\
\> main verb        \>      \>   2         \>\         gp\\
\\
12. \>           \> The bird\=\ found in the nest died.\\
\> reduced relative \>      \>             \>\       ok\\
\> main verb        \>      \>   2                     
}
\Chunk{
out of context:\\
13. \>           \> Without her contributions\=\ dwindled.\\
\> determiner       \>            \\
\> object pronoun   \>                       \>\  5 gp\\
\\
14. \>           \> Without her contributions\=\ the charity failed.\\
\> determiner      \>                        \>\  ok\\
\> object pronoun  \>                        \> 5\\
\\
15. \>           \> The poet said that the psychologist fell yesterday\= .\\
\> low attachment   \>                                                \>\ ok\\
\> high attachment  \>                                                \>  6 \\
\\
16. \>           \> The poet said that the psychologist will fall 
yesterday\= .\\
\> low attachment    \>                                            
        \>   1\\
\> high attachment   \>                                             
       \>   6 gp-awkwardness
}
}

\subsection{Fitting The Data}
\label{sec:fitting}

The simplest conceivable state discarding algorithm is

\startxl{simplest}
If at least one state has no penalties\\
then discard every state that has one or more penalties
\stopx

There are two problems with \ref{simplest}.  The first has to do with the fact
that not all penalties are of equal strength: scenario 16 shows that penalty~6
is stronger than penalty~1.  Using similar reasoning, one can deduce the
following constraints among penalty-strengths: (s1 stands for the strength of
penalty~1)

\centerline{\begin{tabular}{|rc|} \hline
scenario & constraint provided \\ \hline
6       & $s1 \geq s4$ \\
7,8     & $s4 = s7$ \\
16      & $s6 > s1$ \\ \hline
\end{tabular}}

These constraints underdetermine the ranking of the penalties with respect to
strength.\footnote{In fact, I am making a great simplification by treating all
instances of a penalty as having the same strength.  For example,
implausibility is surely a graded judgement, as is the degree of complexity of
accommodation.} The following is one of many schemes which are consistent with
the constraints.  It uses two strength levels, the minimum possible.

\centerline{
\begin{tabular}{|cc|} 
\hline penalty & strength (in number of points)\\ \hline
1 & 1\\
2 & 1\\
3 & 1\\
4 & 1\\
5 & 1\\
6 & 2\\
7 & 1\\ \hline
\end{tabular}}

The second problem with \ref{simplest} is that of timing.  Scenarios 6 and 8
show that sometimes a state which has a penalty is not discarded, even when it
is competing with one that has none.  In these scenarios, information which
arrives one or two words after the first detection of a penalty (penalty~4) is
brought to bear and prevents discarding.  This is in contrast with scenario 13
where as soon as a penalty (penalty~5) is detected, the offending state is
discarded.  I let each penalty type carry a {\em grace period} --- an interval
of time.  When the penalty is detected, a countdown clock associated with it is
started.  The penalty is ignored until its clock reaches zero.

The scenarios above provide the following constraints on the assignment of
grace periods: (where g3 stands for `the grace period of penalty~3', measured
in words\footnote{Using the word as a measure of time is intended to be an
approximation.  Clearly the time course of processing function words is very
different from that of processing long or novel content words.  Given the
currently available psycholinguistic evidence, only a crude timing analysis is
possible at this time. (But see Trueswell and Tanenhaus
(1992)\nocite{TrueswellTanenhaus92} for some preliminary work at trying to
understand the time course of the interaction of competing penalties ---
`constraints' in their terms.)}.)

\centerline{
\begin{tabular}{|rl|} 
\hline scenario & constraint provided \\ \hline
2  &       $g3 < 4$\\
4  &       $g2 < 4$\\
5  &       $g4 < 4$\\
6  &       $g4 \geq g1 + 1$\\
7  &       $g4 \leq g7 + 2$\\
8  &       $g4 \geq g7 + 2$\\
13 &       $g5 = 0$\\ \hline
\end{tabular}}

No timing constraint is provided by scenario 16 because the interaction between
penalty~6 and 1 occurs at the end of the sentence.

These constraints again underdetermine the grace periods.  Here is one
solution, which minimizes the grace period values.

\centerline{
\begin{tabular}{|llcc|}
\hline
penalty & name                           & strength & grace period \\ \hline
1.      & implausibility                        & 2 & 0\\
2.      & underspecified\UL ref                 & 1 & 0\\
3.      & overspecified\UL ref                  & 1 & 0\\
4.      & accom\UL complex\UL description       & 1 & 2\\
5.      & new\UL subject                        & 1 & 0\\ 
6.      & heavy\UL arg\UL light\UL modifier     & 3 & 0\\
7.      & shifted\UL past\UL non\UL given       & 1 & 0\\ \hline
\end{tabular}}

The revised algorithm, then is 

\startxl{RevisedAlgorithm}
\begin{enumerate}
\item For each state, let its penalty score be the sum of the strengths 
of all penalties whose
grace periods have passed.
\item Find the minimum score.
\item Discard each state whose score is above the minimum.
\end{enumerate}
\stopx

It must be emphasized that the algorithm and parameters serve merely to
demonstrate the consistency of the set of penalties listed in the beginning of
this section; so the particular numbers, or even exactly what they measure,
should not be construed as a proposed theory.
  
\ifprediction 

\section{A Prediction}
\label{sec:prediction}

Despite the preliminary nature of the specifics of the state-discarding
algorithm, it is nevertheless possible to derive an interesting prediction from
the system as developed so far, in particular, from the interaction of the
choice of the theory of syntax and the state discarding procedure.

The account presented here assumes penalties for heavy shift that is
infelicitous in context, and for accommodating a complex NP.  Recall the
example in \refsec{sec:example}.  The verb `read' has three categories: a
reduced relative clause, and two main-verb categories: transitive and
intransitive. Consider what the account does when faced with each sentence in
\ref{dung} out of context.

\startxl{dung}
a. The poet read in the garden stank.\\
b. The poet read in the garden a lengthy article about Canadian earthworms.
\stopx

In \ref{dung}a, the complex NP accommodation penalty correctly excludes the
reduced relative analysis, resulting in a garden path.  What is the predicted
status of \ref{dung}b?  Given that the reduced-relative analysis is discarded,
one would expect the main-verb analysis to persist.  But CCG has two completely
separate `main-verb' analyses.  The transitive analysis requires heavy-NP
shift, which is deemed infelicitous out of context.  The surviving analysis is
of the intransitive verb, and cannot cope with the shifted NP.  So the account
predicts a garden path in \ref{dung}b. This prediction arises, of course,
because of the lexicalized nature of CCG: every combinatory potential of a word
is treated as a separate lexical entry.  In other words, CCG does not
distinguish small differences between categories (e.g.\ subcategorization) from
major differences (e.g.\ main verb versus reduced relative clause).

So while CCG predicts a garden path for \ref{dung}b, a more traditional,
phrase-structure theory of grammar might not, depending on whether it
distinguishes analyses on the basis of lexical subcategory.  The garden-path
status of \ref{dung}b is an empirical one, and necessitates teasing apart any
processing difficulty associated with the infelicity of the heavy NP shift from
truly syntactic/parsing effects indicative of a garden path.  It remains for
future research.

\else 
\fi   

\section{Summary}

Using the meaning-based criteria developed in chapters 2 through 4: referential
felicity, felicity with respect to givenness, plausibility; and the parsing
algorithm presented in chapter 5, I have presented a simple model of the
process of sentence comprehension.  The point of this demonstration is to show
that it is possible to construct a simple sentence processor which can account
for significant subset of the data available about when garden paths arise in
English.  This enterprise is largely successful: the data structures and
algorithms needed by top-level of the architecture are obvious and straight
forward.  Complexity arises from the specific requirements necessitated by the
grammar formalism, CCG, and by the scope of the state discarding criterion.
\ifprediction 
The latter is severely underdetermined
by the available data.

The long term goals of this work is to provide a detailed model of sentence
processing, one which makes clear and testable prediction.  While this is still
a long way off, I have nevertheless shown that already it is possible to make
some sort of predictions from the interaction of the various ingredients.
\else 
The latter is severely underdetermined by the available data, so very little
else can be said about it without additional research into the interaction of
the various penalties.
\fi 

The program, as described in this chapter, is written in Quintus Prolog, and is
called {\tt arfi}: Ambiguity Resolution From Interpretation.  It is accompanied
by a graphical user interface which provides an easy-to-use facility for
inspecting the execution trace of the processor on a particular input string.
The inspector program, called the {\tt viewer} is written for the X window
system and requires Common LISP and the software package {\tt CLIM.}  {\tt
arfi} and {\tt viewer} are have been available on the Internet by anonymous
FTP.  They are on the host ftp.cis.upenn.edu in the directory /pub/niv/thesis.

\chapter{Conclusion}
\label{ch:conclusion}

Of the class of computational functions performed by the human language
faculty, ranging from phonetics to the social activity of communication, I have
considered two

\begin{description}
\item[parsing] the application of the rules of syntax to identify the relations
among the
words in the input string
\item[interpretation] the updating of the hearer's mental model of the ongoing 
discourse based on the sense-semantic relations recovered by the parsing
process
\end{description}

I have argued for a particular view of the interaction between these two
functions.  First, I adopted the uncontroversial assumption (almost
uncontroversial, see Marslen-Wilson and Tyler
1987\nocite{Marslen-WilsonTyler87}) that parsing and interpretation occur in
separate modules, and that these modules interact through well-specified
channels: the parser sends nothing but sense-semantic representations to the
interpreter, and the interpreter sends the parser nothing but feedback about
sensibleness of the various analyses.

Second, I adopted the more controversial assumption that the parser computes
all of the grammatically licensed analyses of the string so far and sends them
all to the interpreter for evaluation, in parallel.  I claimed that the parser
does not provide its own ranking or evaluation of the analyses it constructs by
applying structurally stated preference criteria --- that all observable
preferences among ambiguous readings stem from principles of the linguistic
competence, principles such as Grice's maxims of quantity and manner for
evaluating the felicity of definite referring expressions, the competence
principle in English to place high information volume constituents after low
volume ones, to use subject position for encoding given information, etc.  I
did not explore in detail the question of whether the parsing component applies
{\em some\/} evaluation of the various analyses based on strengths of various
alternatives in the competence grammar.  While verb subcategorization
preferences (e.g.\ Ford, Bresnan and Kaplan 1982\nocite{FBK82}; Trueswell,
Tanenhaus and Kello 1993\nocite{TTK93}) can be ascribed either to the lexical
entries (part of grammatical competence) or a `deeper' representation of the
concepts attached to them, there are some preference phenomena which seem to
necessitate grammatically specified strength of preference (Kroch
1989\nocite{Kroch89}).  This issue remains for future research.

Third, I investigated the design of the parser.  My aim was to identify the
simplest design possible.  I investigated Steedman's (1992) thesis that
conceiving of syntactic knowledge of language as a Combinatory Categorial
Grammar (CCG) allows one to construct a parser that is significantly simpler
than what would be needed for traditional grammars, while still maintaining the
input-output behavior necessary to function as the syntactic parsing module in
the overall sentence-processing system design.  It turned out that designing a
parser for CCG runs into its own collection of complexities.  Certain of these
complexities (detection of inevitable ungrammaticality, detection of inevitable
word-order non-canonicality in the form crossing composition, identifying
optional combinations --- e.g.\ picture-noun extraction) can be elegantly
handled by assuming that the ability to parse in adults is composed of an
innate ability to put grammatical constituents together and an acquired skill
of quickly anticipating the consequences of certain combinations.  One final
complexity --- the interaction of CCG's derivational equivalence with the
incremental analysis necessary for timely interpretation --- necessitated
assuming that the history of the derivation is properly a component of a
`grammatical analysis' and augmenting the repertory of the parser with an
operation which explicates the interchangeability of derivational equivalent
analyses by manipulating the history of the derivation.\footnote{Note that it
is logically possible to define a more extreme condition on a parsimonious
parser.  This condition would disallow operations and representations which are
not strictly defined by the well-formedness rules of the grammar.  Since CCG
does not, strictly speaking, define well-formed analyses, only well-formed
constituents, and since it does not explicitly related equivalent derivations,
this view of grammar is not compatible with the derivation-rewrite algorithm I
have presented.}

Fourth, I have constructed a computational simulation of the sentence
comprehension process which allows one to evaluate the viability of the central
claim of the dissertation --- that the syntactic processor blindly and
transparently computes all grammatical analyses, and ambiguity resolution is
based on interpretation. This simulation serves as a computational platform for
experimenting with various analysis pruning strategies in the interpreter.  It
has shown that at the moment, the available psycholinguistic data greatly
underdetermines the precise strategy, but some empirical predictions do emerge.

the dissertation gives rise to \ifprediction three \else two \fi specific
empirical questions:

\begin{itemize}
\item Given the example dialog in \ref{Sabbatical} on \refpage{Sabbatical} 
which shows that
discourse status can affect perceived information volume, just how much of the
information volume, as operationalizable by observing attachment preferences,
can be accounted for by discourse factors, and how much of it is irreducible to
the form of the constituent --- the amount of linguistic material, and other
syntactic attributes such as grammatical category?
\item Does Disconnectedness theory play any role at all in ambiguity
resolution?  To what extent is Avoid New Subjects really responsible for the
data cited in
\refch{ch:leftovers}?  As discussed in detail in \refsec{sec:LCANS}, if one
were to re-run the second experiment reported by Stowe (1989), ruling out the
instrumental reading, and still get implausibility effects in the inanimate
condition, one would have an empirical basis to rule in Avoid New Subjects and
rule out Disconnectedness theory.
\ifprediction
\item Is CCG correct in its equal treatment of major category and subcategory 
distinctions?  That is, can the processor be made to commit to an intransitive
analysis for a verb, thus garden-pathing on its direct object, as predicted in
\ref{dung} on \refpage{dung}?
\else
\fi
\end{itemize}

\appendix

\chapter{Data from Avoid New Subject investigation}
\label{SubjAppendix}

\vbox{
\noindent Brown Corpus:

\centerline{\begin{tabular}{|l|rrrr|rrrr|}  \hline
                 & \multicolumn{4}{c|}{Subjects} &
 \multicolumn{4}{c|}{Non Subjects} \\ \hline
      givenness status &{\sc tc}&{\sc rc}&
{\sc tc+rc}&matrix&{\sc tc}&{\sc rc}&{\sc tc+rc}
   &matrix\\ \hline
  {\sc empty-category} &      0 &     50 &     50 &      0  
&      6 &     41 &     47 &      0 \\
         {\sc pronoun} &    773 &   1027 &   1800 &   7580  
&     79 &    134 &    213 &    956 \\
     {\sc proper-name} &    201 &     81 &    282 &   2838  
&     32 &     21 &     53 &    539 \\
        {\sc definite} &    890 &    266 &   1156 &   6686  
&    351 &    182 &    533 &   3399 \\
      {\sc indefinite} &    617 &    119 &    736 &   4157  
&    555 &    344 &    899 &   5269 \\
  {\sc not-classified} &    259 &    107 &    366 &   3301  
&    167 &     79 &    246 &   1516 \\
\hline    total:       &   2740 &   1650 &   4390 &  24562  
&   1190 &    801 &   1991 &  11679 \\
 \hline
\end{tabular}}}
\vspace{1ex}

\vbox{
\noindent Wall Street Journal Corpus:

\centerline{\begin{tabular}{|l|rrrr|rrrr|}  \hline
                 & \multicolumn{4}{c|}{Subjects} &
 \multicolumn{4}{c|}{Non Subjects} \\ \hline
      givenness status &{\sc tc}&{\sc rc}&{\sc tc+rc}&matrix&{\sc
tc}&{\sc rc}&{\sc tc+rc}
   &matrix\\ \hline
  {\sc empty-category} &      4 &     83 &     87 &      0  
&      2 &     20 &     22 &      1 \\
         {\sc pronoun} &    369 &   2263 &   2632 &   2347  
&     34 &     90 &    124 &    169 \\
     {\sc proper-name} &    167 &    371 &    538 &   3364  
&     29 &     89 &    118 &    377 \\
        {\sc definite} &    610 &   1686 &   2296 &   5385  
&    253 &    729 &    982 &   1959 \\
      {\sc indefinite} &    498 &    805 &   1303 &   3847  
&    484 &   1375 &   1859 &   4039 \\
  {\sc not-classified} &    251 &    278 &    529 &   2402  
&    178 &    581 &    759 &   2138 \\
\hline    total:       &   1899 &   5486 &   7385 &  17345  
&    980 &   2884 &   3864 &   8683 \\
\hline
\end{tabular}}}

(Non-zero cells of empty categories in non post-ZERO-complementizer subjects
are due to annotation errors in the corpus.)

\chapter{A Rewrite System for Derivations}
\label{DRSappendix}

In this appendix I define a formal system, DRS -- a rewrite system\footnote{For
an overview of rewrite systems, the reader is referred to Le Chenadec
(1988)\nocite{LeChenadec}, especially section 2.2.} for CCG derivations, as
sketched in \refsec{sec:Revealing}.  I then prove that DRS preserves the
semantics of a derivation, and show that it can form the basis of a correct and
efficient algorithm for computing the `right-frontier' of a derivation.

{\bf Definition} Two derivations are {\em equivalent} just in case the category
of their respective roots are equal.

I now give the definition of DRS.  \ DRS allows one to describe equivalence
classes of derivations and provide the means of picking out one representative
{}from each equivalence class.

Given the set $D$ of valid derivations, define the relation $ \arr \;
\subseteq \; D \times D$ to hold between a pair of derivations $(d1,d2)$ just
in case exactly one application of one of the derivation rewrite rules in
\ref{drs-forw} and \ref{drs-back} to some node in $d1$ yields $d2$.

Any subtree of a derivation which matches the left-hand-side of either
\ref{drs-forw} or \ref{drs-back} is called a {\em redex}.  The result 
of replacing a redex by the corresponding right-hand-side of a rule is called
the {\em contractum}.  A derivation is in {\em normal form (NF)\/} if it
contains no redex.

\newcommand{\whys}{\up Y\SUB{1} $\cdots$ \up Y\SUB{m-1}}
\newcommand{\zees}{\up Z\SUB{1} $\cdots$ \up Z\SUB{n}}
\newcommand{\dees}[2]{d\SUB{#1} $\cdots$ d\SUB{#2}}
\startxl{drs-forw}
\mbox{\BN{W \whys \zees : \cB\SUP{n} (\cB\SUP{m} a b) c}
         {\BN{W \whys /Y\SUB{m} : \cB\SUP{m} a b} {\NLN{W/X : a}} 
{\NLN{X \whys /Y\SUB{m} :
b}} {\forw m}} {\NLN{Y\SUB{m} \zees : c}} {\forw n}} \\[3mm]
\rule{2in}{0pt}$\arr$ \\[-2mm]
\mbox{\BN{W \whys \zees : \cB\SUP{m+n-1} a (\cB\SUP{n} b c)}
         {\NLN{W/X : a}} {\BN{X \whys \zees : \cB\SUP{n} b c} 
{\NLN{X \whys /Y\SUB{m} : b}}
{\NLN{Y\SUB{m} \zees : c}} {\forw n}} {\forw m+n-1}}
\stopx

\startxl{drs-back}
\mbox{\BN{W \whys \zees : \cB\SUP{m+n-1} a (\cB\SUP{n} b c)}
         {\BN{X \whys \zees : \cB\SUP{n} b c} {\NLN{Y\SUB{m} \zees : c}} 
{\NLN{X \whys
\?Y\SUB{m} : b}} {\back n}} {\NLN{W\?X : a}} {\back m+n-1}}\\[3mm]
\rule{2in}{0pt}$\arr$ \\[-2mm]
\mbox{\BN{W \whys \zees : \cB\SUP{n} (\cB\SUP{m} a b) c}
         {\NLN{Y\SUB{m} \zees : c}} {\BN{W \whys \?Y\SUB{m} : \cB\SUP{m} a b} 
{\NLN{X \whys
\?Y\SUB{m} : b}} {\NLN{W\?X : a}} {\back m}} {\back n}}
\stopx

\lemma{ $\arr$ preserves equivalence of derivations.}

\pf{When the semantic terms from the roots of the left-hand derivation 
and right-hand derivation are compared by applying each of them to sufficiently
many arguments so that all reductions take place, the results are identical:

\begin{tabbing}
\cB\SUP{n} (\cB\SUP{m} a b) c \dees{1}{n+m-1}\\
 = \cB\SUP{m} a b (c \dees{1}{n}) \dees{n+1}{n+m-1}\\ = a (b (c \dees{1}{n})
\dees{n+1}{n+m-1})\\ \\
\cB\SUP{m+n-1} a (\cB\SUP{n} b c) \dees{1}{n+m-1} \\
 = a (\cB\SUP{n} b c \dees{1}{n+m-1}) \\ = a (b (c \dees{1}{n}) 
\dees{n+1}{n+m-1})
\end{tabbing}
}

Let $\larr$ be the converse of $\arr$.  Let $\longleftrightarrow$ be $ \arr
\cup \larr$.  Let $\darr$ be the reflexive transitive closure of $\arr$ and
similarly, $\invdarr$ the reflexive transitive closure of $\larr$, and $\drpl$
the reflexive transitive closure of $\longleftrightarrow$.

Note that $\drpl$ is an equivalence relation, and that $d1 \: \drpl \: d2 \;
\implies d1, d2$ are equivalent, but the converse does not hold, because two
categories could be accidentally equivalent --- an odd property for a
linguistic analysis to have, but a possible one nonetheless.  The reader may
verify that no other combinatory rules may be substituted for
\forw\ in \ref{drs-forw} and \back\ in \ref{drs-back} to yield a 
semantics-preserving derivation rewrite rule.  In particular, the two
combinatory rules must be of the same directionality.

\theorem{ \label{drs-upper-bound} For a derivation with $n$ internal nodes,
every sequence of applications of $\arr$ is finite and is of length at most
$n(n-1)/2$.}

\newcommand{\wate}{\mbox{{\em weight}}}
\newcommand{\skore}{\mbox{{\em score}}}
\newcommand{\kost}{\mbox{{\em cost\hspace{.3em}}}}

\pf{Every derivation with $n$ internal nodes is assigned a positive integer
score which is bounded by $n(n-1)/2$.  An application of $\arr$ is guaranteed
to yield a derivation with a lower score.  This is done by defining the
functions \wate\ and \skore\ for each node of the derivation as follows:

\[ 
\begin{array}{lll}
\wate(x) & \!\!=\!\! & \left\{
\begin{array}{ll}
0 \mbox{\ if $x$ is a leaf node}&\\
 1 + \wate(\mbox{left-child(x)}) + \wate(\mbox{right-child(x)}) &
\mbox{otherwise}
\end{array}\right. \\ 
\mbox{\rule{0pt}{.1mm}} & & \\
\skore(x) & \!\!=\!\! & \left\{
\begin{array}{ll}
0 \mbox{\ if $x$ is a leaf node} &\\
\skore(\mbox{left-child(x)})+\skore(\mbox{right-child(x)})+
	\wate(\mbox{left-child(x)})& 
			\mbox{\hspace{-0.5em}otherwise}
\end{array}\right. 
\end{array}
\]

Each application of $\arr$ decreases the score of the derivation.  This follows
{}from monotonic dependency of the score of the root of the derivation upon the
scores of each sub-derivation, and from the fact that locally, the score of a
redex decreases when $\arr$ is applied: In \reffig{score}, a derivation is
schematically depicted with a redex whose sub-constituents are named a, b, and
c.  Applying $\arr$ reduces the \skore(e), hence the score of the whole
derivation.

\begin{figure}
\centerline{\psfig{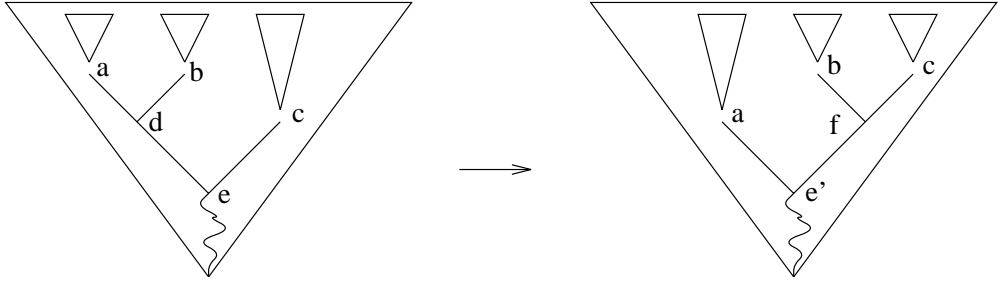}}
\caption{Schema for one redex in DRS}
\label{score}
\end{figure}

in redex:
\begin{eqnarray*}
\wate(d)  & = & \wate(a) + \wate(b) + 1 \\
\skore(d) & = & \skore(a) + \skore(b) + \wate(a) \\
\skore(e) & = & \skore(d) + \skore(c) + \wate(d) \\
          & = & \skore(a) + \skore(b) + \wate(a) + \skore(c) + \wate(a) + 
\wate(b) + 1 \\ & =
& \skore(a) + \skore(b) + \skore(c) + \wate(b) + 2\cdot\wate(a) + 1
\end{eqnarray*} 

in contractum:
\begin{eqnarray*}
\skore(f)  & = & \skore(b) + \skore(c) + \wate(b) \\
\skore(e') & = & \skore(a) + \skore(f) + \wate(a) \\
           & = & \skore(a) + \skore(b) + \skore(c) + \wate(b) + 
\wate(a) \\ & < & \skore(a) +
\skore(b) + \skore(c) + \wate(b) + 2\cdot\wate(a) + 1
\end{eqnarray*}} 

I now show that $n(n-1)/2$ is also the lower bound on sequences of application
of $\arr$.

A {\em left-chain\/} is either a leaf node or a derivation whose left-child is
a left-chain and whose right-child is a leaf node.  A {\em right-chain\/} is
either a leaf node or a derivation whose right-child is a right-chain and whose
left-child is a leaf node.  A {\em quasi-right-chain\/} is a derivation whose
right-child is a leaf and whose left-child is a right-chain.

\lemma{A quasi-right-chain of $n$ internal nodes can be rewritten using $n-1$
application of $\arr$ to a right-chain.}

\begin{figure} 
\centerline{\psfig{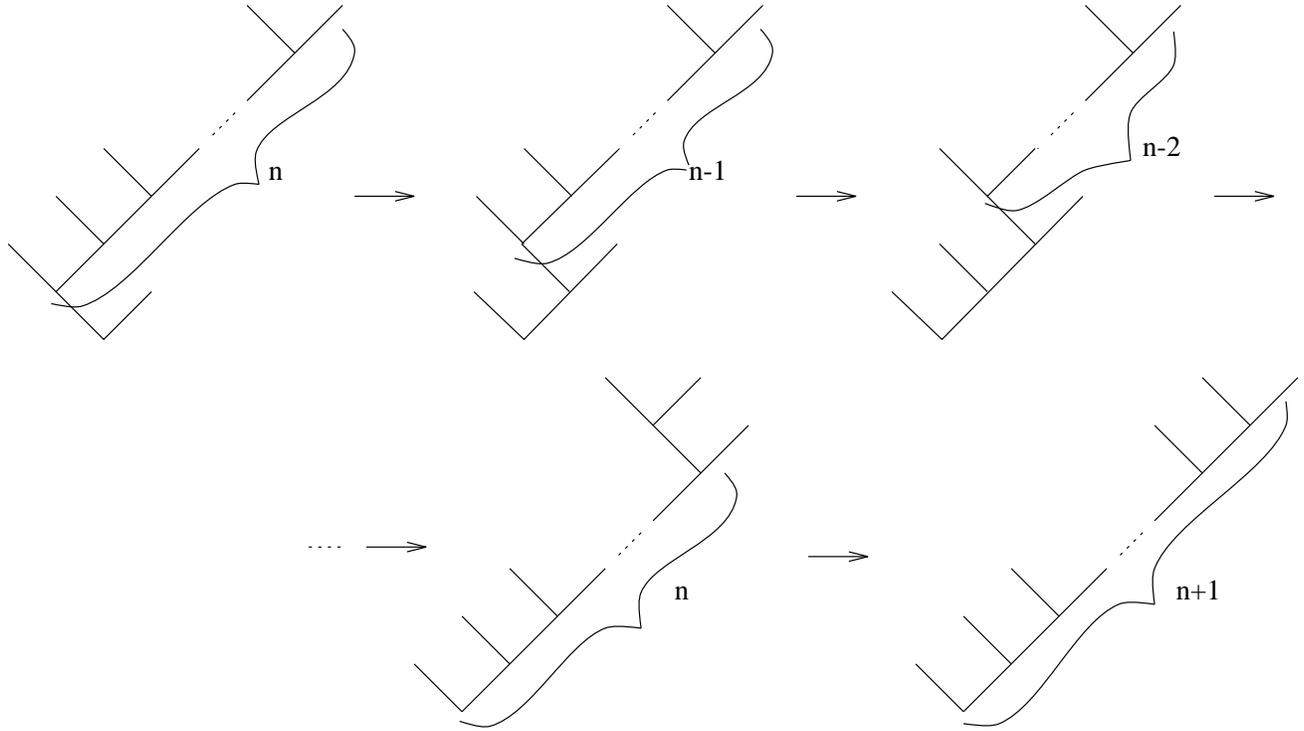}} 
\caption{Normal form computation of a quasi-right-chain}
\label{zip-rb} 
\end{figure}

\pf{At every point in the rewriting operation, there is only one redex.  This
is depicted in \reffig{zip-rb}}

\lemma{A left-chain C of $n$ internal nodes can be rewritten to a right-chain
using a sequence of exactly $n(n-1)/2$ applications of $\arr$.}

\pf{By induction on $n$.
\begin{description}
\item[{\rm $n=1:$}] C is already a right-chain:  0 steps are required.
\item[{\rm $n=2:$}] C is a redex. One application of $\arr$ rewrites it into a 
right-chain.
\item[{\rm $n>2:$}] Suppose this is true for all $m<n$. Apply $\arr$ as 
follows: Rewrite the left-child of C to a right-chain in $(n-1)(n-2)/2$ steps.
The result of this is a quasi-right-chain of $n$ internal nodes, which can be
rewritten to a right-chain using $n-1$ applications of $\arr$.  The total
number of applications of $\arr$ is
\[ \frac{(n-1)(n-2)}{2} + n-1 = \frac{n^2-3n+2+2n-2}{2} = \frac{n^2-n}{2} = 
\frac{n(n-1)}{2} \]
\end{description}
}

A rewrite system is {\em strongly normalizing} (SN) iff every sequence of
applications of $\arr$ is finite.

\corollary{DRS is SN.}
\pf{Immediate corollary of theorem~\ref{drs-upper-bound}.}

So far I have shown that nondeterministic computation of the right-branching NF
of a derivation is quite tractable: quadratic in the size of the derivation.  I
will now show that this is even so on a deterministic machine.

A rewrite system is {\em Church-Rosser (CR)} just in case \[ \forall x,y . (x
\drpl y \implies \exists z . (x \darr z \And y \darr z)) \]

A rewrite system is {\em Weakly Church-Rosser (WCR)} just in case \[ \forall
x,y,w . (w \arr x \And w \arr y) \implies \exists z . (x \darr z \And y \darr
z) \]

\lemma{\label{l:wcr} DRS is WCR.} 

\begin{figure}
\centerline{\psfig{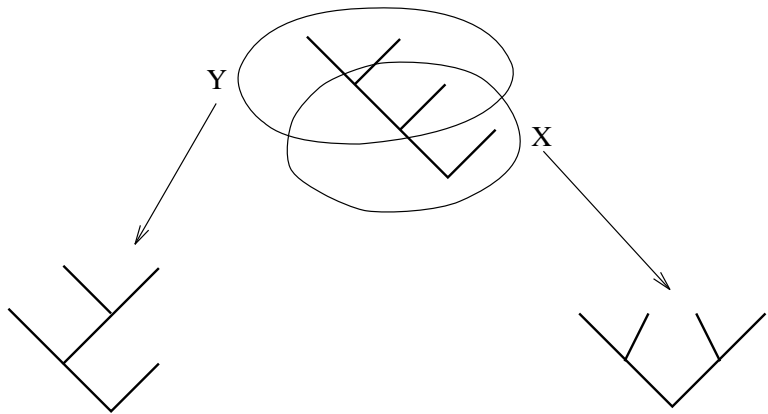}}
\caption{When two redexes are not independent}
\label{conflict}
\end{figure}

\pf{Let $w$ be a derivation with two distinct redexes $x$ and $y$, yielding
the two distinct derivations $w'$ and $w''$ respectively.  There are a few
possibilities:
\kases
\kase{1} $x$ and $y$ have no nodes in common.  There are three subcases: 
$x$ could dominate $y$ (include $y$ as a subconstituent), $x$ could be
dominated by $y$, or $x$ and $y$ could incomparable with respect to dominance.
Either way, it is clear that the order of application of $\arr$ makes no
difference.
\kase{2} $x$ and $y$ share nodes.  Assuming that $x$ and $y$ are distinct, and
without loss of generality, that $y$ does not dominate $x$, we have the
situation depicted in
\reffig{conflict}.  (Note that all three internal nodes in \reffig{conflict} 
are of the same combinatory rule, either \forw\ or \back.) In this case, there
does exist a derivation $z$ such that $w'
\darr z \And w'' \darr z$.  This is depicted in \reffig{wcr}.
\Endkases
}

\begin{figure}
\centerline{\psfig{figure=wcr.eps,silent=y}}
\caption{Why DRS is weakly Church-Rosser}
\label{wcr}
\end{figure}

\lemma{\label{newman} (Newman 1942\nocite{Newman42}) 
$\mbox{WCR} \And \mbox{SN} \implies \mbox{CR}$.}

\theorem { DRS is CR. }

\pf{Follows from lemmas \ref{newman}, \ref{l:wcr}, and corollary 1.}

\corollary{\label{uniqueNF}$\mbox{CR} \implies \forall x,y . (x \drpl y \And 
x,y  \mbox{ are NFs }) \implies x=y$.}

\pf{By contradiction: suppose $\mbox{CR} \And \exists x,y . (x \drpl y \And
x,y \mbox{ are NFs }) \And x \neq y.$ Then $\exists z . (x \darr z \And y
\darr z)$.  Given that $x$ is a NF, $x \darr z$ entails that $x = z$.
Similarly $y=z$.  This contradicts the assumption that $x \neq y$.}

Therefore every DRS equivalence class contains exactly one NF derivation.  It
follows that any deterministic computational path of applying $\arr$ will lead
to the normal form.

As for the efficiency of computing the right branching NF for a derivation of
$n$ internal nodes, theorem~\ref{drs-upper-bound} shows that for a derivation
of $n$ internal nodes, every sequence of applications of $\arr$ is at most
$n(n-1)/2$ steps long.  This is the worst case, which arises from applying
$\arr$ as far away from the root as possible.  An inspection of the definition
of \skore\ suggests that applying $\arr$ as close to the root as possible
yields the largest decrease in \skore\, since \wate(a) is maximized.  In fact,
in case it is always grammatically possible to apply $\arr$ to the closest
redex to the root, every derivation has a CTR (closest to root) rewrite
sequence of length $O(n)$.  The proof requires defining a function which
measures the number of CTR rewrite steps that a derivation requires to reach
NF.  Let us first define the function \CTRarr\ which applies $\arr$ once to the
closest redex to the root of its argument.

\newsavebox{\kombine} \newlength{\kombineL}
\newsavebox{\kombineX}\newlength{\kombineXL}
\savebox{\kombine}{$\CTRarr$(combine(} 
\settowidth{\kombineL}{\usebox{\kombine}}
\savebox{\kombineX}{$\CTRarr$(combine(combine(} 
\settowidth{\kombineXL}{\usebox{\kombineX}}

{
\newcommand{\mumback}{\hspace{-0mm}}
\[
\CTRarr(x) = \left\{
\begin{array}{ll}
x & \mbox{\mumback if $x$ is a leaf node}\\
\mbox{combine(left-child($x$),}\CTRarr(\mbox{right-child($x$)})) & 
				\mbox{\mumback if left-child($x$) is a leaf} \\
\CTRarr\mbox{(combine(left-child(left-child($x$)),} & 
			\mbox{\mumback otherwise}  \\
\mbox{\hspace*{\kombineL}combine(right-child(left-child($x$)),} & \\
\mbox{\hspace*{\kombineXL}right-child($x$))))} & {}
\end{array}\right. 
\]
} Let \kost($x$) be the number of times that \CTRarr\ must be iterated on $x$
so as to yield an NF. \CTRarr\ defines \kost($x$) by the following recurrence
equations:

\savebox{\kombine}{$1 + \kost(\mbox{combine(}$}
\settowidth{\kombineL}{\usebox{\kombine}}

\savebox{\kombineX}{$1 + \kost(\mbox{combine(combine(}$}
\settowidth{\kombineXL}{\usebox{\kombineX}}

\[
\kost(x) = \left\{
\begin{array}{ll}
0 & \mbox{if $x$ is a leaf node} \\
\kost(\mbox{right-child($x$)}) & \mbox{if left-child($x$) is a leaf}\\
1+\kost(\mbox{combine(left-child(left-child($x$)),} & \mbox{otherwise} \\
\mbox{\hspace*{\kombineL}combine(right-child(left-child($x$)),} \\
\mbox{\hspace{\kombineXL}right-child($x$))))} & 
\end{array}\right. 
\]

Observe that in the third case, subsequent applications of \CTRarr\ will
`process' all of left-child(left-child($x$)), then proceed to `process'
right-child(left-child($x$)), and finally process right-child($x$).  This is
illustrated in \reffig{ctrFig}.

\begin{figure}
\centerline{\psfig{figure=ctr.eps,silent=y}}
\caption{One application of \CTRarr}
\label{ctrFig}
\end{figure}

The cost of doing these three steps can be accounted for separately:

\[
\begin{array}{ll}
\kost(x) = & 1 + 	\\
           & \kost(\mbox{combine(left-child(left-child($x$)),$l_0$)}) +\\
           & \kost(\mbox{combine(right-child(left-child($x$)),$l_0$)}) +\\
           & \kost(\mbox{right-child($x$)})
\end{array}
\]

(where $l_0$ is a dummy leaf node.)  

\newcommand{\NN}{\mbox{\#}}

It is now possible to prove by induction on the derivation tree that 

\[\kost(x)=\mbox{\NN$x$ minus the depth of the rightmost leaf in $x$}\]

(where \NN$D$ is the number of internal nodes in the derivation $D$)

\begin{flushleft}
Base cases: \\
$x$ is a leaf: $\kost(x) = 0$\\
$x$ has one internal node:  $\kost(x) = 0$\\[1ex]
Induction:\\
Suppose true for all trees of fewer internal nodes than \NN$x$.
Let $d$ be the depth of the rightmost leaf of $x$.\\
Case 1:  left-child($x$) is a leaf: 
$\kost(x) = \kost(\mbox{right-child(x)}) = (\NN$x$-1) - (d-1) = \NN$x$-d$\\
Case 2: left-child($x$) is not a leaf:\\
Let $a, b, c$ be left-child(left-child($x$)),
right-child(left-child($x$)), right-child($x$), respectively.  
Note that $\NN x = 2+\NN a+\NN
b+\NN c$. 
\end{flushleft}
\[
\begin{array}{ll}
\kost($x$) & = 1 + \kost(\mbox{combine($a$,$l_0$)})+%
\kost(\mbox{combine($b$,$l_0$)})+\kost(c)\\
           & = 1 + \NN a + \NN b + \NN c - \mbox{depth of rightmost leaf in 
$c$}\\
           & = \NN x - 1 - (d-1)\\
           & = \NN x - d 
\end{array}
\] \hfill $\Box$

So while the worst-case sequence of applications of $\arr$ is quadratic in the
size of derivation, the best case (possible as long as the grammar allows it)
is linear.

  \vspace{\fill}

  \centerline{\psfig{figure=horse.eps,silent=y,height=0.7875in,width=2.25in
						}}

  \vspace{\fill}
\addcontentsline{toc}{chapter}{Bibliography}

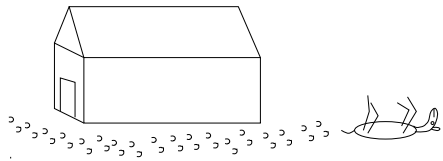
\end{document}